\documentclass[10pt]{article} 
\usepackage[accepted]{tmlr}

\usepackage{amsmath,amsfonts,bm}

\def\eqref#1{equation~\ref{#1}}

\def\1{\bm{1}}

\DeclareMathAlphabet{\mathsfit}{\encodingdefault}{\sfdefault}{m}{sl}
\SetMathAlphabet{\mathsfit}{bold}{\encodingdefault}{\sfdefault}{bx}{n}

\usepackage{hyperref}
\usepackage{url}
\usepackage{graphicx}
\usepackage{booktabs}
\usepackage{multirow}
\usepackage{wrapfig}
\usepackage{adjustbox}
\usepackage{tikz}
\usepackage{multirow}
\usepackage{arydshln}
\usepackage{colortbl}  
\definecolor{mygray}{gray}{.9}
\definecolor{mygreen}{RGB}{93,173,85}
\definecolor{mycolor}{HTML}{cbf1f5}
\usepackage{arydshln}

\usetikzlibrary{positioning,matrix,calc,arrows.meta,decorations.pathreplacing,shapes.geometric}
\usepackage{pgfplotstable}

\definecolor{mywarning}{RGB}{233,144,61}
\definecolor{DarkRed}{RGB}{0,0,0}
\definecolor{azure}{rgb}{0.0, 0.5, 1.0}
\definecolor{gray}{rgb}{0.3, 0.3, 0.3}
\definecolor{DarkGreen}{RGB}{42,110,63}
\newcommand{\hlg}[1]{\textcolor{mygreen}{#1}}
\newcommand{\hlr}[1]{\textcolor{red}{#1}} 

\usepackage{pifont}     

\usepackage{fontawesome5}  

\usepackage{amssymb}  
\usepackage{xcolor}   

\newcommand{\tmark}{\ding{51}} 
\newcommand{\xmark}{\ding{55}} 
\usepackage{latexsym}

\usepackage{verbatim}

\definecolor{mycolor_yellow}{RGB}{254, 248, 230}
\definecolor{mycolor_blond}{RGB}{248, 205, 172}
\definecolor{mycolor_green}{RGB}{224, 255, 205}
\definecolor{mycolor_root}{RGB}{237, 247, 152}
\definecolor{mycolor_pink}{RGB}{250, 219, 223}

\usepackage{tikz}
\usepackage{forest}

\title{A Survey on Federated Fine-Tuning\\ of Large Language Models}

\author{\name Yebo Wu \email yc37926@um.edu.mo \\
      \addr State Key Laboratory of IOTSC, University of Macau
      \AND
      \name Chunlin Tian \email yc27402@um.edu.mo \\
      \addr State Key Laboratory of IOTSC, University of Macau
      \AND
      \name Jingguang Li \email mc45005@um.edu.mo\\
      \addr State Key Laboratory of IOTSC, University of Macau
      \AND
      \name He Sun \email hesun@um.edu.mo\\
      \addr State Key Laboratory of IOTSC, University of Macau
      \AND
      \name Kahou Tam \email yc37436@um.edu.mo\\
      \addr State Key Laboratory of IOTSC, University of Macau
      \AND
      \name Zhanting Zhou \email  ztzhou@std.uestc.edu.cn\\
      \addr University of Electronic Science and Technology of China
      \AND
      \name Haicheng Liao \email yc27979@um.edu.mo\\
      \addr State Key Laboratory of IOTSC, University of Macau
      \AND
      \name Jing Xiong \email junexiong@connect.hku.hk\\
      \addr The University of Hong Kong
      \AND
      \name Zhijiang Guo\thanks{\quad \small Corresponding author.} \email zhijiangguo@hkust-gz.edu.cn\\
      \addr Hong Kong University of Science and Technology (Guangzhou)
      \AND
      \name Li Li\footnotemark[1] \email llili@um.edu.mo\\
      \addr State Key Laboratory of IOTSC, University of Macau
      \AND
      \name Chengzhong Xu \email czxu@um.edu.mo\\
      \addr State Key Laboratory of IOTSC, University of Macau}

\def\month{02}  
\def\year{2026} 
\def\openreview{\url{https://openreview.net/forum?id=rnCqbuIWnn}} 

\begin{document}

\maketitle




\begin{abstract}
Large Language Models (LLMs) have demonstrated impressive success across various tasks. Integrating LLMs with Federated Learning (FL), a paradigm known as FedLLM, offers a promising avenue for collaborative model adaptation while preserving data privacy. This survey provides a systematic and comprehensive review of FedLLM. We begin by tracing the historical development of both LLMs and FL, summarizing relevant prior research to set the context. Subsequently, we delve into an in-depth analysis of the fundamental challenges inherent in deploying FedLLM. Addressing these challenges often requires efficient adaptation strategies; therefore, we conduct an extensive examination of existing Parameter-Efficient Fine-tuning (PEFT) methods and explore their applicability within the FL framework. To rigorously evaluate the performance of FedLLM, we undertake a thorough review of existing fine-tuning datasets and evaluation benchmarks. Furthermore, we discuss FedLLM's diverse real-world applications across multiple domains. Finally, we identify critical open challenges and outline promising research directions to foster future advancements in FedLLM. This survey aims to serve as a foundational resource for researchers and practitioners, offering valuable insights into the rapidly evolving landscape of federated fine-tuning for LLMs. It also establishes a roadmap for future innovations in privacy-preserving AI. We actively maintain a \href{https://github.com/Clin0212/Awesome-Federated-LLM-Learning}{GitHub repo} to track cutting-edge advancements in this field.
\end{abstract}
\section{Introduction}

Large Language Models (LLMs), such as GPT-4o~\citep{gpt4o}, DeepSeek-R1~\citep{guo2025deepseek}, and Qwen3~\citep{yang2025qwen3} have exhibited extraordinary proficiency across a spectrum of downstream tasks~\citep{xu2024copyrightmeter,xu2025videoeraser}.
These LLMs, distinguished by their ability to capture complex semantic knowledge, have established new performance benchmarks in computational linguistics~\citep{xiong2025parallelcomp,xiong2024uncomp,xu2026fraudshield}. However, despite their impressive capabilities, LLMs cannot be directly deployed for specific downstream tasks without appropriate adaptation~\citep{hu2021lora}. 
Furthermore, training LLMs directly on downstream task datasets presents substantial challenges. The massive scale of model parameters leads to significant computational overhead~\citep{tian2025clone}, while the scarcity of task-specific data constrains effective model training and increases the risk of overfitting.
For example, training LLaMA2-65B involves processing approximately 1.4 trillion tokens, requiring 21 days of computation on 2,048 NVIDIA A100 GPUs~\citep{llama}. 
Consequently, fine-tuning pre-trained LLMs has become the dominant paradigm~\citep{dodge2020fine}, enabling more efficient adaptation of LLMs to specific tasks while preserving their foundational knowledge acquired during pre-training.

\begin{figure*}[!t]
    \centering
    \includegraphics[width=0.8\linewidth]{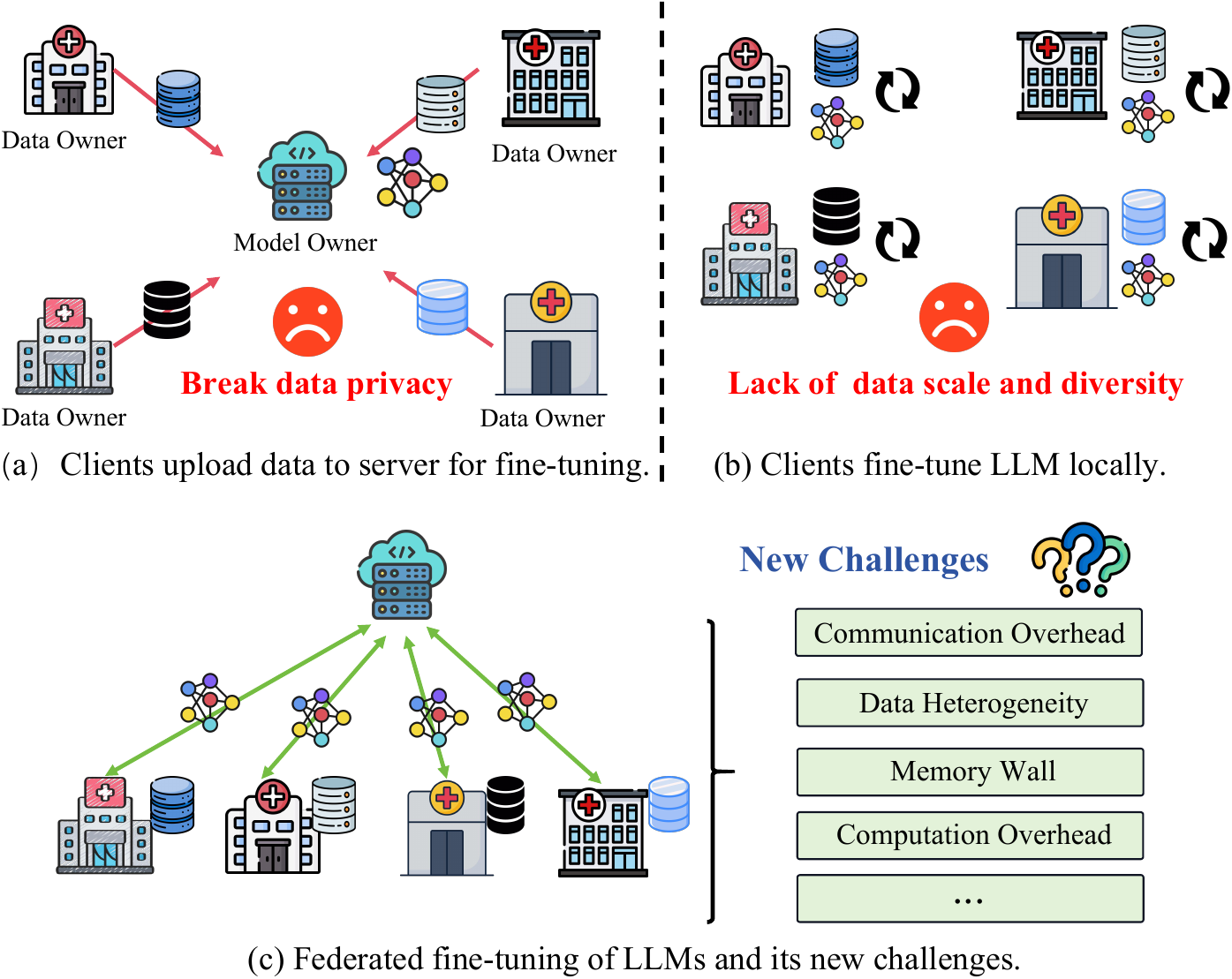}
    \vspace{-2mm}
    \caption{Illustration of three LLM fine-tuning paradigms: (a) \textbf{Centralized Fine-tuning}, where data is aggregated at a central server; (b) \textbf{Local Fine-tuning}, where models are trained independently on private datasets; and (c) \textbf{Federated Fine-tuning}, where data remains local, and model updates are aggregated by a central server to create a global model.}
    \label{fig1}
    \vspace{-3mm}
\end{figure*}

The current mainstream LLM fine-tuning paradigms can be categorized into three approaches: 1) \textbf{Centralized Fine-Tuning} (as shown in Figure~\ref{fig1}(a)): This approach aggregates local datasets from all data owners (clients) and uploads them to a central server for fine-tuning~\citep{zhang2023instruction, ding2023parameter}. 
Despite its effectiveness, this approach raises significant privacy concerns~\citep{wang2023fedins2, ye2024praffl, tam2024towards} and is often impractical in real-world scenarios due to legal restrictions (e.g., GDPR~\citep{voigt2017eu}), which limit the centralization of sensitive personal data. 2) \textbf{Local Fine-Tuning} (as shown in Figure~\ref{fig1}(b)): In this paradigm, each data owner fine-tunes the LLM locally using their private dataset. While this approach preserves data privacy, the limited size and diversity of local datasets often result in suboptimal model performance. For instance, models refined through local fine-tuning demonstrate a substantial performance degradation of up to 7\% on the MMLU benchmark~\citep{hendrycks2020measuring} when compared to federated fine-tuning~\citep{ye2024openfedllm}. 3) \textbf{Federated Fine-Tuning} (as shown in Figure~\ref{fig1}(c)): This approach enables collaborative model improvement while preserving data privacy by allowing clients to train the model locally and only sharing model updates with the central server~\citep{li2025star, li2023reconfigurable}. The server aggregates these updates to construct a global model, which is subsequently redistributed to clients for further refinement. This method addresses both the privacy concerns of centralized fine-tuning and the limited data diversity issues of local fine-tuning, making it a promising paradigm for adapting LLMs to specific downstream tasks~\citep{xu2023federated}.

Despite these benefits, federated fine-tuning encounters several unique challenges, which significantly limit the effective deployment of FedLLM in real-world scenarios:
1) \textbf{Communication Overhead}: LLMs typically contain billions of parameters, such as LLaMA2-7B. Therefore, uploading these massive model parameters in each training round incurs substantial communication overhead, resulting in severe communication latency and excessive bandwidth requirements~\citep{li2022federated}.
2) \textbf{Data Heterogeneity}: Data across participating clients exhibits substantial variation in both quality and statistical distribution~\citep{ning2024fedgcs}. This Non-IID (Non-Independent and Identically Distributed) nature of federated data can introduce significant biases into model updates, leading to weight divergence, slower convergence rates, and ultimately compromised model performance~\citep{fu2024virtual, tian2024ranking}.
3) \textbf{Memory Wall}: Participating clients, especially edge devices, generally possess limited available memory resources~\citep{wu2024heterogeneity, wu2024neulite, zhan2024heterogeneity, li2023breaking, wu2025memory}, which insufficiently support memory-intensive LLM fine-tuning. 
This memory wall fundamentally limits clients' effective participation in the federated fine-tuning process, preventing them from contributing valuable data to the global model and ultimately compromising model performance.
4) \textbf{Computation Overhead}: The hardware processing capabilities of participating clients are often limited~\citep{wang2019adaptive}, making it challenging to meet the high computational demands of fine-tuning LLMs. This computational bottleneck substantially increases local training time and consequently prolongs the overall process. The extended training cycles reduce system efficiency and significantly increase energy consumption on resource-constrained devices, potentially deterring client participation.

To address these challenges, researchers have applied various Parameter-Efficient Fine-Tuning (PEFT) methods to FL, which can be broadly classified into five main categories: LoRA-based tuning~\citep{hu2021lora, tian2024hydralora, wu2025learning}, prompt-based tuning~\citep{lester2021power}, adapter-based tuning~\citep{houlsby2019parameter}, selective-based tuning~\citep{zaken2021bitfit}, and other tuning methods~\citep{shin2023fedsplitx, chen2017zoo}.
The core idea behind these methods is to minimize the number of trainable parameters by focusing on small, task-specific adjustments rather than fine-tuning the entire model. By updating only a subset of parameters or modifying model inputs (e.g., prompts), these approaches significantly reduce the communication overhead, computational burden, and memory usage of model fine-tuning, all while maintaining the performance of LLMs across diverse tasks.

\begin{table}[!t]
\centering
\caption{\textbf{Overview of related surveys.} This comparison highlights whether each work addresses data privacy, aligns with the scope of LLMs, emphasizes efficiency, proposes evaluation benchmarks, and discusses applications and future directions.}
\label{tab:survey_compare}
\resizebox{0.8\linewidth}{!}{
  \begin{tabular}{l|cccccc}
    \toprule[2pt]
    \textbf{Prior Surveys}& \textbf{Privacy} & \textbf{LLM}  &\multirow{1}{*}{\textbf{Efficiency}} &\multirow{1}{*}{\textbf{Benchmark}} &\textbf{Application} &\textbf{Future Direction}\\  
    \midrule[1pt]
    \cite{xu2024survey}&\hlr{\xmark}&\hlg{\tmark{}}&\hlg{\tmark{}}&\hlr{\xmark}&\hlg{\tmark{}}&\hlg{\tmark{}}\\

    \cite{zhao2023survey}&\hlr{\xmark}&\hlg{\tmark{}}&\hlg{\tmark{}}&\hlg{\tmark}&\hlg{\tmark}&\hlg{\tmark}\\

    \cite{gao2024llm}&\hlr{\xmark}&\hlg{\tmark{}}&\hlr{\xmark}&\hlg{\tmark{}}&\hlr{\xmark}&\hlg{\tmark{}}\\

    \cite{li2024survey}&\hlr{\xmark}&\hlg{\tmark{}}&\hlr{\xmark}&\hlg{\tmark{}}&\hlg{\tmark{}}&\hlg{\tmark{}}\\

    \cite{zheng2023survey}&\hlr{\xmark}&\hlg{\tmark{}}&\hlr{\xmark}&\hlg{\tmark{}}&\hlg{\tmark{}}&\hlg{\tmark{}}\\

    \cite{han2024parameter}&\hlr{\xmark}&\hlg{\tmark{}}&\hlg{\tmark{}}&\hlr{\xmark}&\hlg{\tmark{}}&\hlg{\tmark{}}\\

    \cite{xin2024parameter}&\hlr{\xmark}&\hlg{\tmark{}}&\hlg{\tmark{}}&\hlg{\tmark{}}&\hlg{\tmark{}}&\hlg{\tmark{}}\\

    \midrule[0.5pt]
    
    \cite{huang2024federated} & \hlg{\tmark{}}&\hlr{\xmark}&\hlr{\xmark}&\hlg{\tmark{}}&\hlr{\xmark}&\hlg{\tmark{}}\\
    
    \cite{ye2023heterogeneous}& \hlg{\tmark{}}&\hlr{\xmark}&\hlr{\xmark}&\hlr{\xmark}&\hlr{\xmark}&\hlg{\tmark{}}\\

    \cite{jiang2024blockchained}& \hlg{\tmark{}}&\hlr{\xmark}&\hlr{\xmark}&\hlr{\xmark}&\hlg{\tmark{}}&\hlr{\xmark}\\
    
    \cite{yuan2024decentralized}& \hlg{\tmark{}}&\hlr{\xmark}&\hlr{\xmark}&\hlr{\xmark}&\hlg{\tmark{}}&\hlg{\tmark{}}\\

    \cite{chen2024federated}& \hlg{\tmark{}}&\hlr{\xmark}&\hlr{\xmark}&\hlg{\tmark}&\hlg{\tmark}&\hlg{\tmark}\\

    \cite{chai2024survey}& \hlg{\tmark{}}&\hlr{\xmark}&\hlr{\xmark}&\hlg{\tmark{}}&\hlr{\xmark}&\hlg{\tmark{}}\\

    \cite{feng2023fedmultimodal}& \hlg{\tmark{}}&\hlr{\xmark}&\hlr{\xmark}&\hlg{\tmark{}}&\hlg{\tmark{}}&\hlg{\tmark{}}\\

    \cite{zhang2024survey}& \hlg{\tmark{}}&\hlr{\xmark}&\hlr{\xmark}&\hlr{\xmark}&\hlr{\xmark}&\hlg{\tmark{}}\\

    \midrule[0.5pt]

    \cite{yao2024federated} & \hlg{\tmark{}}&\hlg{\tmark{}}&\hlg{\tmark}&\hlr{\xmark}&\hlg{\tmark}&\hlg{\tmark}\\
    
    \cite{li2024synergizing}& \hlg{\tmark{}}&\hlg{\tmark{}}&\hlg{\tmark{}}&\hlr{\xmark}&\hlg{\tmark{}}&\hlg{\tmark{}}\\
    \cite{zhuang2023foundation}&\hlg{\tmark{}}&\hlg{\tmark{}}&\hlr{\xmark}&\hlr{\xmark}&\hlg{\tmark{}}&\hlg{\tmark{}}\\

    \cite{woisetschlager2024survey}&\hlg{\tmark{}}&\hlg{\tmark{}}&\hlg{\tmark{}}&\hlr{\xmark}&\hlr{\xmark}&\hlg{\tmark{}}\\
    \cite{ren2024advances}&\hlg{\tmark{}}&\hlg{\tmark{}}&\hlg{\tmark{}}&\hlr{\xmark}&\hlg{\tmark{}}&\hlg{\tmark{}}\\

    \midrule[1pt]

    \textbf{Ours}&\hlg{\tmark{}}&\hlg{\tmark{}}&\hlg{\tmark{}}&\hlg{\tmark{}}&\hlg{\tmark{}}&\hlg{\tmark{}}\\

    \bottomrule[1pt]
  \end{tabular}
}
\vspace{-5mm}
\end{table}

\noindent
\textbf{Prior Surveys.} Despite the continuous development of innovative federated fine-tuning methods, a significant gap remains in the comprehensive evaluation and comparison of these techniques. While existing surveys on FL provide valuable insights, they are typically limited to either traditional small-model FL settings or fail to offer a detailed analysis and evaluation benchmark specifically for federated fine-tuning of LLMs. Concurrently, although several surveys on PEFT have been proposed, these works predominantly focus on centralized fine-tuning scenarios, overlooking the unique challenges that arise when adapting such techniques to distributed, privacy-preserving settings. A detailed comparison of existing surveys is summarized in Table~\ref{tab:survey_compare}.
To fill this gap, our survey aims to be the \textit{first} to present a systematic examination of federated fine-tuning for LLMs, providing a thorough understanding of their evolution, effectiveness, and practical implementation challenges, along with standardized evaluation benchmarks that enable fair comparison across different approaches. 
This thorough analysis serves as a foundation for researchers and practitioners seeking to navigate the rapidly evolving landscape of federated LLM fine-tuning.

\noindent
\textbf{Contribution.} This paper presents a comprehensive survey on the federated fine-tuning of LLMs. In contrast to existing surveys, the main contributions of this work can be summarized as follows:

\begin{enumerate}
    \item We provide an exhaustive review of all relevant papers on federated fine-tuning up to date, offering an extensive analysis of the state-of-the-art techniques and their evolution in this field.

    \item We conduct a detailed analysis of the key challenges in federated fine-tuning and propose a systematic taxonomy based on different fine-tuning approaches, including LoRA-based, prompt-based, adapter-based, selective-based, and other emerging tuning methods. We further provide an in-depth discussion of the advantages, limitations, and applicability of these methods.

    \item We establish a comprehensive evaluation framework for federated fine-tuning of LLMs, encompassing fine-tuning datasets and evaluation benchmarks across diverse domains, while systematically analyzing and discussing diverse real-world application scenarios.

    \item Finally, we outline promising research directions in FedLLM, aiming to guide future investigations toward more efficient, scalable, and privacy-preserving solutions that bridge the gap between theoretical advances and practical deployments in resource-constrained federated environments.

\end{enumerate}

\begin{figure*}[!t]
    \centering
    \includegraphics[width=0.97\linewidth]{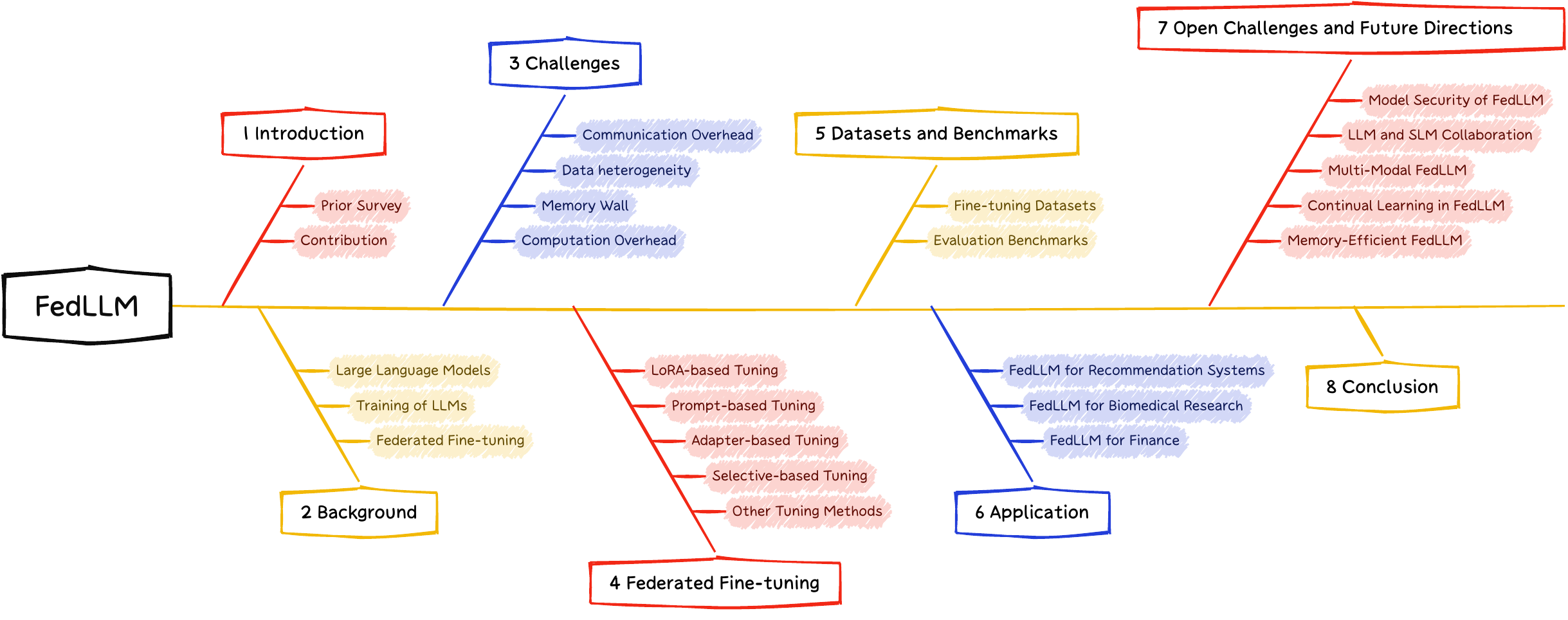}
    \caption{Overall structure of the survey.}
    \label{overview}
    \vspace{-3mm}
\end{figure*}

Figure~\ref{overview} illustrates the organizational structure of this survey. Section~\ref{sec_background} introduces the relevant background and fundamental concepts of LLMs and federated fine-tuning. Section~\ref{sec_challenge} systematically examines the technical challenges and inherent limitations in the federated fine-tuning of LLMs. In Section~\ref{sec_federated_finetuning}, we present a comprehensive review of state-of-the-art federated fine-tuning techniques and methodologies. 
Section~\ref{sec_evaluation} presents representative fine-tuning datasets and evaluation benchmarks across various domains, specifically curated to assess the performance of federated fine-tuning in diverse scenarios. Section~\ref{sec_application} explores practical applications of FedLLM. Section~\ref{sec_outlook} outlines promising research directions, while Section~\ref{sec_conclusion} synthesizes key insights from this survey to inform and guide future research in this rapidly evolving field.

\section{Background}\label{sec_background}


\subsection{Large Language Models}

\begin{wrapfigure}{r}{0.3\linewidth} 
\centering
\vspace{-7mm}
\adjustbox{max width=\linewidth}{\includegraphics{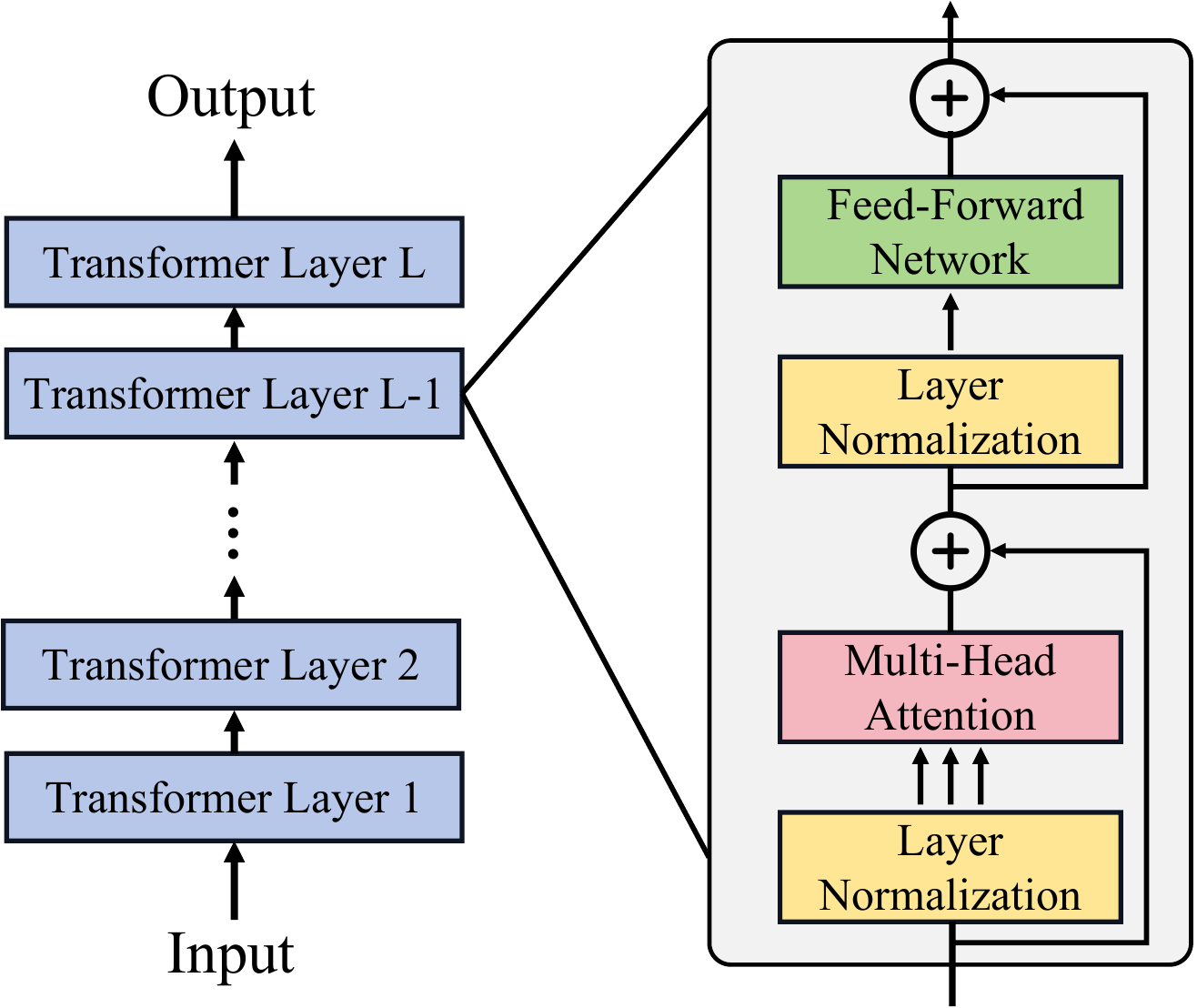}}
\caption{Architecture of LLMs.}
\label{architecture}
\vspace{-6mm}
\end{wrapfigure}

Large Language Models (LLMs) have demonstrated unprecedented capabilities across a wide range of natural language processing tasks, including machine translation~\citep{wang2022progress}, text generation~\citep{yu2022survey}, sentiment analysis~\citep{wankhade2022survey}, and question answering~\citep{zhu2021retrieving}. Their exceptional performance stems from their remarkable ability to encode complex linguistic patterns, capture long-range contextual dependencies, and learn rich semantic representations~\citep{xiong2023dq,xu2026agents}. These capabilities have not only enabled LLMs to achieve state-of-the-art results on a broad spectrum of academic benchmarks, but have also fueled transformative advances in real-world applications such as conversational AI, legal document analysis, medical decision support, and automated content generation.

Architecturally, modern LLMs are typically constructed by stacking dozens or even hundreds of transformer layers, where each layer incrementally refines the input through deep contextualization and abstraction. 
For example, LLaMA2-7B comprises 32 transformer layers stacked sequentially to capture hierarchical linguistic features. This deep, layered architecture enables the model to effectively integrate both local and global contextual information over long sequences, which is essential for complex language understanding tasks.
Figure~\ref{architecture} illustrates the schematic structure of a prototypical LLM, where transformer layers are arranged in a vertically stacked fashion. Each transformer layer consists of two fundamental components: Multi-Head Attention (MHA) and Feed-Forward Network (FFN). Formally, the input to the $l$-th transformer layer is denoted as $h_{l-1} \in \mathbb{R}^{n \times d}$, where $n$ is the sequence length and $d$ is the hidden dimension of the model. The computational process within the $l$-th layer can be expressed as follows:
\begin{align}
    h_{i}^{'} &= \text{MHA}(\text{LN}(h_{i-1})) + h_{i-1}, \\
    h_{i} &= \text{FFN}(\text{LN}(h_{i}^{'})) + h_{i}^{'}
\end{align}
where LN($\cdot$) represents layer normalization, which stabilizes the training dynamics by standardizing the activations, and $h_{i}^{'}$ denotes the intermediate activations after being processed by the MHA module.

\subsection{Training of LLMs}

\begin{figure}[!h]
    \centering
    \includegraphics[width=0.7\linewidth]{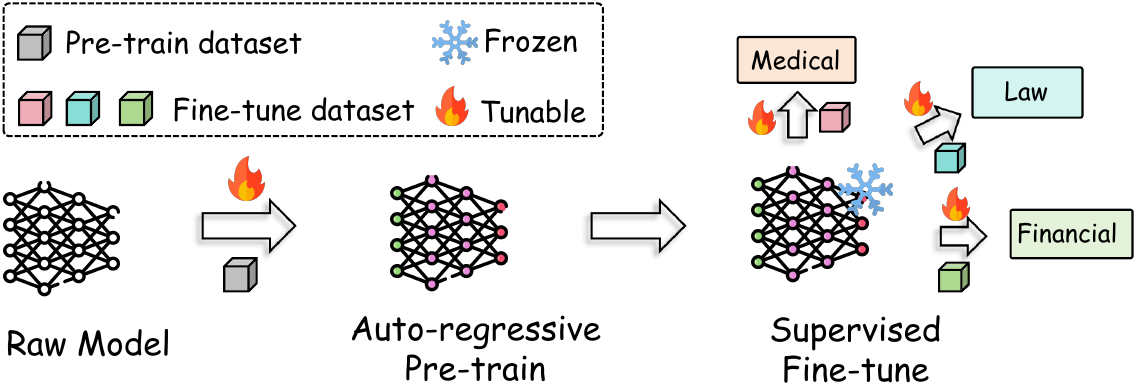}
    \caption{Schematic illustration of the two-stage LLM training process: 1) auto-regressive pre-training on large-scale corpora to develop general linguistic capabilities, followed by 2) supervised fine-tuning to align model outputs with specific task requirements or human preferences.}
    \label{training_of_LLM}
\end{figure}

The training of LLMs encompasses two distinct stages~\citep{xin2024parameter}: pre-training and fine-tuning, as illustrated in Figure~\ref{training_of_LLM}.
1) \textbf{Pre-training} involves training the model on massive unlabeled text corpora (billions to trillions of tokens) drawn from diverse sources such as academic papers, websites, and books. This stage generally adopts the auto-regressive modeling~\citep{yang2019xlnet} approach that predicts each token based on its previous context. 
Through extensive training, LLMs develop a wide range of capabilities, from basic semantic understanding to advanced reasoning across diverse domains. 
While computationally intensive, this unsupervised learning process builds robust and transferable representations that serve as a powerful foundation for various downstream tasks~\citep{naveed2023comprehensive}.

2) The function of \textbf{fine-tuning} is to adapt the pre-trained model to specific downstream tasks through additional training on task-specific datasets~\citep{ding2023parameter}. This stage typically utilizes supervised learning to optimize the model's performance for particular applications. While fine-tuning is highly effective at specializing the model’s general language understanding for targeted tasks, traditional fine-tuning methods often require centralizing data from various sources on a central server~\citep{huang2025survey}, which raises significant privacy and security concerns.
These concerns have sparked growing interest in privacy-preserving fine-tuning paradigms, which seek to retain the benefits of model specialization while ensuring that sensitive user data remains decentralized and secure throughout the training process.

\subsection{Federated Fine-Tuning}

Federated fine-tuning~\citep{zhang2024towards, yi2025fedfld} has emerged as a promising paradigm for adapting LLMs to specific downstream tasks while preserving data privacy. Unlike conventional centralized approaches that require sensitive data aggregation at a central server, federated fine-tuning enables distributed clients to adapt LLMs on local private datasets, sharing only model updates with the coordinating server.
This privacy-preserving approach aligns well with modern data protection requirements and user expectations.
However, the massive scale of LLM parameters and heterogeneous data distributions across clients introduce significant technical challenges. These challenges encompass prohibitive communication bandwidth requirements for transmitting model updates, convergence difficulties when training across heterogeneous data distributions, excessive memory demands that strain client-side resources, and intensive computational overhead that impacts efficiency and energy consumption.
To address these challenges, researchers have proposed various parameter-efficient federated fine-tuning approaches, each strategically designed to mitigate resource constraints while maintaining model performance on downstream tasks. These innovative methods can be broadly categorized as follows:

\begin{itemize}
    \item  1) LoRA-based Tuning~\citep{hu2021lora}: This methodology leverages the intrinsic low-rank nature of weight updates by decomposing them into low-rank approximation matrices, significantly reducing trainable parameters while preserving model's expressiveness.

    \item 2) Prompt-based Tuning~\citep{lester2021power}:  This approach optimizes continuous or discrete prompts in the input space to steer the model's behavior toward specific tasks. By modifying only the prompt embeddings while keeping model weight frozen, it achieves remarkable parameter efficiency in task adaptation.

    \item 3) Adapter-based Tuning~\citep{houlsby2019parameter}: This strategy incorporates specialized adapter modules between the layers of the pre-trained model. By updating only these compact adapters while freezing the original model parameters, it enables efficient task-specific adaptation with minimal architectural modifications to the base model.

    \item 4) Selective-based Tuning~\citep{zaken2021bitfit}: This approach focuses on selectively fine-tuning specific layers or parameters of the model that are most relevant to the downstream task. Through careful selection, it significantly reduces the resource consumption.
    
    \item 5) Other Tuning Methods~\citep{li2024synergizing}: This category encompasses techniques like zeroth-order optimization~\citep{malladi2023fine}, split learning~\citep{thapa2022splitfed}, model compression~\citep{deng2020model}, and data selection~\citep{qin2024federated}, which offer innovative ways to optimize LLM performance with lower resource requirements.

\end{itemize}


\section{Challenges}\label{sec_challenge}

In this section, we provide an in-depth analysis of the challenges encountered in FedLLM, focusing on four key aspects: communication overhead, data heterogeneity, memory constraints, and computation burden.

\subsection{Communication Overhead}

\begin{wrapfigure}{r}{0.4\linewidth} 
\centering
\vspace{-1mm}
\adjustbox{max width=\linewidth}{\includegraphics{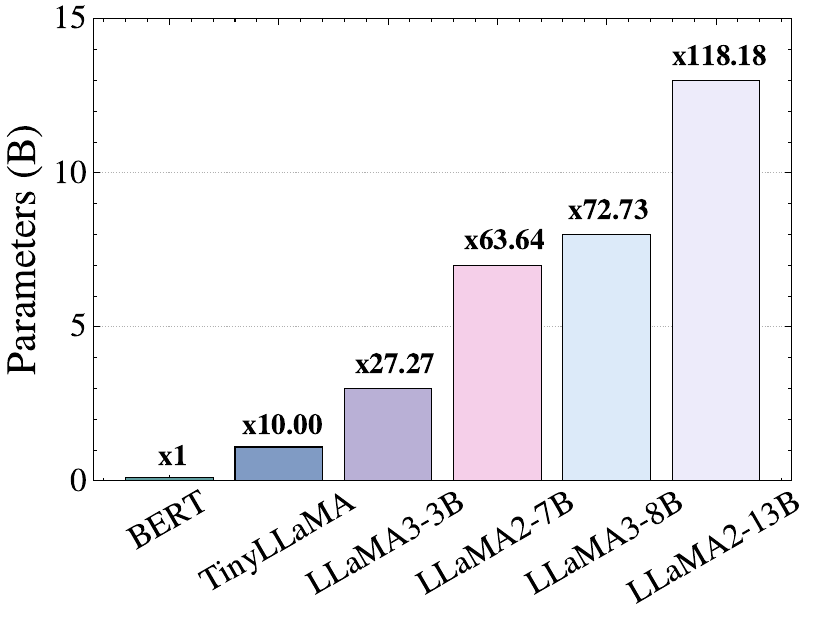}}
\vspace{-5mm}
\caption{Comparison of model parameters across BERT and LLaMA series models.}
\label{Model_parameters}
\end{wrapfigure}

In federated fine-tuning, the learning process necessitates iterative communication between participating clients and the central server, where clients periodically transmit their locally updated model parameters for aggregation~\citep{fu2022federated, fu2024federated, fu2024federated2}.
This iterative exchange continues until model convergence, inherently introducing substantial communication overhead~\citep{li2022one, kou2025fast}. 
The challenge is even more pronounced when fine-tuning LLMs, which consist of billions of parameters.
To quantify this challenge, Figure~\ref{Model_parameters} presents a comparative analysis of parameter sizes across different models, contrasting traditional models like BERT~\citep{BERT} with the LLaMA series, including TinyLLaMA~\citep{zhang2024tinyllama}, LLaMA2-7B, LLaMA2-13B~\citep{llama2}, LLaMA3-3B, and LLaMA3-8B~\citep{dubey2024llama}. Our analysis reveals that LLaMA models are dramatically larger than BERT, with parameter counts ranging from 10 to 118$\times$ greater.
This exponential increase in parameter size directly translates to significantly higher data transmission volumes in each communication round, substantially elevating bandwidth requirements and overall communication costs in federated  environments.

However, in real-world scenarios, communication bandwidth is often severely constrained. According to a 2023 Cisco report, approximately 30\% of edge devices still rely on 2G or 3G networks, which provide bandwidths of less than 10 Mb/s~\citep{wang2023bose}. While 5G networks offer speeds more than 50$\times$ faster, they are accessible to only about 10\% of devices. This significant disparity in network capabilities inevitably results in substantial communication delays, particularly when exchanging large parameter updates.
More critically, the duration of each training round is determined by the slowest device in the network—a phenomenon known as the “straggler effect.” This means that devices with limited connectivity can dramatically hinder the convergence speed of the federated fine-tuning process. Consequently, minimizing communication overhead becomes essential for effective FedLLM implementation.
Efficient management of data transmission can accelerate model convergence while ensuring the practical feasibility of deploying FedLLM in bandwidth-constrained environments. Without addressing the communication challenge, the theoretical privacy benefits of federated fine-tuning may remain inaccessible to many real-world applications, particularly those involving edge devices or regions with limited network infrastructure.


\subsection{Data Heterogeneity}

\begin{wrapfigure}{r}{0.4\linewidth} 
\centering
\vspace{-5mm}
\adjustbox{max width=\linewidth}{\includegraphics{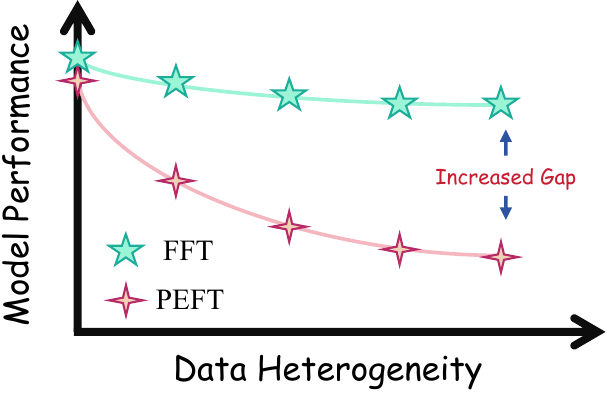}}
\vspace{-5mm}
\caption{Impact of data heterogeneity on model performance. As the degree of data heterogeneity increases, the performance gap between PEFT and FFT widens.}
\label{Fig_data}
\end{wrapfigure}

Data heterogeneity is a notorious challenge in FL, manifesting in significant variations in data distribution~\citep{tian2022harmony}, quality~\citep{tam2023federated, tam2023fedcoop}, and quantity~\citep{yi2022robust} across clients. Such heterogeneity hinders convergence and degrades the global model’s generalization ability, as it must reconcile conflicting updates derived from diverse client populations~\citep{ma2024fedmg, wang2025indoor}.
To mitigate the adverse effects of data heterogeneity, various strategies have been explored in traditional FL, which can be broadly categorized into four groups: 
1) \textbf{Regularization-based methods} incorporate additional penalty terms into the local objective to limit model divergence and encourage alignment with the global model~\citep{li2020federated};
2) \textbf{Aggregation-based methods} modify the server-side aggregation strategy to assign adaptive weights to client updates, reducing the influence of noisy, unreliable, or biased data sources~\citep{wang2020federated, wang2020tackling};
3) \textbf{Data-sharing methods} introduce small, carefully curated auxiliary datasets that are distributed to clients to promote distributional alignment and reduce inter-client drift~\citep{goetz2020federated};
4) \textbf{Personalized FL approaches} aim to address data heterogeneity by learning models that capture the unique characteristics of local data. Rather than converging to a single global model, these methods may produce multiple global models that are individually personalized for each client~\citep{kou2024pfedlvm, sabah2024model}.
While these techniques have demonstrated success in traditional FL scenarios, addressing data heterogeneity in FedLLM remains largely underexplored. This challenge is further exacerbated when applying PEFT techniques, as PEFT tends to be more sensitive to distributional shifts and limited data availability. Figure~\ref{Fig_data} shows that the performance gap between PEFT and full-parameter fine-tuning (FFT) grows wider as data heterogeneity increases, underscoring the need for targeted solutions to improve PEFT robustness in federated settings.



\subsection{Memory Wall}

Memory constraints present a fundamental challenge to the practical deployment of federated fine-tuning~\citep{wu2025breaking}. During the local fine-tuning process, model parameters, intermediate activations, and gradients must be stored in memory, resulting in substantial memory consumption. However, participating clients, especially edge devices, typically have limited available memory, ranging from 4 to 12 GB~\citep{tian2024breaking,tam2024fedhybrid}. This limited memory capacity is insufficient to support fine-tuning mainstream LLMs.

\begin{wrapfigure}{r}{0.4\linewidth} 
\centering
\vspace{-4mm}
\adjustbox{max width=\linewidth}{\includegraphics{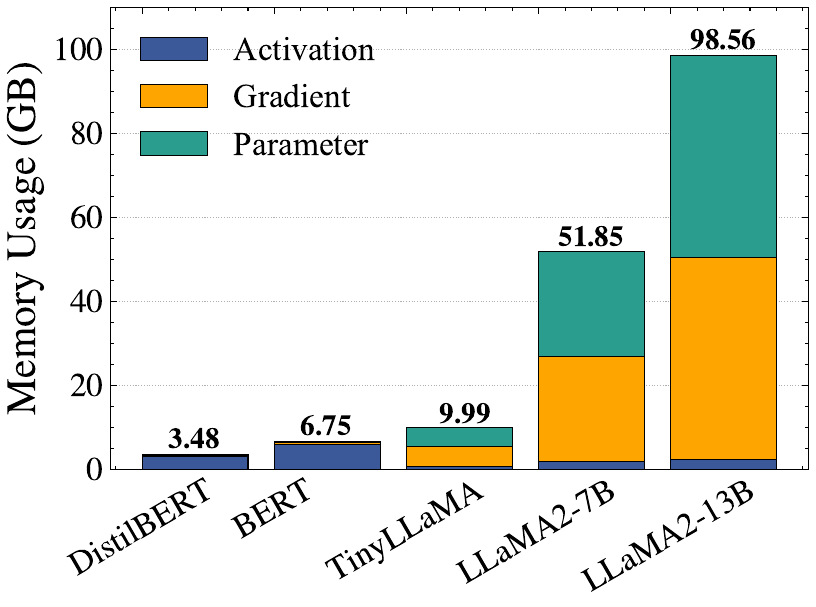}}
\vspace{-5mm}
\caption{Memory usage breakdown for fine-tuning various models. The analysis is conducted on an NVIDIA H800 with precisions of FP32 and FP16, a batch size of 16, and a maximum sequence length of 512.}
\label{Fig_memory}
\vspace{-3mm}
\end{wrapfigure}

To better quantify this challenge, we profile the memory usage during full-parameter fine-tuning for both traditional models (e.g., DistilBERT~\citep{sanh2019distilbert}, BERT~\citep{BERT}) and LLaMA-series models (e.g., TinyLLaMA, LLaMA2-7B, LLaMA2-13B). As shown in Figure~\ref{Fig_memory}, our results reveal a dramatic disparity in resource requirements between these model families.
Fine-tuning LLaMA models demands substantially higher memory resources compared to traditional architectures. Specifically, fine-tuning LLaMA2-7B requires approximately 51.85 GB of GPU memory, which is 7.68$\times$ more than BERT (6.75 GB).
This requirement escalates further with LLaMA2-13B, which demands 98.56 GB of memory, representing a 28.32$\times$ increase over DistilBERT and vastly exceeding the available memory capacity of  edge devices.
This stark mismatch between the memory demands of fine-tuning LLMs and the hardware limitations of participants creates a \textbf{Memory Wall}, a fundamental barrier that severely restricts the feasibility of deploying FedLLM at scale. This memory constraint prevents a significant proportion of devices from participating in the collaborative learning process, thereby compromising model performance through reduced data diversity and limiting the practical application scope of FedLLM.
The memory wall represents not just a technical challenge but a fundamental constraint on democratizing access to advanced AI capabilities through FL.

\subsection{Computation Overhead}

\begin{wrapfigure}{r}{0.4\linewidth} 
\centering
\vspace{-5mm}
\adjustbox{max width=\linewidth}{\includegraphics{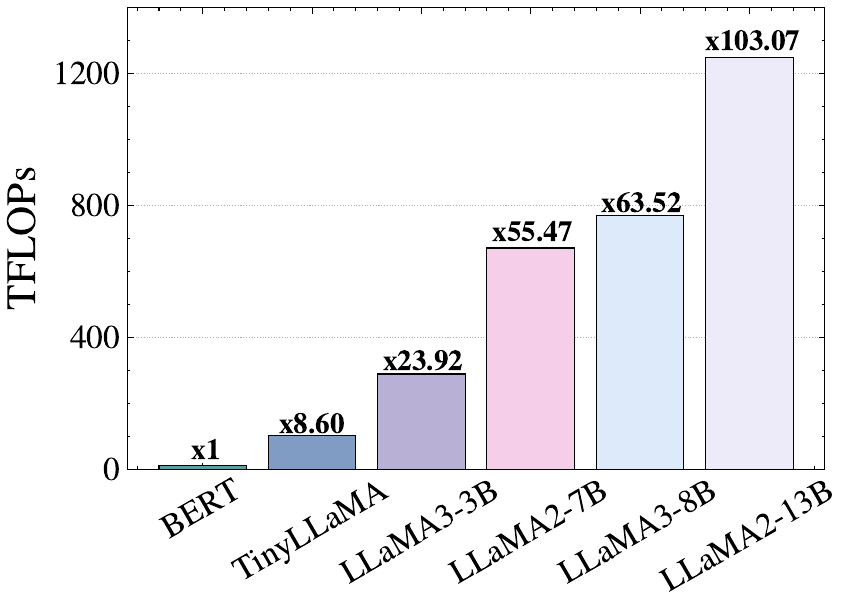}}
\vspace{-5mm}
\caption{Comparison of FLOPs for a single forward and backward pass across models.}
\label{Fig_computation}
\end{wrapfigure}

Computational cost presents another major bottleneck in deploying FedLLM~\citep{tian2022harmony, almanifi2023communication}. The computational demands of fine-tuning LLMs arise from the forward and backward passes during local training iterations, each contributing significantly to the overall processing burden. 
The sheer scale of these models—characterized by billions of parameters, numerous transformer layers, and complex attention mechanisms—makes them inherently compute-intensive, which can quickly overwhelm devices with limited processing capabilities. Moreover, the iterative nature of fine-tuning, which involves repeated forward and backward passes, compounds the computational load, making it difficult to achieve efficient training on resource-constrained devices.
To quantitatively understand this challenge, we profile the computational demands of fine-tuning various models by measuring the floating-point operations (FLOPs) required for a single forward and backward pass with a batch size of 16.

Specifically, we evaluate BERT alongside a suite of LLaMA-based models, including TinyLLaMA, LLaMA3-3B, LLaMA2-7B, LLaMA3-8B, and LLaMA2-13B. As shown in Figure~\ref{Fig_computation}, our results reveal a dramatic escalation in computational requirements for LLaMA-series models compared to BERT. For instance, fine-tuning TinyLLaMA incurs 8.60$\times$ FLOPs of BERT, while LLaMA2-13B demands a staggering 103.07$\times$ more FLOPs. 
This exponential increase in computational complexity directly results in significantly longer training time, excessive energy consumption on battery-powered devices, and thermal management issues, all of which can degrade hardware performance over time, thereby undermining the feasibility of large-scale, real-world deployment~\citep{ning2024fedgcs, DBLP:conf/iclr/TianSL23}.
These findings highlight the pressing need for computation-efficient fine-tuning strategies that can effectively accommodate the heterogeneous and resource-constrained nature of participating devices, while ensuring model performance. 


\section{Federated Fine-Tuning}\label{sec_federated_finetuning}

\begin{figure}[!h]
    \centering
    \includegraphics[width=0.95\linewidth]{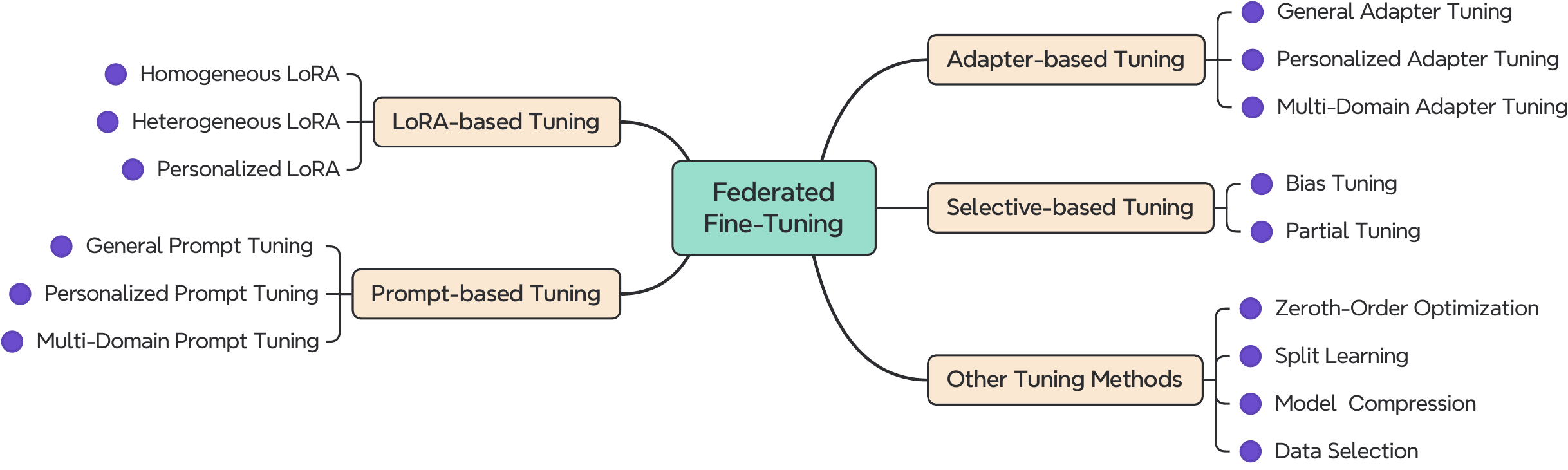}
    \vspace{-2mm}
    \caption{Overview of parameter-efficient federated fine-tuning methods and their corresponding taxonomy.}
    \label{main_table}
    \vspace{-4mm}
\end{figure}

In this section, we introduce various parameter-efficient fine-tuning methods and discuss their applications in FL. Figure~\ref{main_table} provides an overview of representative parameter-efficient federated fine-tuning methods.

\subsection{LoRA-based Tuning}\label{sec_lora}

\subsubsection{Preliminary}

\begin{wrapfigure}{r}{0.4\linewidth} 
\centering
\vspace{-5mm}
\adjustbox{max width=\linewidth}{\includegraphics{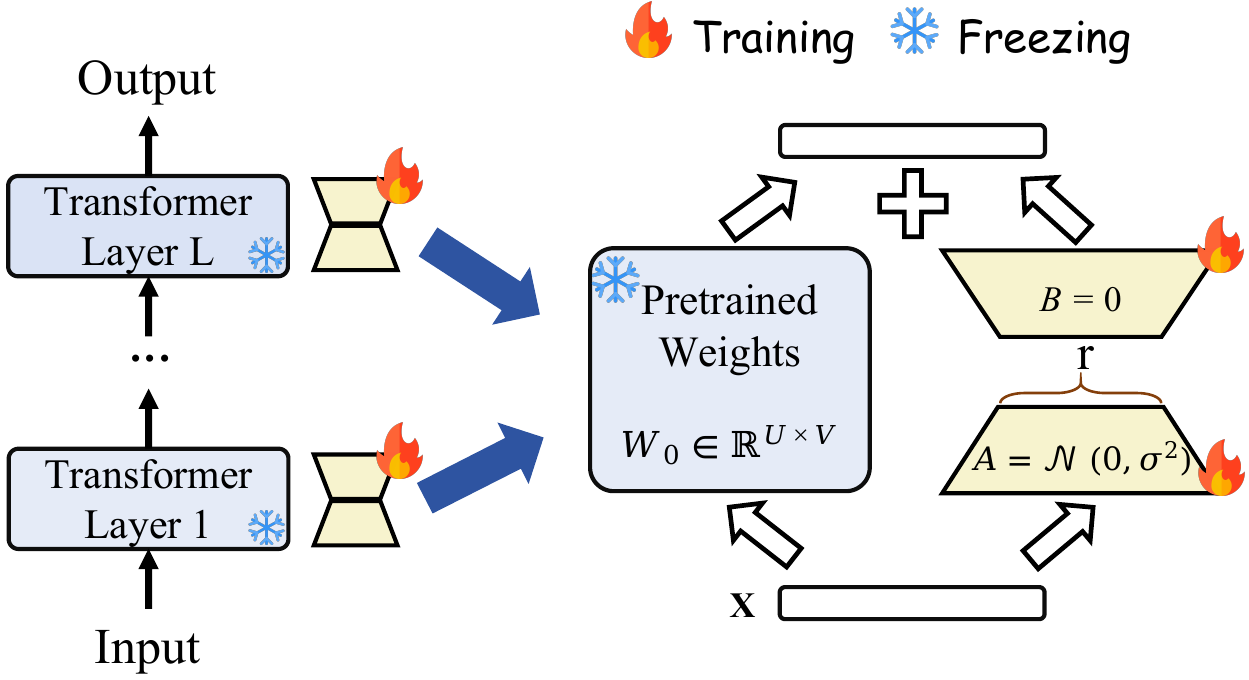}}
\vspace{-5mm}
\caption{The working principle of LoRA.}
\label{Fig_LoRA}
\vspace{-2mm}
\end{wrapfigure}

Low-Rank Adaptation (LoRA)~\citep{hu2021lora, tian2024hydralora} has emerged as a promising approach for efficient fine-tuning of LLMs while maintaining model performance. The core idea of LoRA lies in introducing low-rank matrices into the pre-trained model's weights, allowing for the adaptation of model parameters without altering the original architecture significantly. 
LoRA is based on the observation that fine-tuning does not require updating the full parameter space; instead, meaningful adaptations can often be represented in a low-dimensional subspace. By applying low-rank decomposition to the weight updates, LoRA drastically reduces the number of trainable parameters, leading to lower resource consumption. Furthermore, LoRA's modular design allows it to be easily integrated into a variety of model architectures without altering the original model.

Figure~\ref{Fig_LoRA} illustrates the working principle of LoRA. Specifically, the pre-trained parameter matrix \( \mathbf{W_0} \in \mathbb{R}^{U \times V} \) is decomposed into two matrices, \( \mathbf{A} \in \mathbb{R}^{r \times V} \) and \( \mathbf{B} \in \mathbb{R}^{U \times r} \), where \( r \ll \min(U, V) \) denotes the rank controlling the dimensionality of the low-rank subspace.
The matrix \( \mathbf{A} \) projects the input into a low-dimensional space, and the matrix \( \mathbf{B} \) maps it back to the original space. During fine-tuning, only \( \mathbf{A} \) and \( \mathbf{B} \) are updated, while the original weights \( \mathbf{W}_0 \) remain frozen.
The input data \( X \) is processed by both \( \mathbf{W_0} \) and \( \mathbf{BA} \). The output of \( \mathbf{W_0} X \) is the initial prediction generated by the pre-trained model, while the output of \( \mathbf{BA} X \) represents the task-specific adaptation introduced by the low-rank matrices. These two outputs are then added element-wise to produce the final output. This process can be formulated as:
\begin{equation}
    h = \mathbf{W_0} X + \mathbf{BA} X
\end{equation}
By updating only \( \mathbf{A} \) and \( \mathbf{B} \), LoRA enables efficient task adaptation with minimal resource overhead, effectively capturing task-specific knowledge while preserving the generalization ability of the pre-trained model.

\subsubsection{LoRA in Federated Fine-Tuning}

In the context of federated fine-tuning, LoRA offers notable advantages in both communication and computation efficiency. Since only the low-rank matrices are updated and transmitted, rather than the full model parameters, this lightweight updating mechanism significantly lowers bandwidth requirements and facilitates faster convergence of the global model, making LoRA particularly well-suited for federated environments. In this paper, we introduce a novel taxonomy of LoRA-based federated fine-tuning methods, categorizing them into three primary types: Homogeneous LoRA, where all clients adopt the same rank; Heterogeneous LoRA, where clients use different ranks based on their resources; and Personalized LoRA, which tailors low-rank adaptations to individual client data distributions. This taxonomy is summarized in Figure~\ref{main_table}. In the following sections, we delve into representative methods within each category, analyzing how they address the key challenges identified in Section~\ref{sec_challenge}, beyond the inherent benefits brought by LoRA itself.

\noindent $\bullet$ \textbf{Homogeneous LoRA} refers to scenarios where all clients adopt the same low-rank dimension \( r \) for their LoRA modules. This uniform configuration simplifies aggregation and model synchronization across clients. Table~\ref{LoRA_Homo_table} summarizes representative methods in this category and the specific challenges they address.

\textbf{FedIT}~\citep{zhang2024towards} directly integrates LoRA into the classic FedAvg~\citep{mcmahan2017communication} for instruction tuning. 
\textbf{FedSA-LoRA}~\citep{guo2024selective} identifies that the \( \mathbf{A} \) matrices primarily encode general knowledge, while the \( \mathbf{B} \) matrices capture client-specific features; thus, it only uploads \( \mathbf{A} \) to the server, significantly reducing communication overhead. \textbf{FederatedScope-LLM}~\citep{fsllm} establishes a comprehensive end-to-end pipeline for federated LLM fine-tuning and proposes offsite-tuning strategies to mitigate both communication and computational costs. \textbf{FeDeRA}~\citep{yan2024federa} addresses data heterogeneity by initializing LoRA matrices via singular value decomposition on the pre-trained weights. \textbf{LoRA-FAIR}~\citep{bian2024lora} introduces a correction mechanism on the server to handle aggregation bias and initialization drift across clients.
\textbf{FLASC}~\citep{kuo2024federated} incorporates sparsity into LoRA to further reduce communication overhead.
\textbf{SA-FedLoRA}~\citep{yang2024sa} mitigates client drift through parameter regularization and dynamically allocates communication budgets.

\textbf{SLoRA}~\citep{babakniya2023slora} proposes a novel data-driven initialization scheme to better handle statistical heterogeneity.
\textbf{RoLoRA}~\citep{chen2024robust} adopts an alternating minimization approach to improve robustness under non-IID conditions. \textbf{FedPipe}~\citep{fang2024automated} automatically selects critical parameters for fine-tuning and applies quantization to reduce memory usage.
\textbf{LP-FL}~\citep{lpfl} applies LoRA directly to enable efficient on-device fine-tuning.
\textbf{Fed-piLot}~\citep{zhang2024fed} reduces memory consumption through LoRA assignment strategies and introduces a novel spatial-temporal aggregation (STAgg) rule to address heterogeneity.
\textbf{FedRA}~\citep{su2023fedra} adaptively determines parameter update scopes based on client resource constraints, effectively reducing computational, communication, and memory costs.
\textbf{FedPruner}~\citep{wu2025memory} introduces a macro-micro synergetic pruning framework to mitigate memory constraints on participating devices. To further facilitate large-scale deployment, \textbf{DevFT}~\citep{wu2025learning} proposes a developmental federated tuning paradigm to minimize training resource overheads.

\begin{table}[!t]
\small
\centering
\caption{\textbf{Homogeneous LoRA in federated fine-tuning.}}
\label{LoRA_Homo_table}
\resizebox{0.8\linewidth}{!}{
\begin{tabular}{lccccccc}
\toprule[1pt]
\multirow{2}{*}{\textbf{Method}}& \multicolumn{4}{c}{\textbf{Challenge}}  \\  

\cmidrule{2-5}

& \textbf{Communication} & \textbf{Non-IID} & \textbf{Memory} & \textbf{Computation}   \\ 

\midrule[1pt]
FedIT~\citep{zhang2024towards}&\hlr{\xmark}&\hlr{\xmark}&\hlr{\xmark}&\hlr{\xmark}\\
FedSA-LoRA~\citep{guo2024selective}&\hlg{\tmark{}}&\hlr{\xmark}&\hlr{\xmark}&\hlr{\xmark}\\
FederatedScope-LLM~\citep{fsllm}&\hlg{\tmark{}}&\hlr{\xmark}&\hlr{\xmark}&\hlg{\tmark{}}\\
{FeDeRA~\citep{yan2024federa}}&\hlr{\xmark}&\hlg{\tmark{}}& \hlr{\xmark} &\hlr{\xmark}\\ 
{LoRA-FAIR~\citep{bian2024lora}}& \hlr{\xmark}& \hlr{\xmark} & \hlr{\xmark}& \hlr{\xmark}\\ {FLASC~\citep{kuo2024federated}}&\hlg{\tmark{}}&\hlr{\xmark}&\hlr{\xmark}&\hlr{\xmark}\\
{SA-FedLoRA~\citep{yang2024sa}}&\hlg{\tmark{}}&\hlg{\tmark{}}&\hlr{\xmark}&\hlr{\xmark}\\
{SLoRA~\citep{babakniya2023slora}}&\hlr{\xmark}&\hlg{\tmark{}}& \hlr{\xmark} &\hlr{\xmark}\\ {RoLoRA~\citep{chen2024robust}}&\hlr{\xmark}&\hlg{\tmark{}}& \hlr{\xmark} &\hlr{\xmark} \\
{FedPipe~\citep{fang2024automated}}&\hlg{\tmark{}}&\hlr{\xmark}&\hlg{\tmark{}}&\hlg{\tmark{}}\\ 
{Lp-FL~\citep{lpfl}}&\hlr{\xmark}&\hlr{\xmark}&\hlr{\xmark}&\hlr{\xmark}\\ 
{Fed-piLot~\citep{zhang2024fed}}&\hlr{\xmark}&\hlg{\tmark{}}&\hlg{\tmark{}}&\hlr{\xmark}\\ 
FedRA~\citep{su2023fedra}&\hlg{\tmark{}}&\hlr{\xmark}&\hlg{\tmark{}}&\hlg{\tmark{}}\\
FedPruner~\citep{wu2025memory}&\hlg{\tmark{}}&\hlg{\tmark{}}&\hlg{\tmark{}}&\hlg{\tmark{}}\\
DevFT~\citep{wu2025learning}&\hlg{\tmark{}}&\hlr{\xmark}&\hlg{\tmark{}}&\hlg{\tmark{}}\\

\bottomrule[1pt]
\end{tabular}
}
\vspace{-5mm}
\end{table}
\begin{table}[!t]
\small
\centering
\caption{\textbf{Heterogeneous LoRA in federated fine-tuning.}}
\label{LoRA_Hete_table}
\resizebox{0.8\linewidth}{!}{
\begin{tabular}{lccccccc}
\toprule[1pt]
\multirow{2}{*}{\textbf{Method}}& \multicolumn{4}{c}{\textbf{Challenge}}  \\  

\cmidrule{2-5}

& \textbf{Communication} & \textbf{Non-IID} & \textbf{Memory} & \textbf{Computation}  \\ 

\midrule[1pt]
{HETLoRA~\citep{cho2024heterogeneous}}&\hlg{\tmark{}}&\hlg{\tmark{}}&\hlg{\tmark{}}&\hlg{\tmark{}}\\ 
{FLoRA~\citep{wang2024flora}}&\hlg{\tmark{}}&\hlr{\xmark}&\hlg{\tmark{}}&\hlg{\tmark{}}\\
{FlexLoRA~\citep{bai2024federated}}&\hlg{\tmark{}}&\hlr{\xmark}&\hlg{\tmark{}}&\hlg{\tmark{}}\\ 
{LoRA-A$^{2}$~\citep{koo2024towards}}&\hlg{\tmark{}}&\hlg{\tmark{}}&\hlg{\tmark{}}&\hlg{\tmark{}}\\ 
\citet{byun2024towards}&\hlg{\tmark{}}&\hlr{\xmark}&\hlg{\tmark{}}&\hlg{\tmark{}}\\
{FedHM~\citep{yao2021fedhm}}&\hlg{\tmark{}}&\hlr{\xmark}&\hlg{\tmark{}}&\hlg{\tmark{}}\\
{RBLA~\citep{tavallaie2024rbla}}&\hlg{\tmark{}}&\hlg{\tmark{}}&\hlg{\tmark{}}&\hlg{\tmark{}}\\
{SmartFed~\citep{wu2025elastic}}&\hlg{\tmark{}}&\hlr{\xmark{}}&\hlg{\tmark{}}&\hlg{\tmark{}}\\
                    
\bottomrule[1pt]
\end{tabular}
}
\vspace{-4mm}
\end{table}

\noindent $\bullet$ \textbf{Heterogeneous LoRA} allows clients to adopt different rank values \( r \) based on their data characteristics or resource constraints. This heterogeneity can manifest either across clients (inter-model) or within different layers of the same model (intra-model). By enabling each client to select a rank that best fits its capabilities and local data, this approach introduces greater flexibility and resource-awareness into the federated fine-tuning process. Table~\ref{LoRA_Hete_table} summarizes representative methods and the specific challenges they address.

\textbf{HETLoRA}~\citep{cho2024heterogeneous} assigns heterogeneous ranks across devices and incorporates rank self-pruning along with sparsity-weighted aggregation to tackle data heterogeneity.
\textbf{FLoRA}~\citep{wang2024flora} proposes a stacking-based aggregation scheme and allows devices to select ranks according to their resource budgets.
\textbf{FlexLoRA}~\citep{bai2024federated} enables dynamic adjustment of local LoRA ranks to leverage the heterogeneous device resources, while employing singular value decomposition for weight redistribution. 
\textbf{LoRA-A$^{2}$}~\citep{koo2024towards} introduces alternating freezing and adaptive rank selection mechanisms to fully utilize heterogeneous device resources while addressing statistical heterogeneity.
\citet{byun2024towards} propose a replication-based padding technique to enable aggregation across clients with varying LoRA ranks.
\textbf{FEDHM}~\citep{yao2021fedhm} addresses resource constraints by distributing low-rank models with heterogeneous capacities to clients.
\textbf{RBLA}~\citep{tavallaie2024rbla} improves aggregation robustness by simultaneously maintaining and aligning both low-rank and high-rank feature components.
\textbf{SmartFed}~\citep{wu2025elastic} achieves efficient task adaptation by leveraging rank-wise reconfiguration of existing LoRA modules, thereby avoiding the prohibitive computational overhead of training from scratch.

\begin{table}[!t]
\small
\centering
\caption{\textbf{Personalized LoRA in federated fine-tuning.}}
\label{LoRA_Personalize_table}
\resizebox{0.8\linewidth}{!}{
\begin{tabular}{lccccccc}
\toprule[1pt]
\multirow{2}{*}{\textbf{Method}}& \multicolumn{4}{c}{\textbf{Challenge}}  \\  

\cmidrule{2-5}

& \textbf{Communication} & \textbf{Non-IID} & \textbf{Memory} & \textbf{Computation}  \\ 

\midrule[1pt]
{FDLoRA~\citep{qi2024fdlora}}&\hlr{\xmark}&\hlg{\tmark{}}&\hlr{\xmark}&\hlr{\xmark}\\ 
{pFedLoRA~\citep{yi2023fedlora}}&\hlr{\xmark}&\hlg{\tmark{}}&\hlr{\xmark}&\hlr{\xmark}\\  {FedLoRA~\citep{wu2024fedlora}}&\hlr{\xmark}&\hlg{\tmark{}}&\hlr{\xmark}&\hlr{\xmark}\\ {FedDPA~\citep{yang2024dual}}&\hlr{\xmark}&\hlg{\tmark{}}&\hlr{\xmark}&\hlr{\xmark}\\ 
{PerFIT~\citep{zhang2024personalized}}&\hlr{\xmark}&\hlg{\tmark{}}&\hlr{\xmark}&\hlr{\xmark}\\ {FedMEM~\citep{du2024communication}}&\hlr{\xmark}&\hlg{\tmark{}}&\hlr{\xmark}&\hlr{\xmark}\\ 
{FedAMoLE~\citep{zhang2024personalized_new}}&\hlr{\xmark}&\hlg{\tmark{}}&\hlr{\xmark}&\hlr{\xmark}\\               
\bottomrule[1pt]
\end{tabular}
}
\vspace{-4mm}
\end{table}

\noindent $\bullet$ \textbf{Personalized LoRA} enables each participant to fine-tune its model using personalized low-rank adaptation matrices, allowing for better alignment with local data characteristics. This approach enhances the ability of the global model to generalize across clients while retaining client-specific nuances. Table~\ref{LoRA_Personalize_table} summarizes representative methods and the specific challenges they aim to address.

\textbf{FDLoRA}~\citep{qi2024fdlora} introduces dual LoRA modules on each client to separately capture global and personalized knowledge.
\textbf{pFedLoRA}~\citep{yi2023fedlora} designs a homogeneous small adapter to facilitate federated clients' heterogeneous local model training, with a proposed iterative training process for global-local knowledge exchange. 
\textbf{FedLoRA}~\citep{wu2024fedlora} maintains shared general knowledge in a global full-rank matrix while encoding client-specific knowledge in a personalized low-rank module.
\textbf{FedDPA}~\citep{yang2024dual} utilizes a global adapter and a local adapter to jointly address test-time distribution shifts and client-specific personalization. \textbf{PerFIT}~\citep{zhang2024personalized} allows each client to search for a personalized architecture by expanding the trainable parameter space of the global model to address data heterogeneity. \textbf{FEDMEM}~\citep{du2024communication} equips the global model with a KNN classifier that captures client-specific distributional shifts, achieving personalization and overcoming data heterogeneity. \textbf{FedAMoLE}~\citep{zhang2024personalized_new} features a mixture of LoRA experts module for aggregating heterogeneous models and a reverse selection-based expert assignment strategy that optimizes model architectures based on data distributions.

\subsection{Prompt-based Tuning}\label{sec_prompt}

\subsubsection{Preliminary}

\begin{wrapfigure}{r}{0.3\linewidth} 
\centering
\vspace{-4mm}
\adjustbox{max width=\linewidth}{\includegraphics{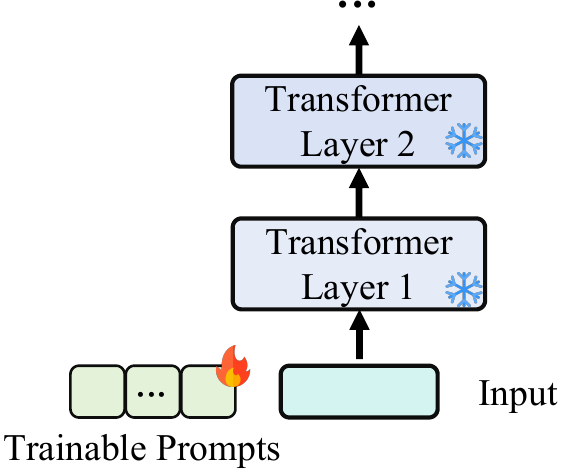}}
\vspace{-5mm}
\caption{The working principle of prompt tuning.}
\label{Fig_Prompt}
\vspace{-4mm}
\end{wrapfigure}
Prompt-based tuning~\citep{lester2021power} has emerged as a highly effective and resource-efficient alternative to conventional fine-tuning approaches for LLMs. Unlike traditional methods that update the model’s parameters directly, prompt-based tuning learns a set of trainable prompts or input embeddings that steer the model’s behavior on downstream tasks. By modifying only the input space, this approach leverages the pre-trained knowledge of LLMs without altering their weights.
As illustrated in Figure~\ref{Fig_Prompt}, a sequence of trainable prompt embeddings \( P \in \mathbb{R}^{l_p \times d} \) is prepended to the original input tokens \( X \in \mathbb{R}^{l_x \times d} \), where \( l_p \) and \( l_x \) denote the lengths of the prompt and input sequences, respectively, and \( d \) represents the model’s hidden dimension. The concatenated sequence is then fed into the frozen model: $Z = f([P; X]; \theta)$, where \( f(\cdot; \theta) \) denotes the pre-trained LLM with frozen parameters \( \theta \), and \( [P; X] \) represents the concatenation of the trainable prompts and the original input tokens. By optimizing the prompt embeddings \( P \), the model can effectively adapt to new tasks while reusing its pre-trained knowledge. This approach enables efficient task adaptation without modifying the model weights, thereby significantly reducing memory and computational overhead. Through prompt-based tuning, task-specific guidance is embedded within the prompts, allowing the model to generate desired outputs by attending to relevant information stored in its pre-trained parameters.

\subsubsection{Prompt in Federated Fine-Tuning}

In the context of federated fine-tuning, prompt-based tuning offers significant advantages in communication efficiency and model adaptability. Since only the trainable prompt embeddings are updated and exchanged, rather than model parameters, this approach substantially reduces communication overhead between clients and the central server. Additionally, by freezing the base model, prompt-based tuning allows clients with heterogeneous data distributions to personalize their behavior effectively, while still benefiting from globally shared knowledge encoded in the pre-trained model. These properties make prompt-based tuning a compelling choice for federated fine-tuning. In this paper, we propose a novel taxonomy of prompt-based federated fine-tuning approaches, categorizing them into three primary types: General Prompt Tuning, Personalized Prompt Tuning, and Multi-Domain Prompt Tuning, as illustrated in Figure~\ref{main_table}. In the following sections, we examine representative methods within each category and analyze how they address the challenges outlined in Section~\ref{sec_challenge}, beyond the inherent benefits brought by prompt tuning itself.

\noindent $\bullet$ \textbf{General Prompt Tuning} refers to approaches in which a shared set of prompt embeddings is learned and applied uniformly across all participating clients. In this setting, the same prompts are prepended to each client’s input sequences, providing consistent task-specific guidance and enabling the global model to generalize across diverse data sources. Table~\ref{Prompt_General_table} summarizes representative methods and the specific challenges they aim to address.

\begin{table}[!t]
\small
\centering
\caption{\textbf{General prompt tuning in federated fine-tuning.}}
\label{Prompt_General_table}
\resizebox{0.8\linewidth}{!}{
\begin{tabular}{lccccccc}
\toprule[1pt]
\multirow{2}{*}{\textbf{Method}}& \multicolumn{4}{c}{\textbf{Challenge}}  \\  

\cmidrule{2-5}

& \textbf{Communication} & \textbf{Non-IID} & \textbf{Memory} & \textbf{Computation} \\ 

\midrule[1pt]

{MetePFL~\citep{chen2023prompt}}&\hlr{\xmark}&\hlr{\xmark}&\hlr{\xmark}&\hlr{\xmark}\\ 
{PromptFL~\citep{guo2023promptfl}}&\hlr{\xmark}&\hlr{\xmark}&\hlr{\xmark}&\hlr{\xmark}\\
{FedBPT~\citep{sun2023fedbpt}}&\hlr{\xmark}&\hlr{\xmark}&\hlg{\tmark{}}&\hlg{\tmark{}}\\
{FedPepTAO~\citep{che2023federated}}&\hlg{\tmark{}}&\hlg{\tmark{}}&\hlr{\xmark}&\hlr{\xmark}\\
{Fed-BBPT~\citep{lin2023efficient}}&\hlr{\xmark}&\hlr{\xmark}&\hlg{\tmark{}}&\hlg{\tmark{}}\\
{FedTPG~\citep{qiu2023text}}&\hlr{\xmark}&\hlr{\xmark}&\hlr{\xmark}&\hlr{\xmark}\\ 
{FedPR~\citep{feng2023learning}}&\hlr{\xmark}&\hlg{\tmark{}}&\hlr{\xmark}&\hlr{\xmark}\\
{Fed-CPrompt~\citep{bagwe2023fed}}&\hlr{\xmark}&\hlg{\tmark{}}&\hlr{\xmark}&\hlr{\xmark}\\
{Fedprompt~\citep{zhao2023fedprompt}}&\hlr{\xmark}&\hlr{\xmark}&\hlr{\xmark}&\hlr{\xmark}\\
{FedSP~\citep{dong2023tunable}}&\hlr{\xmark}&\hlr{\xmark}&\hlg{\tmark{}}&\hlg{\tmark{}}\\
{HePCo~\citep{halbe2023hepco}}&\hlr{\xmark}&\hlg{\tmark{}}&\hlr{\xmark}&\hlr{\xmark}\\
{PFL-GCN~\citep{ahmad2023prompt}}&\hlr{\xmark}&\hlr{\xmark}&\hlr{\xmark}&\hlr{\xmark}\\
{AUG-FedPrompt~\citep{cai2023towards}}&\hlr{\xmark}&\hlr{\xmark}&\hlr{\xmark}&\hlr{\xmark}\\
{\citet{liu2023federated}}&\hlr{\xmark}&\hlr{\xmark}&\hlr{\xmark}&\hlr{\xmark}\\
{FedHPL~\citep{ma2024fedhpl}}&\hlr{\xmark}&\hlg{\tmark{}}&\hlr{\xmark}&\hlr{\xmark}\\
{PFPT~\citep{wengprobabilistic}}&\hlr{\xmark}&\hlg{\tmark{}}&\hlr{\xmark}&\hlr{\xmark}\\
{FCILPT~\citep{liu2023federated1}}&\hlr{\xmark}&\hlg{\tmark{}}&\hlr{\xmark}&\hlr{\xmark}\\
{CaFPT~\citep{guo2024explore}}&\hlr{\xmark}&\hlr{\xmark}&\hlr{\xmark}&\hlr{\xmark}\\
{FedPoD~\citep{chen2023federated2}}&\hlr{\xmark}&\hlg{\tmark{}}&\hlr{\xmark}&\hlr{\xmark}\\

\bottomrule[1pt]
\end{tabular}
}
\vspace{-3mm}
\end{table}

\textbf{MetePFL}~\citep{chen2023prompt} applies prompt tuning to fine-tune a spatio-temporal Transformer-based foundation model for weather forecasting tasks in a federated setting.
\textbf{PromptFL}~\citep{guo2023promptfl} adapts CLIP models for vision-language tasks in FL using prompt-based tuning.
\textbf{FedBPT}~\citep{sun2023fedbpt} employs prompt-based tuning to efficiently adapt black-box LLMs using gradient-free optimization, eliminating the need for clients to access model parameters and requiring only forward propagation for local training. 
\textbf{FedPepTAO}~\citep{che2023federated} introduces a partial prompt tuning mechanism to reduce communication costs, along with an adaptive optimization algorithm to address data heterogeneity.
\textbf{Fed-BBPT}~\citep{lin2023efficient} enables clients to utilize a zeroth-order optimizer locally, obviating the need for full LLM deployment, effectively reducing memory consumption and computational costs. 
\textbf{FedTPG}~\citep{qiu2023text} learns a unified, task-aware prompt generation network conditioned on input text, improving generalization to both seen and unseen classes.

\textbf{FedPR}~\citep{feng2023learning} enhances federated visual prompt tuning by projecting local prompt updates into an approximate null space of the global prompt, mitigating gradient interference and improving global performance.
\textbf{Fed-CPrompt}~\citep{bagwe2023fed} addresses asynchronous task arrivals and heterogeneous data distributions via asynchronous prompt updates and a contrastive continual learning loss.
\textbf{FedPrompt}~\citep{zhao2023fedprompt} employs a split aggregation strategy, freezing the extensive parameters of LLMs and only tuning and aggregating soft prompts. 
\textbf{FedSP}~\citep{dong2023tunable} reduces computational and memory overhead by utilizing a lightweight auxiliary model for prompt learning.
\textbf{HePCo}~\citep{halbe2023hepco} mitigates catastrophic forgetting and data heterogeneity through a data-free distillation method performed in the model’s latent space.
\textbf{PFL-GCN}~\citep{ahmad2023prompt} employs prompt tuning specifically for sentiment analysis.

\textbf{AUG-FedPrompt}~\citep{cai2023towards} exploits abundant unlabeled data for data augmentation to address the issue of data scarcity. 
\citet{liu2023federated} integrate self-consistency and chain-of-thought prompting to improve zero-shot performance of LLMs. 
\textbf{FedHPL}~\citep{ma2024fedhpl} introduces a global logit distillation framework to handle model heterogeneity and guide the local training process.
\textbf{PFPT}~\citep{wengprobabilistic} proposes a probabilistic prompt aggregation mechanism to address data heterogeneity and imbalanced data distribution.
\textbf{FCILPT}~\citep{liu2023federated1} jointly encodes task-relevant and task-irrelevant knowledge into prompts to preserve both previous and newly learned knowledge, alleviating catastrophic forgetting.
\textbf{CaFPT}~\citep{guo2024explore} leverages information-theoretic principles to facilitates the retrieval process by conditioning on examples that activate the most relevant knowledge inside pre-trained models.
\textbf{FedPoD}~\citep{chen2023federated2} employs lightweight prompts to guide frozen foundation models and introduces multi-level prompt-based communication to enable multi-source knowledge fusion and controlled optimization.

\noindent $\bullet$ \textbf{Personalized Prompt Tuning} enables each client to tailor its prompt embeddings based on local data distributions and task-specific requirements. By fine-tuning prompts individually, clients can better capture local nuances and context-specific information that a one-size-fits-all prompt might overlook. 
This approach directly addresses the challenge of data heterogeneity by facilitating local adaptation, while still allowing clients to benefit from global knowledge aggregated during training. Table~\ref{Prompt_personlized_table} summarizes representative methods and the specific challenges they target.

\textbf{pFedPG}~\citep{yang2023efficient} deploys a personalized prompt generator on the server to produce client-specific visual prompts, enabling efficient adaptation of frozen backbones to diverse local data.
\textbf{SGPT}~\citep{deng2024unlocking} combines generalized and personalized FL by learning a mix of shared and group-specific prompts to capture both commonalities and group-specific variations.
\textbf{pFedPrompt}~\citep{guo2023pfedprompt} leverages the unique multimodal capabilities of vision-language models by learning client consensus in the linguistic space and adapting to client characteristics in the visual space in a non-parametric manner. 
\textbf{FedOTP}~\citep{li2024global} introduces efficient collaborative prompt learning strategies to capture diverse category traits on a per-client basis. 
\textbf{pFedPT}~\citep{li2023visual} utilizes personalized visual prompts to implicitly represent local data distribution information and provides this information to the aggregation model to enhance classification tasks. 
\textbf{FedMGP}~\citep{yu2024personalized} uses coarse-grained global prompts for shared knowledge and fine-grained local prompts for personalization, and introduces a selective fusion mechanism for prompt aggregation.

\textbf{FedLPPA}~\citep{lin2024fedlppa} jointly learns personalized prompts and aggregation strategies for weakly-supervised medical image segmentation.
\textbf{FedPGP}~\citep{cui2024harmonizing} employs pre-trained CLIP to provide knowledge-guidance for the global prompt, enhancing generalization while incorporating a low-rank adaptation term to personalize the global prompt. 
\textbf{FedPFT}~\citep{wu2024tackling} addresses feature-classifier mismatch through prompt-driven feature transformation.
\citet{wang2024personalized} propose a discrete local search strategy for gradient-free local training and a token-based compression method inspired by linear word analogies, substantially reducing resource costs.
\textbf{pFedMoAP}~\citep{luo2024mixture} introduces a personalized prompt learning framework based on the mixture-of-experts  paradigm~\citep{cai2024survey}.
\textbf{CP$^2$GFed}~\citep{gao2024cp} introduces a cross-granularity knowledge transfer mechanism and dynamic personalized prompt generation to improve model performance.

\begin{table}[!t]
\small
\centering
\caption{\textbf{Personalized prompt tuning in federated fine-tuning.}}
\label{Prompt_personlized_table}
\resizebox{0.8\linewidth}{!}{
\begin{tabular}{lccccccc}
\toprule[1pt]
\multirow{2}{*}{\textbf{Method}}& \multicolumn{4}{c}{\textbf{Challenge}}  \\  

\cmidrule{2-5}

& \textbf{Communication} & \textbf{Non-IID} & \textbf{Memory} & \textbf{Computation}  \\ 

\midrule[1pt]
{pFedPG~\citep{yang2023efficient}}&\hlr{\xmark}&\hlg{\tmark{}}&\hlr{\xmark}&\hlr{\xmark}\\ 
{SGPT~\citep{deng2024unlocking}}&\hlr{\xmark}&\hlg{\tmark{}}&\hlr{\xmark}&\hlr{\xmark}\\ 
{pFedPrompt~\citep{guo2023pfedprompt}}&\hlr{\xmark}&\hlg{\tmark{}}&\hlr{\xmark}&\hlr{\xmark}\\ 
{FedOTP~\citep{li2024global}}&\hlr{\xmark}&\hlg{\tmark{}}&\hlr{\xmark}&\hlr{\xmark}\\ 
{pFedPT~\citep{li2023visual}}&\hlr{\xmark}&\hlg{\tmark{}}&\hlr{\xmark}&\hlr{\xmark}\\ 
{FedMGP~\citep{yu2024personalized}}&\hlr{\xmark}&\hlg{\tmark{}}&\hlr{\xmark}&\hlr{\xmark}\\ 
{FedLPPA~\citep{lin2024fedlppa}}&\hlr{\xmark}&\hlg{\tmark{}}&\hlr{\xmark}&\hlr{\xmark}\\ 
{FedPGP~\citep{cui2024harmonizing}}&\hlr{\xmark}&\hlg{\tmark{}}&\hlr{\xmark}&\hlr{\xmark}\\ 
{FedPFT~\citep{wu2024tackling}}&\hlr{\xmark}&\hlg{\tmark{}}&\hlr{\xmark}&\hlr{\xmark}\\ 
{\citet{wang2024personalized}}&\hlg{\tmark{}}&\hlg{\tmark{}}&\hlg{\tmark{}}&\hlg{\tmark{}}\\
{pFedMoAP~\citep{luo2024mixture}}&\hlr{\xmark}&\hlg{\tmark{}}&\hlr{\xmark}&\hlr{\xmark}\\ 
{CP$^2$GFed~\citep{gao2024cp}}&\hlr{\xmark}&\hlg{\tmark{}}&\hlr{\xmark}&\hlg{\tmark{}}\\
\bottomrule[1pt]
\end{tabular}
}
\vspace{-5mm}
\end{table}

\begin{table}[!t]
\small
\centering
\caption{\textbf{Multi-domain prompt tuning in federated fine-tuning.}}
\label{Prompt_multidomain_table}
\resizebox{0.8\linewidth}{!}{
\begin{tabular}{lccccccc}
\toprule[1pt]
\multirow{2}{*}{\textbf{Method}}& \multicolumn{4}{c}{\textbf{Challenge}}  \\  

\cmidrule{2-5}

& \textbf{Communication} & \textbf{Non-IID} & \textbf{Memory} & \textbf{Computation}  \\ 

\midrule[1pt]

{DiPrompT~\citep{bai2024diprompt}}&\hlr{\xmark}&\hlg{\tmark{}}&\hlr{\xmark}&\hlr{\xmark}\\ 
{PFCR~\citep{guo2024prompt}}&\hlr{\xmark}&\hlg{\tmark{}}&\hlr{\xmark}&\hlr{\xmark}\\
{Fed-DPT~\citep{wei2023dual}}&\hlr{\xmark}&\hlg{\tmark{}}&\hlr{\xmark}&\hlr{\xmark}\\
{FedAPT~\citep{su2024federated2}}&\hlr{\xmark}&\hlg{\tmark{}}&\hlr{\xmark}&\hlr{\xmark}\\
{\citet{zhao2024breaking}}&\hlr{\xmark}&\hlg{\tmark{}}&\hlr{\xmark}&\hlr{\xmark}\\
{FedDG~\citep{gong2024federated}}&\hlr{\xmark}&\hlg{\tmark{}}&\hlr{\xmark}&\hlr{\xmark}\\
{CP-Prompt~\citep{feng2024cp}}&\hlr{\xmark}&\hlg{\tmark{}}&\hlr{\xmark}&\hlr{\xmark}\\
\bottomrule[1pt]
\end{tabular}
}
\vspace{-3mm}
\end{table}

\noindent $\bullet$ \textbf{Multi-Domain Prompt Tuning} extends the prompt-based approach to environments where federated clients operate across distinct domains or application contexts. In such scenarios, each client is equipped with domain-specific prompt embeddings that adapt the shared global model to diverse contextual and distributional conditions. This approach enhances the model’s generalization ability across heterogeneous domains while maintaining a shared global foundation. It is particularly valuable in real-world deployments spanning multiple industries or task categories. Table~\ref{Prompt_multidomain_table} summarizes representative methods and the specific challenges they address.

\textbf{DiPrompT}~\citep{bai2024diprompt} proposes a distributed domain generalization approach using adaptive prompts, introducing global prompts for shared knowledge and domain prompts for domain-specific adaptation. 
\textbf{PFCR}~\citep{guo2024prompt} eliminates the need for raw data sharing via encrypted gradient updates, models items in a unified feature space using descriptive text, and facilitates cross-domain knowledge transfer through federated content representations and prompt tuning.
\textbf{Fed-DPT}~\citep{wei2023dual} leverages a pre-trained vision-language model and applies dual prompt tuning—combining visual and textual prompts—for improved domain alignment across decentralized data sources. 
\textbf{FedAPT}~\citep{su2024federated2} introduces a meta prompt, an adaptive network, and frozen keys to personalize prompts for each test sample, thereby enhancing multi-domain image classification. \citet{zhao2024breaking} propose a language distance metric to improve data efficiency and facilitate cross-linguistic generalization. 
\textbf{FedDG}~\citep{gong2024federated} allows clients to learn text and visual prompts locally while maintaining indirect alignment via global prompts used as a shared reference. Domain-specific prompts are exchanged among clients and selectively integrated into global prompts using lightweight attention-based aggregators.
\textbf{CP-Prompt}~\citep{feng2024cp} captures intra-domain knowledge by inserting personalized prompts into the multi-head attention modules and subsequently learns inter-domain representations through a shared prompting mechanism.

\subsection{Adapter-based Tuning}\label{sec_adapter}


\subsubsection{Preliminary}

\begin{wrapfigure}{r}{0.3\linewidth} 
\centering
\vspace{-2mm}
\adjustbox{max width=\linewidth}{\includegraphics{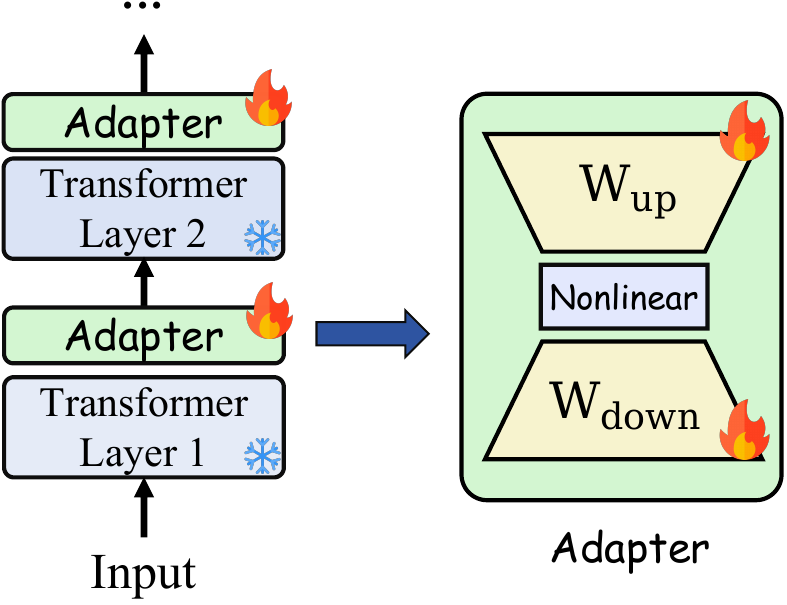}}
\vspace{-6mm}
\caption{The working principle of adapter tuning.}
\label{Fig_Adapter}
\vspace{-5mm}
\end{wrapfigure}

Adapter-based tuning is another parameter-efficient alternative to full-parameter fine-tuning for LLMs~\citep{pfeiffer2020adapterhub}. It introduces lightweight, trainable adapter modules into the model while keeping the pre-trained weights frozen. These modules act as task-specific components that transform intermediate representations in a controlled manner, enabling efficient adaptation to downstream tasks with minimal memory and computational overhead. A standard adapter module consists of three key operations: \textbf{down-projection}, \textbf{non-linearity}, and \textbf{up-projection}, as shown in Figure~\ref{Fig_Adapter}. For activations \( h_{i} \in \mathbb{R}^{n \times d} \), the adapter transformation proceeds as follows:

\textbf{1) Down-Projection:} The high-dimensional hidden state \( h_{i} \) is projected into a low-dimensional space using a learnable weight matrix \( W_{DP} \in \mathbb{R}^{d \times r} \), where \( r \ll d \) controls the bottleneck size. This step reduces the number of trainable parameters while capturing essential features:
\begin{equation}
    h_{i}^{'} = h_{i} W_{DP} 
\end{equation}

\textbf{2) Non-Linearity:} A non-linear activation function \( \sigma(\cdot) \), such as ReLU or GELU, is applied to introduce expressive transformations while retaining important task-specific patterns:
\begin{equation}
    h_{i}^{''} = \sigma(h_{i}^{'})
\end{equation}

\textbf{3) Up-Projection:} The transformed low-dimensional representation is mapped back to the original feature space using an \textbf{up-projection} matrix \( W_{UP} \in \mathbb{R}^{r \times d} \):
\begin{equation}
    h_{i}^{'''} =  h_{i}^{''} W_{UP}
\end{equation}

\textbf{4) Residual Connection:} The final output of the adapter is then residually added to the original hidden state, preserving the pre-trained knowledge while incorporating task-specific adjustments:
\begin{equation}
    Z = h_{i} + h_{i}^{'''}
\end{equation}
where \( Z \) represents the adapted hidden representation. This residual connection ensures that the pre-trained model remains largely intact while allowing task-specific fine-tuning through the lightweight adapter layers.  Notably, only the adapter parameters \( W_{DP} \) and \( W_{UP} \) are updated during training, resulting in significantly lower memory and computation costs compared to full-parameter fine-tuning.

\subsubsection{Adapter in Federated Fine-Tuning}

In the context of federated fine-tuning, adapter-based tuning provides significant advantages in both resource efficiency and model adaptability. Since only the lightweight adapter modules are updated and exchanged, rather than the full model parameters, this approach greatly reduces computation and communication overhead. Moreover, by freezing the base model, adapter-based tuning enables clients to fine-tune efficiently on heterogeneous local data while still benefiting from the shared global knowledge encoded in the pre-trained model.
This modular design facilitates the seamless integration of task-specific adaptations without compromising the generalization capability of the base model. To better understand the landscape of adapter-based methods in federated fine-tuning, we propose a new taxonomy comprising three categories: General Adapter Tuning, Personalized Adapter Tuning, and Multi-Domain Adapter Tuning, as illustrated in Figure~\ref{main_table}.
In the following sections, we explore representative methods within each category and analyze how they address the core challenges identified in Section~\ref{sec_challenge}, beyond the inherent  benefits brought by adapter itself.

\begin{table}[!t]
\small
\centering
\caption{\textbf{General, personalized, and multi-domain adapter tuning in federated fine-tuning.}}
\label{Adapter_General_table}
\resizebox{0.8\linewidth}{!}{
\begin{tabular}{lccccccc}
\toprule[1pt]
\multirow{2}{*}{\textbf{Method}}& \multirow{2}{*}{\textbf{Type}}&\multicolumn{4}{c}{\textbf{Challenge}}  \\  
\cmidrule{3-6}
& &\textbf{Communication} & \textbf{Non-IID} & \textbf{Memory} & \textbf{Computation} \\ 

\midrule[1pt]
{FedAdapter~\citep{cai2023efficient}}&General&\hlg{\tmark{}}&\hlr{\xmark}&\hlr{\xmark}&\hlg{\tmark{}}\\
{\citet{kim2023efficient2}}&General&\hlg{\tmark{}}&\hlg{\tmark{}}&\hlg{\tmark{}}&\hlg{\tmark{}}\\{FedTT+~\citep{ghiasvand2024communication}}&General&\hlg{\tmark{}}&\hlg{\tmark{}}&\hlr{\xmark}&\hlr{\xmark}\\

\midrule[1pt]
{C2A~\citep{kim2023client}}&Personalized&\hlr{\xmark}&\hlg{\tmark{}}&\hlr{\xmark}&\hlr{\xmark}\\
{FedCLIP~\citep{lu2023fedclip}}&Personalized&\hlr{\xmark}&\hlg{\tmark{}}&\hlr{\xmark}&\hlr{\xmark}\\

\midrule[1pt]
{Fed-MNMT~\citep{liu2023communication}}&Multi-domain&\hlr{\xmark}&\hlg{\tmark{}}&\hlr{\xmark}&\hlr{\xmark}\\
{AdaFedSelecKD~\citep{feng2024adapter}}&Multi-domain&\hlr{\xmark}&\hlg{\tmark{}}&\hlr{\xmark}&\hlr{\xmark}\\
{FedDAT~\citep{chen2024feddat}}&Multi-domain&\hlr{\xmark}&\hlg{\tmark{}}&\hlr{\xmark}&\hlr{\xmark}\\

\bottomrule[1pt]
\end{tabular}
}
\vspace{-3mm}
\end{table}

\noindent $\bullet$ \textbf{General Adapter Tuning} refers to scenarios in which all clients utilize a shared adapter structure with identical initialization. In this setting, the same adapter modules are inserted into the transformer layers of each client’s model, enabling consistent adaptation mechanisms across the federation. This uniformity facilitates stable aggregation and coordinated updates during federated training. Such an approach is particularly effective when clients operate on similar tasks or share relatively homogeneous data distributions, as a globally optimized adapter can generalize well across participants. Table~\ref{Adapter_General_table} summarizes representative methods and the specific challenges they aim to address.

\textbf{FedAdapter}~\citep{cai2023efficient} proposes a progressive adapter tuning strategy, combined with continuous device profiling, to dynamically optimize adapter configurations across clients, improving efficiency without sacrificing accuracy.
\citet{kim2023efficient2} leverage adapters to address the high communication costs associated with federated fine-tuning of LLMs. 
\textbf{FedTT+}~\citep{ghiasvand2024communication} integrates tensorized adapters for LLM adaptation and further improves robustness to data heterogeneity by freezing portions of the tensor factors, significantly reducing the number of trainable parameters while maintaining model performance.

\noindent $\bullet$ \textbf{Personalized Adapter Tuning} enables each client to independently fine-tune its adapter modules based on its local data distribution and task-specific requirements. In contrast to general adapter tuning, this approach does not enforce uniformity across clients; instead, it allows for the retention of personalized adapter parameters that better capture client-specific knowledge. This strategy is particularly advantageous in federated settings characterized by high degrees of data heterogeneity. By leveraging personalized adapters, clients can achieve improved local performance while still benefiting from shared global knowledge. Table~\ref{Adapter_General_table} summarizes representative methods and the specific challenges they address.

\textbf{C2A}~\citep{kim2023client} employs a hypernetwork to generate client-specific adapters, effectively addressing data heterogeneity by enabling on-demand parameter generation tailored to each client.
\textbf{FedCLIP}~\citep{lu2023fedclip} introduces an attention-based adapter design that utilizes the pre-trained model's knowledge to facilitate both rapid generalization and efficient personalization while minimizing resource overhead.

\noindent $\bullet$ \textbf{Multi-Domain Adapter Tuning} extends the federated fine-tuning paradigm to clients operating across distinct domains, enabling efficient adaptation to domain-specific tasks. In this setting, each client maintains its own domain-specific adapter while contributing to a shared global model. The global model aggregates adapter updates across domains to capture  domain-invariant representations to support generalization.
This approach is particularly effective in cross-domain scenarios such as multilingual natural language processing. By decoupling domain-specific learning from the shared backbone, this strategy balances personalization and collaboration. Table~\ref{Adapter_General_table} summarizes representative methods and the challenges they address.

\textbf{Fed-MNMT}~\citep{liu2023communication} applies adapter-based fine-tuning for multilingual neural machine translation, significantly reducing communication overhead. It further explores parameter clustering strategies to mitigate conflicts during aggregation.
\textbf{AdaFedSelecKD}~\citep{feng2024adapter} performs adapter-based summarization to minimize transmitted parameters and introduces selective knowledge distillation for efficient domain-adaptive learning.
\textbf{FedDAT}~\citep{chen2024feddat} proposes a dual-adapter teacher framework to regularize local updates under data heterogeneity, and employs mutual knowledge distillation for effective cross-client knowledge transfer.

\subsection{Selective-based Tuning}\label{sec_selective}

\subsubsection{Preliminary}

Selective-based tuning has emerged as an efficient strategy for fine-tuning LLMs by updating specific parameters of the model while keeping the majority of pre-trained weights frozen~\citep{kornblith2019better}. This approach significantly reduces computational and memory overhead compared to full-parameter fine-tuning, while maintaining the model’s generalization ability.
Among selective-based tuning techniques, two widely adopted strategies are bias tuning~\citep{zaken2021bitfit} and partial tuning~\citep{houlsby2019parameter}, both of which optimize only a subset of parameters rather than the entire model.

Bias tuning updates only the bias terms of the model while keeping all other parameters frozen. Despite its simplicity, this approach has demonstrated strong performance across a variety of tasks with minimal overhead.
Partial tuning generalizes this idea by allowing updates to a carefully selected subset of model parameters, such as layer normalization parameters, feed-forward network biases, or specific attention blocks.
By focusing updates on the most relevant parameters, selective-based tuning methods improve training efficiency, mitigate catastrophic forgetting, and enable rapid adaptation using limited data and resources.

\subsubsection{Selective Fine-Tuning Methods}

\textbf{DP-BiTFiT}~\citep{bu2022differentially} applies differentially private bias-term tuning in centralized training scenarios to ensure privacy-preserving adaptation while reducing resource demands.
\textbf{FedPEFT}~\citep{sun2022conquering} shares only a small subset of model weights, such as bias parameters, significantly reducing the communication overhead. 
\textbf{RaFFM}~\citep{yu2023bridging} introduces a resource-aware model compression framework tailored for FL, which includes salient parameter prioritization and subnetwork extraction to support dynamic model scaling across heterogeneous edge devices.
\citet{sun2024exploring} propose a selective layer-wise fine-tuning approach to reduce training cost while preserving model performance.

\subsection{Other Tuning Methods}\label{sec_hybrid}

In addition to mainstream fine-tuning strategies, several alternative approaches have been explored to optimize LLMs in FL. These methods primarily include \textbf{zeroth-order optimization}~\citep{chen2017zoo}, \textbf{split learning}~\citep{thapa2022splitfed}, \textbf{model compression}~\citep{choudhary2020comprehensive}, and \textbf{data selection}~\citep{shen2024rethinking} (as shown in Figure~\ref{main_table}). For example, \textbf{FedKSeed}~\citep{qin2023federated} employs zeroth-order optimization with a finite set of random seeds, enabling LLM fine-tuning without storing intermediate activations and reducing communication overhead. \textbf{FedBERT}~\citep{tian2022fedbert} combines FL with split learning to pre-train BERT in a distributed, privacy-preserving manner, achieving efficient model training across decentralized clients. \textbf{FedBiOT}~\citep{wu2024fedbiot} compresses the LLM at the server side while allowing clients to fine-tune lightweight adapters, significantly reducing resource consumption. \textbf{FedHDS}~\citep{qin2024federated} introduces a hierarchical data selection framework that identifies representative coresets for instruction tuning, minimizing redundancy at both intra- and inter-client levels to improve training efficiency in FL.

\section{Datasets and Benchmarks}\label{sec_evaluation}

A comprehensive evaluation framework is essential for systematically assessing the effectiveness and generalization ability of federated fine-tuning methods. In this section, we begin by presenting widely-used fine-tuning datasets spanning multiple domains. We then detail a suite of domain-specific evaluation benchmarks that enable consistent, fine-grained, and standardized assessment of FedLLM performance across diverse and heterogeneous task settings.

\subsection{Instruction Fine-Tuning Datasets}

Table~\ref{tab_finetune_dataset} and Table~\ref{tab_finetune_dataset2} present a comprehensive collection of instruction fine-tuning datasets spanning several key domains, including general language understanding, finance, medicine, code, math, and law. For each domain, we select representative datasets from Hugging Face that are widely adopted by the community and closely aligned with real-world FedLLM applications. Building on this overview, we next provide a detailed introduction to the datasets within each domain.

\subsubsection{General Instruction Fine-Tuning Datasets}

General instruction fine-tuning datasets are primarily designed to improve the overall instruction-following capability of LLMs across a wide range of tasks and domains~\citep{zhou2023instruction}. These datasets serve as a foundation for aligning LLMs with human intent, and are particularly useful in federated settings where clients may engage in diverse yet general-purpose interactions.

For English instruction fine-tuning, representative datasets include Alpaca~\citep{taori2023stanford}, which is generated by Text-Davinci-003 with Alpaca-style instruction prompts, and Alpaca-GPT4~\citep{peng2023instruction}, which builds on this with GPT-4-generated, multi-turn dialogues to improve linguistic nuance and contextual coherence. Other commonly used datasets include Self-Instruct~\citep{wang2022self}, UltraChat 200k~\citep{ding2023enhancing}, OpenOrca~\citep{lian2023openorca}, ShareGPT-90K\footnote{https://huggingface.co/datasets/liyucheng/ShareGPT90K}, WizardLM Evol-Instruct V2 196k~\citep{xu2023wizardlm}, Databricks Dolly 15K~\citep{conover2023free}, Baize~\citep{xu2023baize}, OpenChat~\citep{wang2023openchat}, and Flan-v2\footnote{https://huggingface.co/datasets/SirNeural/flan$\_$v2}. 


For Chinese instruction tuning, the landscape includes representative datasets such as BELLE-CN~\citep{ji2023belle}, which focuses on improving model alignment with Chinese user instructions. This is complemented by additional resources like Firefly-train-1.1M~\citep{Firefly}, Wizard-LM-Chinese-Instruct-Evol~\citep{xu2023wizardlm}, and HC3-Chinese~\citep{guo2023close}. These resources facilitate a diverse range of instruction tasks, including reasoning, question answering, and domain-specific knowledge modeling.

Bilingual instruction datasets are also included to support multilingual and cross-lingual instruction tuning in federated contexts. HC3~\citep{guo2023close} offers both human- and model-generated responses in English and Chinese for evaluating factual consistency and detection capabilities. ShareGPT-Chinese-English-90k~\citep{ShareGPT-Chinese-English-90k} provides high-quality, bilingual conversations, making it particularly suitable for instruction tuning in multilingual FedLLM scenarios.
A detailed comparison is provided in Table~\ref{tab_finetune_dataset}.

\begin{table}[!t]
\footnotesize
\centering
\renewcommand{\arraystretch}{0.98}

\caption{A summary of representative {instruction fine-tuning datasets} in \textbf{general, financial, and medical} domains. The construction methods are categorized into three types: human construct, which refers to datasets written or annotated by humans; model construct, which refers to data generated by prompting LLMs; and synthetic, which refers to data produced through rule-based or programmatic generation. }

\label{tab_finetune_dataset}
\resizebox{1\linewidth}{!}{
\begin{tabular}{lccccccc}
\toprule[1pt]
\midrule[0.5pt]

\textbf{Dataset}& \textbf{Language} &\textbf{Construction} & \textbf{Domain} & \textbf{Description} \\ 

\midrule[0.5pt]
\href{https://huggingface.co/datasets/tatsu-lab/alpaca}{Alpaca} & English & Model & General & Generated by Text-Davinci-003 (Alpaca-style)\\

\href{https://huggingface.co/datasets/vicgalle/alpaca-gpt4}{Alpaca-GPT4} & English & Model & General 
& Generated by GPT-4 (multi-turn Alpaca prompts)\\

\href{https://huggingface.co/datasets/yizhongw/self_instruct}{Self-Instruct} & English & Human + Model & General & Generated from seed instructions via GPT-3\\

\href{https://huggingface.co/datasets/HuggingFaceH4/ultrachat_200k}{UltraChat 200k} & English & Model & General & Multi-turn dialogues filtered from UltraChat\\

\href{https://huggingface.co/datasets/Open-Orca/OpenOrca}{OpenOrca} & English & Model & General & 
~4.2M GPT-3.5/4 augmented FLAN examples\\

\href{https://huggingface.co/datasets/liyucheng/ShareGPT90K}{ShareGPT90K} & English & Model & General & 90K multi-turn dialogues from ShareGPT \\

\href{https://huggingface.co/datasets/WizardLMTeam/WizardLM_evol_instruct_V2_196k}{WizardLM Evol-Instruct V2} & English & Model & General & 196K samples generated via Evol-Instruct \\

\href{https://huggingface.co/datasets/databricks/databricks-dolly-15k}{Databricks Dolly 15K} & English & Human & General & 15K human-generated prompt-response pairs \\

\href{https://huggingface.co/datasets/linkanjarad/baize-chat-data}{Baize} & English & Model & General 
& Dialogues from ChatGPT (prompted by user queries)\\

\href{https://huggingface.co/openchat}{OpenChat} & English & Model & General 
& Dialogues from open LLMs for multi-turn alignment \\

\href{https://huggingface.co/datasets/SirNeural/flan_v2}{Flan-v2} & English & Model & General 
& Combination of Flan, P3, SNI, CoT, and Dialog tasks\\



\href{https://huggingface.co/datasets/BelleGroup/train_2M_CN}{BELLE-CN} & Chinese & Human + Model & General & 2M Chinese instruction-following samples\\

\href{https://huggingface.co/datasets/YeungNLP/firefly-train-1.1M}{Firefly-train-1.1M} & Chinese & Human & General & 1.65M Chinese samples (23 tasks, human templates) \\

\href{https://huggingface.co/datasets/silk-road/Wizard-LM-Chinese-instruct-evol}{Wizard-LM-Chinese-instruct-evol} & Chinese & Human + Model & General & 
70K Chinese samples (translated from WizardLM) \\

\href{https://huggingface.co/datasets/Hello-SimpleAI/HC3-Chinese}{HC3-Chinese} & Chinese & Human + Model & General & Chinese Human-ChatGPT QA pairs (multi-domain) \\

\href{https://huggingface.co/datasets/Hello-SimpleAI/HC3}{HC3} & English / Chinese & Human + Model & General & Human-ChatGPT QA pairs (multi-domain) \\

\href{https://huggingface.co/datasets/shareAI/ShareGPT-Chinese-English-90k}{ShareGPT-Chinese-English-90k} & English / Chinese & Model & General 
& 90K bilingual human-machine QA pairs\\

\midrule[0.5pt]

\href{https://huggingface.co/datasets/FinGPT/fingpt-finred}{FinGPT} & English & Human + Model & Financial 
& Instruction-tuning data for financial tasks\\

\href{https://huggingface.co/datasets/Josephgflowers/Finance-Instruct-500k}{Finance-Instruct-500k} & English & Human + Model & Financial 
& Large-scale instruction-tuning data for financial tasks\\

\href{https://huggingface.co/datasets/gbharti/finance-alpaca}{Finance-Alpaca} & English & Human + Model & Financial & Instruction-tuning data combining Alpaca and FiQA\\


\href{https://huggingface.co/datasets/takala/financial_phrasebank}{Financial PhraseBank} & English & Human & Financial & Manually annotated financial news for sentiment\\

\href{https://huggingface.co/datasets/bwzheng2010/yahoo-finance-data}{Yahoo-Finance-Data} & English & Human & Financial & Yahoo finance data \\

\href{https://huggingface.co/datasets/virattt/financial-qa-10K}{Financial-QA-10K} & English &  Model & Financial & Contextual QA for financial answering and retrieval\\

\href{https://huggingface.co/datasets/nickmuchi/financial-classification}{Financial-Classification} & English & Human  & Financial & Financial PhraseBank and Kaggle financial texts\\

\href{https://huggingface.co/datasets/zeroshot/twitter-financial-news-topic}{Twitter-Financial-News-Topic} & English & Human & Financial & Annotated tweets for financial topic classification\\

\href{https://huggingface.co/datasets/ashraq/financial-news-articles}{Financial-News-Articles} & English & Human & Financial & Financial articles for classification and sentiment analysis\\

\href{https://huggingface.co/datasets/BeIR/fiqa}{FiQA} & English & Human & Financial & QA from financial texts and forums\\

\href{https://huggingface.co/datasets/lamini/earnings-calls-qa}{Earnings-Call} & English & Human & Financial & QA pairs from CEO/CFO earnings calls\\

\href{https://github.com/shun-zheng/Doc2EDAG}{Doc2EDAG} & Chinese & Human & Financial & Financial reports annotated for event graph extraction\\

\href{https://huggingface.co/datasets/gretelai/synthetic_pii_finance_multilingual}{Synthetic-PII-Finance-Multilingual} & Multilingual & Synthetic & Financial & Synthetic financial docs with labeled PII \\

\midrule[0.5pt]

\href{https://huggingface.co/datasets/LinhDuong/chatdoctor-200k}{ChatDoctor-200K} & English & Human + Model & Medical & Medical QA and dialogue instructions\\

\href{https://huggingface.co/datasets/lavita/ChatDoctor-HealthCareMagic-100k}{ChatDoctor-HealthCareMagic-100k} & English & Human & Medical & Real-world doctor-patient conversations\\

\href{https://huggingface.co/datasets/medalpaca/medical_meadow_cord19}{Medical Meadow CORD-19} & English & Human & Medical & Literature summarization instructions (CORD-19)\\

\href{https://huggingface.co/datasets/medalpaca/medical_meadow_medqa}{Medical Meadow MedQA} & English & Human & Medical & Medical QA from professional board exams\\

\href{https://huggingface.co/datasets/wangrongsheng/HealthCareMagic-100k-en}{HealthCareMagic-100k-en} & English & Human & Medical & Medical and real doctor-patient consultations\\

\href{https://huggingface.co/datasets/michaelwzhu/ChatMed_Consult_Dataset}{ChatMed-Consult-Dataset} & Chinese & Human + Model & Medical & Medical consultation QA and dialogue instructions\\

\href{https://github.com/SupritYoung/Zhongjing}{CMtMedQA} & Chinese & Human & Medical & Multi-turn real medical QA\\

\href{https://huggingface.co/datasets/Flmc/DISC-Med-SFT}{DISC-Med-SFT} & Chinese & Human + Model & Medical & Real dialogues and knowledge graph QA pairs\\


\href{https://huggingface.co/datasets/FreedomIntelligence/HuatuoGPT-sft-data-v1}{HuatuoGPT-SFT} & Chinese & Human + Model & Medical & ChatGPT-generated and real doctor-patient dialogues\\

\href{https://huggingface.co/datasets/FreedomIntelligence/Huatuo26M-Lite}{Huatuo26M-Lite} & Chinese & Human + Model & Medical & Refined subset of Huatuo-26M (ChatGPT-rewritten)\\

\href{https://huggingface.co/datasets/michaelwzhu/ShenNong_TCM_Dataset}{ShenNong-TCM} & Chinese & Human + Model & Medical & QA pairs from TCM knowledge graph\\

\href{https://huggingface.co/datasets/bigbio/meddialog}{MedDialog} & English / Chinese & Human & Medical & Large-scale doctor-patient dialogues\\

\midrule[0.5pt]
\bottomrule[1pt]
\end{tabular}
}
\end{table}

\subsubsection{Financial Domain Instruction Fine-Tuning Datasets}

Instruction fine-tuning datasets in the financial domain are tailored to equip language models with specialized knowledge and reasoning capabilities relevant to financial tasks~\citep{li2023large}. In the context of FedLLM, such datasets are particularly valuable for enabling privacy-preserving and institution-specific applications, including investment recommendation, market sentiment analysis, financial reporting assistance, and regulatory compliance. These use cases often involve sensitive and proprietary data that cannot be centrally aggregated due to privacy, confidentiality, or regulatory constraints~\citep{byrd2020differentially}, making federated fine-tuning an ideal solution.

For English instruction tuning, representative datasets include FinGPT~\citep{zhang2023instructfingpt}, which provides a large corpus of financial question-answering pairs and document summaries tailored to real-world financial analysis. Finance-Instruct-500k~\citep{josephgflowers2025financeinstruct} and Finance-Alpaca\footnote{https://huggingface.co/datasets/gbharti/finance-alpaca} extend general instruction-tuning formats to financial scenarios, offering instruction–response pairs related to stock prediction, portfolio analysis, and macroeconomic commentary. Additional datasets such as Financial PhraseBank~\citep{malo2014good}, Yahoo-Finance-Data\footnote{https://huggingface.co/datasets/bwzheng2010/yahoo-finance-data}, Financial-QA-10K\footnote{https://huggingface.co/datasets/virattt/financial-qa-10K}, Financial-Classification\footnote{https://huggingface.co/datasets/nickmuchi/financial-classification}, Twitter-Financial-News-Topic\footnote{https://huggingface.co/datasets/zeroshot/twitter-financial-news-topic}, Financial-News-Articles\footnote{https://huggingface.co/datasets/ashraq/financial-news-articles}, and FiQA~\citep{maia201818} cover a broad spectrum of financial tasks including sentiment classification, time-series event extraction, and financial question answering. Transcripts from earnings calls further support fine-tuning for multi-turn, dialogue-based financial reasoning.

For Chinese financial applications, Doc2EDAG~\citep{zheng2019doc2edag} provides a rich dataset for event detection and argument generation from financial documents, supporting instruction-style tasks. In addition, the Synthetic-PII-Finance-Multilingual dataset~\citep{gretel-synthetic-pii-finance-multilingual-2024} includes multi-language synthetic financial records annotated with privacy-related attributes. A comparative summary of these finance-related datasets is presented in Table~\ref{tab_finetune_dataset}.

\subsubsection{Medical Domain Instruction Fine-Tuning Datasets}

Instruction fine-tuning datasets in the medical domain are designed to enable LLMs to perform tasks such as medical reasoning, patient interaction, and clinical decision support~\citep{zhang2023alpacare}. Within the context of FedLLM, these datasets are particularly relevant for privacy-preserving healthcare applications, where sensitive patient data is inherently distributed across hospitals, clinics, and personal health devices, and cannot be centrally aggregated due to strict privacy regulations~\citep{nguyen2022federated}. Federated fine-tuning with medical instruction data empowers models to generalize across diverse clinical intents while preserving the confidentiality and heterogeneity of local medical records, thereby supporting robust and compliant deployment in real-world healthcare settings.

For English-language datasets, ChatDoctor-200K~\citep{li2023chatdoctor} and ChatDoctor-HealthCareMagic-100k~\citep{li2023chatdoctor} contain medical dialogues derived from professional consultation platforms, facilitating multi-turn reasoning and symptom analysis. The Medical Meadow~\citep{han2023medalpaca} suite offers curated datasets from sources such as CORD-19\footnote{https://huggingface.co/datasets/medalpaca/medical$\_$meadow$\_$cord19} and MedQA\footnote{https://huggingface.co/datasets/medalpaca/medical$\_$meadow$\_$medqa}, targeting tasks like biomedical question answering, literature summarization, and clinical fact verification. Additionally, HealthCareMagic-100k-en\footnote{https://huggingface.co/datasets/wangrongsheng/HealthCareMagic-100k-en} supports English patient–doctor interactions across various medical specialties.

For Chinese medical instruction tuning, a rich set of datasets has been developed to address the unique linguistic and clinical characteristics of Chinese healthcare scenarios. Notable examples include ChatMed-Consult-Dataset~\citep{zhu2023chatmed} and CMtMedQA~\citep{yang2024zhongjing}, which focus on consultation-style QA pairs. DISC-Med-SFT~\citep{bao2023disc}, HuatuoGPT-SFT~\citep{li2023huatuo}, and Huatuo26M-Lite\footnote{https://huggingface.co/datasets/FreedomIntelligence/Huatuo26M-Lite} offer large-scale instruction–response pairs in clinical medicine, public health, and disease treatment. Traditional Chinese Medicine (TCM) is also covered through datasets like ShenNong-TCM~\citep{zhu2023shennong}, enabling LLMs to support specific diagnostic and treatment tasks.

The MedDialog~\citep{zeng2020meddialog} dataset provides bilingual (English and Chinese) medical dialogues, supporting cross-lingual instruction tuning and evaluation in federated medical environments where linguistic diversity and clinical protocol variance are prevalent. A comparative summary of these medical-specific datasets, including language, construction method, and description, is provided in Table~\ref{tab_finetune_dataset}.

\begin{table}[!t]
\footnotesize
\centering
\renewcommand{\arraystretch}{0.98}  

\caption{A summary of representative instruction fine-tuning datasets in \textbf{code, math, and law} domains.}

\label{tab_finetune_dataset2}
\resizebox{1\linewidth}{!}{
\begin{tabular}{lccccccc}
\toprule[1pt]
\midrule[0.5pt]

\textbf{Dataset}& \textbf{Language} &\textbf{Construction} & \textbf{Domain} & \textbf{Description} \\ 

\midrule[0.5pt]

\href{https://huggingface.co/datasets/sahil2801/CodeAlpaca-20k}{CodeAlpaca} & English & Model & Code & GPT-generated code instructions (Alpaca-style)\\

\href{https://huggingface.co/datasets/iamtarun/code_instructions_120k_alpaca}{Code Instructions 120k Alpaca} & English & Human + Model & Code & Code generation instructions (Alpaca-style)\\

\href{https://huggingface.co/datasets/deepmind/code_contests}{CodeContests} & English & Human & Code & Competitive programming problems \\

\href{https://huggingface.co/datasets/bigcode/commitpackft}{CommitPackFT} & English & Human & Code & Filtered GitHub commits with high-quality messages\\

\href{https://github.com/OpenBMB/ToolBench}{ToolBench} & English & Human + Model & Code & Instructions for multi-tool API usage\\

\href{https://huggingface.co/datasets/codeparrot/codeparrot-clean}{CodeParrot} & English & Human & Code & Deduplicated and filtered Python files from GitHub \\

\href{https://huggingface.co/datasets/bigcode/the-stack-v2-dedup}{The Stack v2 Dedup} & English & Human & Code & Large-scale deduplicated source code\\

\href{https://huggingface.co/datasets/irds/codesearchnet}{CodeSearchNet} & English & Human & Code & 
Code-language pairs for retrieval and semantic search\\

\href{https://huggingface.co/datasets/open-r1/codeforces-cots}{CodeForces-CoTs} & English & Human + Model & Code & 10k CodeForces problems with CoT (DeepSeek R1) \\

\href{https://huggingface.co/datasets/google/code_x_glue_cc_code_refinement}{CodeXGLUE Code Refinement} & English & Human & Code & Buggy and fixed Java functions for code refinement \\

\midrule[0.5pt]

\href{https://huggingface.co/datasets/openai/gsm8k}{GSM8K} & English & Human & Math & 8.5k grade school math word problems \\

\href{https://huggingface.co/datasets/ankner/gsm8k-CoT}{CoT-GSM8k} & English & Human + Model & Math 
& GSM8K extended with CoT reasoning steps \\

\href{https://huggingface.co/datasets/TIGER-Lab/MathInstruct}{MathInstruct} & English & Human + Model & Math & CoT and Program-of-Thought rationales \\

\href{https://huggingface.co/datasets/meta-math/MetaMathQA}{MetaMathQA} & English & Human + Model & Math 
& Question augmentations from GSM8K and MATH \\

\href{https://huggingface.co/datasets/open-r1/OpenR1-Math-220k}{OpenR1-Math-220k} & English & Human + Model & Math & 220k math problems with CoT (DeepSeek R1) \\

\href{https://huggingface.co/datasets/hendrycks/competition_math}{Competition MATH} & English & Human & Math & 12.5K high school competition math problems \\

\href{https://huggingface.co/datasets/deepmind/math_dataset}{DeepMind Mathematics Dataset} & English & Synthetic & Math & Algorithmically generated math problems\\

\href{https://huggingface.co/datasets/nvidia/OpenMathInstruct-1}{OpenMathInstruct-1} & English & Human + Model & Math & 1.8M math problems with code-interpreter solutions\\

\href{https://huggingface.co/datasets/microsoft/orca-math-word-problems-200k}{Orca-Math Word Problems 200k} & English & Synthetic & Math & 200K synthetic grade school math word problems  \\

\href{https://huggingface.co/datasets/BytedTsinghua-SIA/DAPO-Math-17k}{DAPO-Math-17k} & English & Human + Model & Math & 17K diverse math problems for reasoning \\

\href{https://huggingface.co/datasets/SynthLabsAI/Big-Math-RL-Verified}{Big-Math-RL-Verified} & English & Human + Model & Math & 251K math problems with verifiable answers for RL \\

\href{https://huggingface.co/datasets/BelleGroup/school_math_0.25M}{BELLE-math-zh} & Chinese & Human + Model & Math & Chinese elementary school math problems\\

\href{https://huggingface.co/datasets/ALmonster/MathInstruct-Chinese}{MathInstruct-Chinese} & Chinese & Model & Math & Chinese version of MathInstruct\\

\midrule[0.5pt]

\href{https://huggingface.co/datasets/dzunggg/legal-qa-v1}{Legal-QA-v1} & English & Human & Law & 3.7K legal QA pairs from legal forums \\

\href{https://huggingface.co/datasets/pile-of-law/pile-of-law}{Pile of Law} & English & Human & Law 
& Dataset for legal-domain tasks\\

\href{https://huggingface.co/datasets/theatticusproject/cuad}{CUAD} & English & Human & Law & 26K expert-annotated legal contract QA \\

\href{https://huggingface.co/datasets/coastalchp/ledgar}{LEDGAR} & English & Human & Law & 1.45M expert-labeled legal contract clauses \\

\href{https://huggingface.co/datasets/ShengbinYue/DISC-Law-SFT}{DISC-Law-SFT} & Chinese & Human + Model & Law & Comprehensive instructions for diverse legal tasks \\

\href{https://huggingface.co/datasets/sentence-transformers/law-gpt}{Law-GPT-zh} & Chinese & Human & Law & Legal sentence pairs for embedding models \\

\href{https://huggingface.co/datasets/Dusker/lawyer-llama}{Lawyer LLaMA} & Chinese & Human + Model & Law & Dataset for Chinese legal tasks\\

\midrule[0.5pt]
\bottomrule[1pt]
\end{tabular}
}
\end{table}

\subsubsection{Code Domain Instruction Fine-Tuning Datasets}

Instruction fine-tuning datasets in the code domain are curated to improve a language model’s ability to understand, generate, and reason about source code across various programming languages and tasks~\citep{muennighoff2023octopack}. In the context of FedLLM, such datasets are particularly valuable for enabling privacy-preserving applications like on-device programming assistants, secure code generation within enterprise environments, and personalized developer support. Given that source code repositories often contain proprietary algorithms, sensitive business logic, or embedded credentials, federated fine-tuning offers a compelling alternative to centralized training on raw code, allowing organizations to harness LLM capabilities without compromising code confidentiality.

Representative datasets include CodeAlpaca~\citep{chaudhary2023code} and Code Instructions 120k Alpaca\footnote{https://huggingface.co/datasets/iamtarun/code$\_$instructions$\_$120k$\_$alpaca}, which extend the Alpaca instruction format to software engineering tasks such as debugging, function generation, and refactoring. CodeContests~\citep{li2022competition} focuses on competitive programming tasks and provides instruction–response pairs related to algorithmic problem solving. CommitPackFT~\citep{muennighoff2023octopack} offers commit-message generation tasks based on source code diffs, reflecting real-world software maintenance scenarios. ToolBench~\citep{qin2023toolllm} is designed to help models learn tool-augmented code generation through instruction-following examples.

Several large-scale pretraining and fine-tuning datasets are also widely used for code instruction alignment. CodeParrot\footnote{https://huggingface.co/datasets/codeparrot/codeparrot-clean} and The Stack v2 Dedup~\citep{kocetkov2022stack} provide diverse and deduplicated code corpora across multiple programming languages, while CodeSearchNet~\citep{husain2019codesearchnet} and CodeXGLUE~\citep{lu2021codexglue} support retrieval, summarization, and translation tasks with instruction-style prompts. CodeForces-CoTs~\citep{penedo2025codeforces} incorporates chain-of-thought annotations for programming tasks, supporting more explainable and step-wise code generation. These code-specific datasets play a vital role in developing FedLLMs that can support privacy-sensitive, domain-specific programming environments. 
A comparative summary of these code-related datasets is presented in Table~\ref{tab_finetune_dataset2}.

\subsubsection{Math Domain Instruction Fine-Tuning Datasets}

Instruction fine-tuning datasets in the math domain are developed to enhance a language model’s proficiency in mathematical reasoning, symbolic computation, and step-by-step problem solving~\citep{tang2024mathscale}. These datasets are particularly valuable in FedLLM scenarios such as personalized math tutoring, intelligent educational platforms, and localized STEM applications, where sensitive student information or institution-specific curricular content must remain on-device. Fine-tuning LLMs in the math domain poses unique challenges, as it requires not only a strong grasp of language but also precise logical inference and numerical accuracy—skills essential for generating correct and interpretable mathematical solutions.

For English-language datasets, GSM8K~\citep{cobbe2021gsm8k} is a widely used benchmark for grade-school math word problems with detailed rationales. CoT-GSM8K\footnote{https://huggingface.co/datasets/Dahoas/cot$\_$gsm8k} augments it with chain-of-thought explanations to support intermediate reasoning steps. Datasets such as MathInstruct~\citep{yue2023mammoth}, MetaMathQA~\citep{yu2023metamath}, OpenR1-Math-220k~\citep{allal2025open}, and Orca-Math Word Problems 200k~\citep{mitra2024orcamath} provide high-quality, instruction-based math problems covering arithmetic, algebra, and word problem solving. The Competition MATH Benchmark~\citep{math} and DeepMind Mathematics\footnote{https://huggingface.co/datasets/di-zhang-fdu/DeepMind$\_$Mathematics$\_$QA} Dataset offer more advanced, competition-style problems suitable for evaluating formal mathematical reasoning. Recent datasets like DAPO-Math-17k~\citep{yu2025dapoopensourcellmreinforcement}, Big-Math-RL-Verified~\citep{albalak2025bigmathlargescalehighqualitymath}, and OpenMathInstruct-1~\citep{toshniwal2024openmath} integrate reinforcement signals, verifiable proofs, or multi-step derivations, further pushing the boundaries of instruction-aligned mathematical LLMs.

For Chinese-language instruction tuning datasets, BELLE-math-zh\footnote{https://huggingface.co/datasets/frankminors123/belle-math-zh} and MathInstruct-Chinese\footnote{https://huggingface.co/datasets/ALmonster/MathInstruct-Chinese} provide diverse mathematical problems adapted to Chinese curricula and linguistic structures. 
These datasets facilitate the training and evaluation of FedLLMs in multilingual educational contexts, supporting privacy-preserving and culturally contextualized math assistance. A comparative summary of these math-focused instruction tuning datasets is provided in Table~\ref{tab_finetune_dataset2}.

\subsubsection{Legal Domain Instruction Fine-Tuning Datasets}

Instruction fine-tuning datasets in the legal domain are designed to improve a language model’s capability in legal reasoning, contract analysis, statute interpretation, and other tasks that require deep, domain-specific understanding of legal language and structure~\citep{yue2023disc}. In the context of FedLLM, these datasets are particularly important for enabling decentralized legal assistance systems, confidential contract review, and on-device compliance monitoring—applications where legal data is often sensitive, jurisdiction-bound, and subject to strict confidentiality constraints. As centralized training on legal documents is frequently infeasible due to regulatory and privacy concerns, federated fine-tuning offers a promising approach to leveraging LLMs in legal settings without compromising data security or legal integrity.

For English-language instruction tuning, Legal-QA-v1\footnote{https://huggingface.co/datasets/dzunggg/legal-qa-v1} provides question–answer pairs covering legal concepts and procedures across various subfields of law. Pile of Law~\citep{gao2020pile} is a large-scale corpus of U.S. legal documents—including court opinions, contracts, and regulations—that supports open-ended legal instruction tuning. CUAD~\citep{hendrycks2021cuad} focuses on contract understanding, with annotated question–answer pairs tailored for clause extraction and risk analysis. LEDGAR\footnote{https://huggingface.co/datasets/coastalchp/ledgar} contains a large set of contractual clauses categorized into fine-grained legal functions, useful for classification and retrieval tasks in instruction-based formats.

For Chinese-language legal modeling, DISC-Law-SFT~\citep{yue2023disc} offers supervised instruction–response pairs across a wide spectrum of Chinese legal domains, including civil law, criminal law, and administrative law. Law-GPT-zh\footnote{https://huggingface.co/datasets/sentence-transformers/law-gpt} consolidates multiple sources of Chinese legal texts into an instruction tuning format to support legal consultation and statutory reasoning. Lawyer LLaMA~\citep{huang2023lawyer} further augments legal dialogue capabilities with simulated lawyer–client conversations, making it well-suited for on-device legal assistants in federated environments. A comparative overview of these law-specific instruction tuning datasets is provided in Table~\ref{tab_finetune_dataset2}.

\subsection{Evaluation Benchmarks}

\subsubsection{General Evaluation Benchmarks}

General-purpose evaluation benchmarks play a critical role in systematically assessing the instruction-following ability, reasoning competence, and overall robustness of LLMs across a wide range of tasks and domains~\citep{lou2024large}. In the context of FedLLM, such benchmarks are especially valuable for evaluating model generalization under heterogeneous data distributions, identifying robustness gaps in decentralized training settings, and facilitating consistent comparisons between personalized and globally aggregated models. Existing benchmarks can be broadly categorized into four groups: (i) general reasoning and instruction-following, (ii) robustness, alignment, and meta-evaluation, (iii) multilingual and Chinese-specific benchmarks, and (iv) long-context understanding. These benchmarks provide a foundation for rigorous and reproducible evaluation of FedLLM across heterogeneous and dynamic environments.

\noindent\textbf{(i) General reasoning and instruction-following.}
This category includes evaluation benchmarks such as MMLU~\citep{hendrycks2020measuring}, BIG-bench~\citep{srivastava2022beyond}, DROP~\citep{dua2019drop}, CRASS~\citep{frohberg2021crass}, and ARC~\citep{clark2018think}, which assess multitask knowledge, discrete reasoning, and science question answering. AGIEval~\citep{zhong2023agieval}, M3Exam~\citep{zhang2023m3exam}, and SCIBENCH~\citep{wang2023scibench} extend evaluation to standardized exams and college-level math, physics, and chemistry. Instruction-following quality is addressed by Vicuna Evaluation\footnote{https://github.com/lm-sys/vicuna-blog-eval}, MT-Bench~\citep{zheng2023judging}, AlpacaEval~\citep{dubois2023alpacafarm}, Chatbot Arena~\citep{zheng2023judging}, and PandaLM~\citep{wang2023pandalm}. Datasets like HellaSwag~\citep{zellers2019hellaswag} and TruthfulQA~\citep{lin2021truthfulqa} focus on commonsense inference and factual accuracy, respectively.

\begin{table}[!t]
\footnotesize
\centering

\caption{Overview of \textbf{general} evaluation benchmarks. 
Abbreviations: EM (Exact Match), OOD (Out-of-Distribution), NER (Named Entity Recognition), ICL (In-Context Learning), RE (Relation Extraction), MRR (Mean Reciprocal Rank), ASR (Adversarial Success Rate).}

\label{tab_eva_benchmark}
\resizebox{1\linewidth}{!}{
\begin{tabular}{lccccccc}
\toprule[1pt]
\midrule[0.5pt]

\textbf{Benchmark} & \textbf{Domain} & \textbf{Evaluation Objective} & \textbf{Main Evaluation Criteria} \\ 
\midrule[0.5pt]

\href{https://huggingface.co/datasets/cais/mmlu}{MMLU} & Gen. & Multitask understanding (57 subjects) & Accuracy \\
\href{https://huggingface.co/datasets/tasksource/bigbench}{BIG-bench} & Gen. & Advanced reasoning capabilities & EM, ROUGE, and Accuracy \\

\href{https://huggingface.co/datasets/ucinlp/drop}{DROP} & Gen. & Discrete reasoning over paragraphs & EM \& F1 \\

\href{https://github.com/apergo-ai/CRASS-data-set}{CRASS} & Gen. & Counterfactual reasoning & Accuracy \\

\href{https://huggingface.co/datasets/allenai/ai2_arc}{ARC} & Gen. & Grade-school science reasoning & Accuracy \\

\href{https://github.com/ruixiangcui/AGIEval}{AGIEval} & Gen. & Performance on human-centric exams & Accuracy \\

\href{https://github.com/DAMO-NLP-SG/M3Exam}{M3Exam} & Gen. & Multilingual, multimodal, multilevel reasoning & Accuracy \\
\href{https://github.com/mandyyyyii/scibench}{SCIBENCH} & Gen. & College-level scientific problem-solving & Accuracy \& Error Attribution \\

\href{https://github.com/lm-sys/vicuna-blog-eval}{Vicuna Evaluation} & Gen. & Instruction-following quality & Human/GPT-4 Preference \\

\href{https://github.com/lm-sys/FastChat/tree/main/fastchat/llm_judge}{MT‑Bench} & Gen. & Multi-turn conversation \& instruction-following & GPT-4 Win Rate \\
\href{https://github.com/tatsu-lab/alpaca_eval}{AlpacaEval} & Gen. & Instruction-following (LLM auto-annotation) & Win Rate\\
\href{https://github.com/lm-sys/FastChat/blob/main/docs/arena.md}{Chatbot Arena} & Gen. & Human-preference & Elo Score \\

\href{https://github.com/WeOpenML/PandaLM}{PandaLM} & Gen. & Instruction-following \& hyperparameter impact & PandaLM Win Rate \\ 
\href{https://rowanzellers.com/hellaswag/}{HellaSwag} & Gen. & Commonsense inference (continuation selection) & Accuracy \\

\href{https://huggingface.co/datasets/truthfulqa/truthful_qa}{TruthfulQA} & Gen. & Truthfulness \& falsehood avoidance & Truthfulness Rate \& Accuracy \\
\href{https://huggingface.co/datasets/derek-thomas/ScienceQA}{ScienceQA} & Gen. & Multimodal scientific reasoning \& explanation & Accuracy \& Explanation Quality \\
\href{https://github.com/FranxYao/chain-of-thought-hub}{CoT Hub} & Gen. & Multi-step reasoning via CoT & Accuracy\\
\href{https://github.com/DeepReasoning/NeuLR}{NeuLR} & Gen. & Deductive, inductive, and abductive reasoning & Accuracy \\

\href{https://github.com/Arvid-pku/ALCUNA}{ALCUNA} & Gen. & Novel knowledge comprehension \& reasoning & Accuracy \\
\href{https://proceedings.neurips.cc/paper_files/paper/2023/file/f64e55d03e2fe61aa4114e49cb654acb-Paper-Datasets_and_Benchmarks.pdf}{LMExamQA} & Gen. & Knowledge recall, understanding, and analysis & Accuracy \\
\href{https://github.com/minjechoi/SOCKET}{SocKET} & Gen. & Social knowledge understanding & Accuracy \\
\href{https://github.com/JoeyHou/choice-75}{Choice-75} & Gen. & Decision reasoning (scripted scenarios) & Accuracy \\
\href{https://github.com/stanford-crfm/helm}{HELM} & Gen. & Holistic evaluation (multi-metric scenarios) & Composite Score \\
\href{https://github.com/bentoml/openllm-benchmark}{OpenLLM} & Gen. & Open-style reasoning (multi-benchmark) & Normalized Accuracy \\
\midrule[0.5pt]

\href{https://github.com/lifan-yuan/OOD_NLP}{BOSS} & Gen. & OOD robustness & OOD Accuracy Drop \\
\href{https://github.com/YangLinyi/GLUE-X}{GLUE‑X} & Gen. & OOD robustness & OOD Accuracy Drop \\
\href{https://github.com/microsoft/promptbench}{PromptBench} & Gen. & Robustness \& prompt-engineering & ASR \& Accuracy \\
\href{https://github.com/facebookresearch/dynabench}{DynaBench} & Gen. & Robustness (dynamic human-in-loop) & Error Rate \\
\href{https://github.com/THU-KEG/KoLA}{KoLA} & Gen. & Evolving world knowledge (19 tasks) & Self-Contrast Calibration \\
\href{https://github.com/Abbey4799/CELLO}{CELLO} & Gen. & Complex instruction-following & Accuracy \& BLEU  \\
\href{https://github.com/UMass-Meta-LLM-Eval/llm_eval}{LLMEval} & Gen. & Meta-evaluation of LLM evaluators & Meta-evaluator Agreement \\
\midrule[0.5pt]

\href{https://github.com/MikeGu721/XiezhiBenchmark}{Xiezhi} & Gen. & Holistic domain knowledge (516 disciplines) & MRR \\
\href{https://github.com/hkust-nlp/ceval}{C‑Eval} & Gen. & Chinese domain knowledge \& reasoning & Accuracy \\
\href{https://github.com/LianjiaTech/BELLE/tree/main/eval}{BELLE-eval} & Gen. & Chinese instruction-following \& multi-skill & GPT-4 Win Rate \\
\href{https://github.com/CLUEbenchmark/SuperCLUE}{SuperCLUE} & Gen. & Chinese instruction-following (human-aligned) & GPT-4 Win Rate \& Accuracy \\ 
\href{https://github.com/tjunlp-lab/M3KE}{M3KE} & Gen. & Chinese knowledge (71 disciplines, 4 levels) & Accuracy \\
\href{https://github.com/ictnlp/BayLing}{BayLing-80} & Gen. & Cross-lingual \& conversational capabilities & GPT-4 Win Rate \\

\href{https://huggingface.co/datasets/Besteasy/MMCU}{MMCU} & Gen. & Multitask Chinese understanding & Accuracy \\

\href{https://github.com/jizijing/C-CLUE}{C‑CLUE} & Gen. & Classical Chinese NER \& RE & Accuracy, Recall, and F1\\

\midrule[0.5pt]
\bottomrule[1pt]
\end{tabular}
}
\vspace{-5mm}
\end{table}

\begin{table}[!t]
\footnotesize
\centering

\caption{Overview of \textbf{general long-context (LC)} evaluation benchmarks.}

\label{tab_eva_benchmark_lc}
\resizebox{1\linewidth}{!}{
\begin{tabular}{lccccccc}
\toprule[1pt]
\midrule[0.5pt]

\textbf{Benchmark} & \textbf{Domain} & \textbf{Evaluation Objective} & \textbf{Main Evaluation Criteria} \\ 
\midrule[0.5pt]

\href{https://huggingface.co/datasets/THUDM/LongBench}{LongBench} & Gen. (LC) & Bilingual long-context understanding & Accuracy, F1, and ROUGE \\

\href{https://github.com/OpenLMLab/LEval}{L-Eval} & Gen. (LC) & Long-context reasoning (up to 60k) & Accuracy \& Win Rate \\

\href{https://github.com/OpenBMB/InfiniteBench}{InfiniteBench} & Gen. (LC) & Long-context understanding (up to 2M) & Accuracy \& ROUGE \\

\href{https://github.com/Hambaobao/Marathon}{Marathon} & Gen. (LC) & Long-context understanding (multi-domain) & Accuracy \\

\href{https://github.com/DachengLi1/LongChat}{LongEval} & Gen. (LC) & Long-context retrieval & Accuracy \\

\href{https://github.com/booydar/babilong}{BABILong} & Gen. (LC) & Long-context reasoning (haystack) & Accuracy \\

\href{https://github.com/Phospheneser/DetectiveQA}{DetectiveQA} & Gen. (LC) & Long-context reasoning & Accuracy \& GPT-4 Judge Score \\

\href{https://github.com/marzenakrp/nocha}{NoCha} & Gen. (LC) & Narrative comprehension & Accuracy \\

\href{https://github.com/MozerWang/Loong}{Loong} & Gen. (LC) & Document reasoning \& extended-context QA & GPT-4 Judge Score \\

\href{https://github.com/Zhihan72/TCELongBench}{TCELongBench} & Gen. (LC) & Temporal reasoning over long narratives & Accuracy\\

\href{https://github.com/ameliadai/DENIAHL}{DENIAHL} & Gen. (LC) & Context feature influence & ROUGE \& EM \\

\href{https://github.com/xiaowu0162/LongMemEval}{LongMemEval} & Gen. (LC) & Long-term interactive memory & EM \\

\href{https://github.com/QZH-777/longrag}{Long2RAG} & Gen. (LC) & Long-text RAG capabilities & BLEU, ROUGE, and EM \\

\href{https://github.com/ZetangForward/L-CITEEVAL}{L-CiteEval} & Gen. (LC) & Citation \& evidence utilization & ROUGE \& Accuracy \\

\href{https://github.com/sheldonwu0327/lif-bench-2024}{LIFBENCH} & Gen. (LC) & Long-context instruction-following & Accuracy, ROUGE, and Pass@k\\

\href{https://huggingface.co/datasets/lz1bytedance/LongReason}{LongReason} & Gen. (LC) & Long-context reasoning & Accuracy \& ROUGE-L\\

\href{https://github.com/RUCAIBox/BAMBOO}{BAMBOO} & Gen. (LC) & Long-text modeling (diverse tasks) & Accuracy, F1, and ROUGE\\

\href{https://github.com/dmis-lab/ETHIC}{ETHIC} & Gen. (LC) & Long-context understanding & F1, Win Rate, and EM \\

\href{https://github.com/bigai-nlco/LooGLE}{LooGLE} & Gen. (LC) & Long-context understanding and reasoning & GPT-4 Judge Score \& Auto Metric\\

\href{https://princeton-nlp.github.io/HELMET/}{HELMET} & Gen. (LC) & Multi-skill long-context capabilities & Task Metrics \& Human Evaluation \\

\href{https://huggingface.co/datasets/megagonlabs/holobench}{HoloBench} & Gen. (LC) & Holistic reasoning (database-style text) & Accuracy\\

\href{https://github.com/google-deepmind/loft}{LOFT} & Gen. (LC) & Long-context reasoning & Accuracy, EM, and Recall \\

\href{https://github.com/infinigence/LVEval}{Lv-Eval} & Gen. (LC) & Long-context comprehension (up to 256k) & Keyword Recall \& F1 \\

\href{https://github.com/launchnlp/ManyICLBench}{ManyICLBench} & Gen. (LC) & Many-shot ICL capabilities & Accuracy \\

\href{https://github.com/tau-nlp/zero_scrolls}{ZeroSCROLLS} & Gen. (LC) & Zero-shot inference capabilities & ROUGE \& F1 \\

\href{https://github.com/TIGER-AI-Lab/LongICLBench}{LongICLBench} & Gen. (LC) & Long-context ICL & Accuracy, ROUGE-L, and Pass@k \\

\href{https://github.com/ai-forever/LIBRA}{LIBRA} & Gen. (LC) & Russian long-context understanding & Accuracy, Recall, and F1 \\

\midrule[0.5pt]
\bottomrule[1pt]
\end{tabular}
}
\vspace{-5mm}
\end{table}

Several benchmarks target more specialized reasoning abilities: ScienceQA~\citep{lu2022learn} evaluates multimodal scientific question answering; Chain-of-Thought Hub~\citep{fu2023chain} focuses on step-wise reasoning; NeuLR~\citep{xu2023large} benchmarks deductive, inductive, and abductive reasoning; ALCUNA~\citep{yin2023alcuna} assesses generalization to novel knowledge; LMExamQA~\citep{bai2023benchmarking} tests models on recall, understanding, and analysis across over academic questions. Benchmarks like SocKET~\citep{choi2023llms} and Choice-75~\citep{hou2023choice} address social knowledge and decision-making in scripted scenarios, respectively. Broad-scoped platforms such as HELM~\citep{liang2022holistic} and OpenLLM\footnote{https://huggingface.co/spaces/open-llm-leaderboard/open$\_$llm$\_$leaderboard} integrate multiple datasets and offer normalized aggregate scores across diverse metrics and tasks.

\noindent\textbf{(ii) Robustness, alignment, and meta-evaluation.}
To simulate the variability and noise of federated environments, benchmarks such as BOSS~\citep{yuan2023revisiting}, GLUE-X~\citep{yang2022glue}, PromptBench~\citep{zhu2023promptbench}, and DynaBench~\citep{kiela2021dynabench} test model robustness to input distribution shifts, prompt perturbations, and adversarial examples. 
KoLA~\citep{yu2023kola} emphasizes world knowledge calibration, while CELLO~\citep{he2024can} further tests compliance under real-world constraints. LLMEval~\citep{zhang2023wider} functions as a meta-evaluation benchmark to assess the consistency and reliability of LLM-based evaluators, an important consideration when deploying automated evaluation in federated settings.

\noindent\textbf{(iii) Multilingual and Chinese-specific benchmarks.}
Benchmarks such as Xiezhi~\citep{gu2024xiezhi}, C-Eval~\citep{huang2023c}, BELLE-eval~\citep{ji2023exploring}, SuperCLUE~\citep{xu2023superclue}, M3KE~\citep{liu2023m3ke}, BayLing-80~\citep{zhang2023bayling}, MMCU~\citep{zeng2023measuring}, and C-CLUE\footnote{https://github.com/jizijing/C-CLUE} provide a diverse evaluation landscape for Chinese and multilingual LLMs. These benchmarks span topics ranging from academic disciplines to sociocultural reasoning, using metrics including GPT-4 preference scoring, multitask accuracy, F1 scores, and normalized Elo rankings.

\noindent\textbf{(iv) Long-context understanding.}
Long-context reasoning is critical for FedLLM applications involving document-intensive tasks, extended dialogue, and memory retention. Benchmarks such as LongBench~\citep{bai2023longbench}, L-Eval~\citep{an2023eval}, InfiniteBench~\citep{zhang2024inftybench}, Marathon~\citep{zhang2023marathon}, LongEval~\citep{li2023long}, and BABILong~\citep{kuratov2024babilong} form the backbone of this category, measuring comprehension, retrieval, and reasoning over contexts up to 2 million tokens. Further specialized benchmarks include: 1) Narrative and temporal reasoning: DetectiveQA~\citep{xu2024detectiveqa}, NoCha~\citep{karpinska2024one}, Loong~\citep{wang2024leave} and TCELongBench~\citep{zhang2024analyzing} focus on long-range narrative understanding and event-based temporal reasoning in complex textual sequences. 2) Retrieval and memory evaluation: DENIAHL~\citep{dai2024deniahl}, LongMemEval~\citep{wu2024longmemeval}, Long2RAG~\citep{qi2024long}, and L-CiteEval~\citep{tang2024citeeval}  assess in-context feature sensitivity, long-term memory utilization, retrieval grounding, and the model’s ability to incorporate external citations. 

3) Instruction-following and generation under long input: LIFBENCH~\citep{wu2024lifbench}, LongReason~\citep{ling2025longreason}, BAMBOO~\citep{dong2023bamboo}, ETHIC~\citep{lee2024ethic}, and LooGLE~\citep{li2023loogle} evaluate multi-criteria instruction-following accuracy, response stability, and coherence under extended prompts. HELMET~\citep{yen2024helmet} further integrates summarization, retrieval, and reasoning in unified evaluation pipelines, supporting both automatic and human metrics. 4) Database-style and structured input reasoning: HoloBench~\citep{maekawa2024holistic} and LOFT~\citep{lee2024can} benchmark the ability to perform complex reasoning over structured or database-like inputs, including table querying, execution accuracy, and factual consistency, especially in scenarios where retrieval-augmented methods are substituted by long-context modeling.

5) Scaling and generalization with longer context:
Lv-Eval~\citep{yuan2024lv}, ManyICLBench~\citep{zou2024retrieval}, ZeroSCROLLS~\citep{shaham2023zeroscrolls}, and LongICLBench~\citep{li2024long} examine the scalability of in-context learning and multi-task alignment as input length increases, making them valuable tools for analyzing the performance ceiling of long-context FedLLMs.
6) Cross-lingual long-context evaluation:
LIBRA~\citep{churin2024long} tests instruction-following and long-form coherence in Russian, contributing to the evaluation of multilingual long-context capabilities.
These long-context benchmarks are crucial for validating the scalability and persistence of FedLLMs, especially in environments requiring private document analysis, extended user sessions, or continual knowledge tracking.
Table~\ref{tab_eva_benchmark} and Table~\ref{tab_eva_benchmark_lc} summarize the general evaluation benchmarks discussed above.

\subsubsection{Financial Domain Evaluation Benchmarks}

Finance-specific evaluation benchmarks are crucial for assessing the domain alignment, factual accuracy, and reasoning capabilities of LLMs in high-stakes financial contexts~\citep{li2023large}. In federated settings, where sensitive data from banks, asset managers, and regulatory bodies cannot be centralized due to confidentiality and compliance constraints, these benchmarks serve as vital tools for evaluating model performance under decentralized, privacy-preserving, and task-diverse conditions. Effective financial evaluation must encompass a broad range of tasks—including open-book question answering, multi-step quantitative reasoning, and document classification—while also accounting for linguistic nuances, regulatory requirements, and the structural complexity of financial texts. Such benchmarks are indispensable for ensuring the robustness and reliability of FedLLM in real-world financial applications.

\noindent\textbf{(i) Multi-task and agent-style financial evaluation.} Several benchmarks offer broad-spectrum evaluation across multiple financial tasks, aligning well with FedLLM’s need to support diverse client use cases (e.g., auditing, compliance, trading). FinBen~\citep{xie2024finben} evaluates holistic financial reasoning over 24 tasks using automatic metrics, retrieval-augmented generation accuracy, and expert human judgment. PIXIU~\citep{xie2023pixiu}, FLUE~\citep{shah2022flue}, and BBT-CFLEB~\citep{lu2023bbt} benchmark LLMs on multi-task setups including sentiment classification, QA, event detection, and stock prediction. CFinBench~\citep{nie2024cfinbench} and SuperCLUEFin~\citep{xu2024superclue} extend this paradigm to Chinese financial scenarios, assessing models across regulatory knowledge, certification preparation, and real-world task instructions using multi-type question formats.
ICE‑PIXIU~\citep{xie2023pixiu} further supports bilingual Chinese–English evaluation, suitable for cross-regional FedLLM deployments. FLARE-ES~\citep{zhang2024dolares} enables bilingual Spanish–English testing to evaluate cross-lingual transfer and domain-specific reasoning.

\noindent\textbf{(ii) Task-specific capability evaluation.} 
Fine-grained benchmarks evaluate specific financial NLP capabilities that are critical for real-world FedLLM deployment, including document question answering, information extraction, and numerical reasoning.
FinanceBench~\citep{islam2023financebench} targets open-book question answering using real-world, company-related financial documents, with a focus on factual correctness and evidence alignment—a critical requirement for FedLLM deployed in compliance, auditing, and enterprise document analysis scenarios. FiNER‑ORD~\citep{shah2023finer} and FinRED~\citep{sharma2022finred} evaluate financial Named Entity Recognition and relation extraction from news and earnings transcripts—key for localized data processing on devices with limited connectivity. FinQA~\citep{chen2021finqa} and BizBench~\citep{koncel2023bizbench} benchmark multi-step numerical and quantitative reasoning, involving both tabular data and executable financial logic, useful in portfolio analysis, valuation, and budgeting applications. EconLogicQA~\citep{quan2024econlogicqa} addresses sequential economic reasoning over multi-event scenarios. In the Chinese financial domain, FinEval~\citep{zhang2023fineval} and  CFBenchmark~\citep{lei2023cfbenchmark} assess task-specific instruction following across topics such as taxation, accounting, and investment strategy.
Hirano~\citep{hirano2024construction} enables financial language understanding evaluation in Japanese, while MultiFin~\citep{jorgensen2023multifin} focuses on multilingual topic classification, useful in federated settings with cross-border clients.

\noindent\textbf{(iii) Long-context financial reasoning.} Many practical financial tasks involve extended documents such as annual reports, investor briefs, and regulatory filings. Long-context benchmarks are critical to evaluate whether FedLLM can perform document-level comprehension and multi-step derivation under memory and privacy constraints.
DocFinQA~\citep{reddy2024docfinqa} simulates multi-step numerical reasoning over financial reports, measuring exact match, F1, and reasoning traceability. FinTextQA~\citep{chen2024fintextqa} further tests open-ended QA over long textual contexts using BLEU and ROUGE for generation quality. These benchmarks reflect realistic federated scenarios, such as on-device due diligence support or local regulatory interpretation, where global models must adapt to long-form content without accessing raw documents.
A summary of these finance-specific evaluation benchmarks, including their evaluation objectives and corresponding metrics, is presented in Table~\ref{tab_eva_benchmark_domain}.

\begin{table}[!t]
\footnotesize
\centering
\renewcommand{\arraystretch}{1.0}  

\caption{Overview of evaluation benchmarks in the \textbf{financial} and \textbf{medical} domains.}

\label{tab_eva_benchmark_domain}
\resizebox{1\linewidth}{!}{
\begin{tabular}{lccccccc}
\toprule[1pt]
\midrule[0.5pt]

\textbf{Benchmark} & \textbf{Domain} & \textbf{Evaluation Objective} & \textbf{Main Evaluation Criteria} \\ 
\midrule[0.5pt]

\href{https://huggingface.co/datasets/TheFinAI/finben-finer-ord}{FinBen} & Financial & Holistic financial capabilities & Auto metrics \& Human eval \\

\href{https://github.com/The-FinAI/PIXIU}{PIXIU} & Financial & Financial NLP tasks & Accuracy \& F1 \\

\href{https://salt-nlp.github.io/FLANG/}{FLUE} & Financial & Financial language understanding & Accuracy, F1, nDCG, and MRR \\

\href{https://github.com/ssymmetry/BBT-FinCUGE-Applications}{BBT-CFLEB} & Financial & Chinese financial understanding and generation & ROUGE, F1, and Accuracy \\

\href{https://cfinbench.github.io}{CFinBench} & Financial & Chinese financial knowledge & Accuracy \\

\href{https://www.cluebenchmarks.com/}{SuperCLUEFin} & Financial & Chinese financial assistant capabilities & Win Rate \& Accuracy \\

\href{https://github.com/YY0649/ICE-PIXIU}{ICE‑PIXIU} & Financial & Bilingual (Chinese–English) financial reasoning & Accuracy, F1, and ROUGE \\

\href{https://github.com/The-FinAI/PIXIU}{FLARE‑ES} & Financial & Bilingual (Spanish–English) financial reasoning & Accuracy, F1, and ROUGE \\

\href{https://github.com/patronus-ai/financebench}{FinanceBench} & Financial & Financial open-book QA (real-world) & Factual Correctness \\

\href{https://huggingface.co/datasets/gtfintechlab/finer-ord}{FiNER‑ORD} & Financial & Financial NER & F1, Precision, and Recall \\

\href{https://github.com/soummyaah/FinRED/}{FinRED} & Financial & Financial RE (news, transcripts) & F1, Precision, and Recall \\

\href{https://github.com/czyssrs/FinQA}{FinQA} & Financial & Numerical reasoning over financial reports & Accuracy \\

\href{https://huggingface.co/datasets/kensho/bizbench}{BizBench} & Financial & Quantitative reasoning on realistic problems & Accuracy \& F1 \\

\href{https://huggingface.co/datasets/yinzhu-quan/econ_logic_qa}{EconLogicQA} & Financial & Economic sequential reasoning & Accuracy \\

\href{https://github.com/SUFE-AIFLM-Lab/FinEval}{FinEval} & Financial & Chinese financial knowledge and reasoning & Accuracy \\

\href{https://github.com/TongjiFinLab/CFBenchmark}{CFBenchmark} & Financial & Chinese financial assistant capabilities & LLM Judge Score \& Accuracy\\

\href{https://github.com/pfnet-research/japanese-lm-fin-harness}{Hirano} & Financial & Japanese financial understanding & Accuracy \\

\href{https://huggingface.co/datasets/awinml/MultiFin}{MultiFin} & Financial & Multilingual financial topic classification & F1 \& Accuracy \\

\href{https://huggingface.co/datasets/kensho/DocFinQA}{DocFinQA} & Financial (LC) & Long-context financial document reasoning & Accuracy \\

\href{https://huggingface.co/datasets/GPS-Lab/FinTextQA}{FinTextQA} & Financial (LC) & Long-form financial QA & ROUGE \& LLM Judge Score\\

\midrule

\href{https://github.com/CBLUEbenchmark/CBLUE}{CBLUE} & Medical & Chinese biomedical understanding & Accuracy \& F1 \\

\href{https://huggingface.co/datasets/Juvenilecris/CHIP2023-PromptCBLUE-peft}{PromptCBLUE} & Medical & Chinese medical knowledge (prompt-based) & Accuracy \& F1 \\

\href{https://huggingface.co/datasets/FreedomIntelligence/CMB}{CMB} & Medical & Chinese medical knowledge & Accuracy \& LLM Judge Score \\

\href{https://huggingface.co/datasets/FreedomIntelligence/huatuo26M-testdatasets}{HuaTuo26M} & Medical 
& Chinese medical knowledge and QA & ROUGE \& and LLM Judge Score\\

\href{https://huggingface.co/datasets/fzkuji/CMExam}{CMExam} & Medical & Chinese medical licensing exam & Accuracy, BLEU, and ROUGE \\

\href{https://huggingface.co/datasets/openlifescienceai/multimedqa}{MultiMedQA} & Medical & Clinical knowledge and open-ended QA & Accuracy \& Human Evaluation \\

\href{https://github.com/CMKRG/QiZhenGPT}{QiZhenGPT} & Medical & Drug indication identification & Accuracy \\

\href{https://github.com/knowlab/MedExQA}{MedExQA} & Medical & Medical knowledge and explanation generation & Accuracy \& LLM judge Score \\

\href{https://github.com/HanjieChen/ChallengeClinicalQA}{JAMA and Medbullets} & Medical & Challenging clinical QA & Accuracy \& Human Evaluation\\

\href{https://huggingface.co/datasets/TsinghuaC3I/MedXpertQA}{MedXpertQA} & Medical & Expert-level medical reasoning (multi-modal) & Accuracy\\

\href{https://github.com/Medical-AI-Learning/MedJourney}{MedJourney} & Medical 
& Full clinical patient journey tasks & Accuracy, ROUGE, BLEU, and F1 \\

\href{https://github.com/gersteinlab/medagents-benchmark}{MedAgentsBench} & Medical & Complex multi-step clinical reasoning & Accuracy\\

\href{https://github.com/kbressem/LongHealth}{LongHealth} & Medical (LC) & QA over long-form clinical documents & Accuracy \\

\href{https://github.com/JOHNNY-fans/MedOdyssey}{MedOdyssey} & Medical (LC) & Long-context medical understanding & Accuracy, Recall, and F1\\
\midrule[0.5pt]
\bottomrule[1pt]
\end{tabular}
}
\end{table}

\subsubsection{Medical Domain Evaluation Benchmarks}

Medical-specific evaluation benchmarks are essential for assessing the performance of LLMs in healthcare applications, where accuracy, safety, and domain-specific understanding are critical~\citep{thirunavukarasu2023large}. In the context of FedLLM, these benchmarks are particularly important for evaluating models deployed in privacy-sensitive environments, such as hospital intranets, personal health monitoring devices, and clinical support systems, where patient data cannot be centralized due to regulatory and ethical constraints. To reflect the complexity of medical reasoning and language comprehension under such federated conditions, medical benchmarks cover a wide range of evaluation targets, including diagnostic reasoning, medical question answering, clinical document understanding, and guideline adherence.


\noindent\textbf{(i) Medical knowledge and QA evaluation.} This category focuses on assessing LLMs' general and specialized medical knowledge through structured question-answering tasks.
CBLUE~\citep{zhang2021cblue} and PromptCBLUE\footnote{https://github.com/michael-wzhu/PromptCBLUE} benchmark Chinese biomedical NLP tasks across information extraction, document classification, and QA, with the latter emphasizing prompt-based generation.
CMB~\citep{wang2023cmb}, HuaTuo26M~\citep{li2023huatuo}, and CMExam~\citep{liu2023benchmarking} evaluate models on Chinese medical exam-style QA, clinical diagnosis tasks, and real-world query understanding. These benchmarks are well-suited for evaluating FedLLM tailored to localized clinical documentation and public health information.
MultiMedQA~\citep{singhal2023large} serves as a high-quality English benchmark encompassing multiple-choice and open-ended medical QA, with expert-based evaluation of factuality, helpfulness, and safety—three pillars critical for deploying FedLLM in patient-facing applications.
QiZhenGPT\footnote{https://github.com/CMKRG/QiZhenGPT} provides an annotation-based evaluation of drug indication extraction from natural language prompts, useful for drug interaction checking at the edge.
MedExQA~\citep{kim2024medexqa} expands QA evaluation to underrepresented medical specialties, while JAMA and Medbullets~\citep{chen2025benchmarking} challenge models with high-difficulty US medical exam-style questions and demand strong explanation quality.

\noindent\textbf{(ii) Clinical reasoning and care process modeling.} In practice, many medical applications require multi-step diagnostic reasoning, care pathway modeling, and treatment planning, especially in multi-agent or longitudinal clinical scenarios. MedXpertQA~\citep{zuo2025medxpertqa} tests both textual and multimodal reasoning, simulating specialist-level question answering across medical images and textual reports. MedJourney~\citep{wu2024medjourney} evaluates LLMs across the full patient care pipeline, from chief complaint triage to follow-up guidance, with both task-specific and human expert evaluation.
MedAgentsBench~\citep{tang2025medagentsbench} explicitly focuses on multi-turn clinical planning, assessing the model’s ability to generate coherent and correct diagnostic and treatment steps over multiple interactions—an ideal setup for privacy-preserving agent-based FedLLM deployment.

\noindent\textbf{(iii) Long-context clinical understanding.} Federated medical applications often involve lengthy patient histories, radiology reports, or guideline documents. Benchmarks in this group evaluate the ability of LLMs to perform robust QA and inference over long-form inputs.
LongHealth~\citep{adams2024longhealth} focuses on QA over long clinical narratives, measuring exact match and F1 accuracy.
MedOdyssey~\citep{fan2024medodyssey} targets long-context understanding across clinical specialties, incorporating ROUGE, EM, and human preference scoring to assess consistency and informativeness over extended reasoning chains.
A comprehensive summary of these medical evaluation benchmarks is provided in Table~\ref{tab_eva_benchmark_domain}.

\subsubsection{Code Domain Evaluation Benchmarks}

Code-specific evaluation benchmarks are essential for assessing the ability of FedLLM to understand, generate, and reason over programming logic across multiple languages and task types~\citep{jiang2024survey}. In federated settings, code-related applications are commonly deployed in heterogeneous environments, including personalized coding assistants, on-device tutoring systems, and localized software development workflows. These scenarios place unique demands on LLMs: they must follow fine-grained instructions, ensure functional correctness, and maintain high reliability—all while operating under the resource constraints and privacy requirements that are characteristic of FedLLM deployment. Benchmarks in this domain typically evaluate code synthesis, completion, bug fixing, and multi-turn problem solving.


\noindent\textbf{(i) Basic code generation and functional reasoning.}
This group of benchmarks focuses on evaluating models' ability to generate syntactically correct and semantically valid code from natural language instructions. HumanEval~\citep{chen2021evaluating} and MBPP~\citep{austin2021program} are foundational benchmarks evaluating one-shot Python function generation, judged via execution-based metrics like pass@k. APPS~\citep{hendrycks2021measuring2} and DS-1000~\citep{lai2023ds} extend this to more complex and real-world tasks, with DS-1000 focusing on data science scenarios across multiple Python libraries.
CodeXGLUE~\citep{lu2021codexglue} offers a broad suite of tasks—spanning code summarization, translation, and generation—making it suitable for assessing FedLLM that may specialize in different sub-tasks across clients.
CruxEval~\citep{gu2024cruxeval} further emphasizes logical reasoning and error-free execution in high-stakes contexts.
ODEX~\citep{wang2022execution} introduces cross-lingual code generation, evaluating the model’s ability to translate natural language into code in four different programming languages, a valuable benchmark for multilingual FedLLM deployment.
These datasets are particularly useful for evaluating client-level specialization in federated setups, where users might work with domain-specific codebases or tools.

\noindent\textbf{(ii) Multi-turn synthesis and structural understanding.} Modern code development involves iterative and structural logic, making multi-turn and component-aware benchmarks crucial.
MTPB~\citep{nijkamp2022codegen} focuses on multi-turn program synthesis, assessing the ability to generate partial, compositional subprograms in sequence.
ClassEval~\citep{du2023classeval} evaluates whether LLMs can generate coherent Python classes, including method dependencies and variable interactions—reflecting real-world object-oriented programming needs.
BigCodeBench~\citep{zhuo2024bigcodebench} challenges models with complex, multi-functional instruction following, and rich code behavior evaluation, using criteria like test case accuracy and branch coverage.
HumanEvalPack~\citep{muennighoff2023octopack} extends HumanEval’s principles to six languages and multiple code-related subtasks, enabling multilingual federated evaluation.
BIRD~\citep{li2023can} adds a structured dimension by testing text-to-SQL generation for database querying, emphasizing schema-aware reasoning and executable correctness—an increasingly relevant task in enterprise AI agents and personal data querying under private data constraints.
Such benchmarks are particularly relevant for FedLLM deployed in collaborative or enterprise environments, where partial programs must be incrementally refined by agents or users with limited compute.

\noindent\textbf{(iii) Long-context code understanding and retrieval.}
Real-world software development often requires reasoning over extended codebases or repositories, posing a challenge for memory-constrained clients. RepoQA~\citep{liu2024repoqa} and LongCodeArena~\citep{bogomolov2024long} evaluate models' comprehension and retrieval-augmented reasoning over entire repositories or large code documents. They measure not only token-level accuracy but also structural coherence and retrieval efficacy. 
These benchmarks offer valuable insights into the performance of long-context-aware FedLLM on realistic software engineering tasks such as bug fixing, documentation generation, and legacy code comprehension—particularly under constraints imposed by limited local computational resources.

Together, these code-specific benchmarks provide a comprehensive foundation for assessing the capabilities and limitations of FedLLM in coding applications. A detailed overview of benchmark objectives and evaluation metrics is provided in Table~\ref{tab_eva_benchmark_domain2}.

\begin{table}[!t]
\footnotesize
\centering

\caption{Overview of evaluation benchmarks in the \textbf{code}, \textbf{math}, and \textbf{law} domains.}

\label{tab_eva_benchmark_domain2}
\resizebox{1\linewidth}{!}{
\begin{tabular}{lccccccc}
\toprule[1pt]
\midrule[0.5pt]

\textbf{Benchmark} & \textbf{Domain} & \textbf{Evaluation Objective} & \textbf{Main Evaluation Criteria} \\ 
\midrule[0.5pt]

\href{https://huggingface.co/datasets/openai/openai_humaneval}{HumanEval} & Code & Code generation and algorithmic reasoning & Pass@k \\

\href{https://huggingface.co/datasets/Muennighoff/mbpp}{MBPP} & Code & Basic Python code generation & Pass@k \\

\href{https://huggingface.co/datasets/codeparrot/apps}{APPS} & Code & Coding challenge competence & Pass@k \\

\href{https://huggingface.co/datasets/xlangai/DS-1000}{DS-1000} & Code & Data science code generation & Accuracy \& Pass@k \\

\href{https://github.com/microsoft/CodeXGLUE}{CodeXGLUE} & Code & Code understanding and generation & Accuracy, BLEU, and MRR \\

\href{https://github.com/facebookresearch/cruxeval}{CruxEval} & Code & Code reasoning, understanding, and  execution & Pass@k \\

\href{https://huggingface.co/datasets/neulab/odex}{ODEX} & Code & Code generation and execution & Pass@k \\

\href{https://github.com/salesforce/CodeGen}{MTPB} & Code & Multi-turn program synthesis & Pass@k \\

\href{https://huggingface.co/datasets/FudanSELab/ClassEval}{ClassEval} & Code & Class-level code generation & Pass@k\\

\href{https://huggingface.co/datasets/bigcode/bigcodebench}{BigCodeBench} & Code & Complex and practical code generation & Pass@k\\

\href{https://huggingface.co/datasets/bigcode/humanevalpack}{HumanEvalPack} & Code & Multilingual code tasks & Pass@k\\

\href{https://github.com/bird-bench/mini_dev}{BIRD} & Code 
& Database-grounded text-to-SQL & Execution Accuracy \& F1 \\

\href{https://github.com/evalplus/repoqa}{RepoQA} & Code (LC) & Long-context code understanding & Pass Rate \\

\href{https://github.com/JetBrains-Research/lca-baselines}{LongCodeArena} & Code (LC) & Long-context code tasks & Pass@k, Accuracy, and EM \\

\midrule

\href{https://huggingface.co/datasets/openai/gsm8k}{GSM8K} & Math & Grade school math reasoning & Accuracy \\

\href{https://huggingface.co/datasets/hendrycks/competition_math}{Competition MATH} & Math & Competition math problem-solving & Accuracy \\

\href{https://huggingface.co/datasets/MathOdyssey/MathOdyssey}{MathOdyssey} & Math & High-level math problem-solving (Olympiad) & Accuracy \\

\href{https://github.com/open-compass/MathBench}{MathBench} & Math & Comprehensive and hierarchical math evaluation & Accuracy \& Circular Evaluation\\

\href{https://github.com/YilunZhou/champ-dataset}{CHAMP} & Math & Fine-grained math reasoning & Accuracy \\

\href{https://huggingface.co/datasets/allenai/lila}{LILA} & Math & Meta-benchmark for math reasoning & Accuracy \& Pass@k \\

\href{https://github.com/openai/miniF2F}{MiniF2F} & Math 
& Formal math reasoning (Olympiad) & Proof Accuracy \\

\href{https://huggingface.co/datasets/hoskinson-center/proofnet}{ProofNet} & Math & Auto-formalization and formal proof generation & Accuracy \& Success Rate \\

\href{https://github.com/google-deepmind/alphageometry}{AlphaGeometry} & Math & Neuro-symbolic reasoning (Euclidean geometry) & Solve Rate \\

\href{https://huggingface.co/datasets/AI4Math/MathVerse}{MathVerse} & Math & Visual math reasoning & Accuracy \& CoT Score \\

\href{https://huggingface.co/datasets/We-Math/We-Math}{We-Math} & Math & Multi-step math reasoning & Accuracy\\

\href{https://huggingface.co/datasets/toloka/u-math}{U-MATH} & Math & Open-ended visual math (university-level) & LLM Judge Score \\

\href{https://github.com/lupantech/PromptPG}{TabMWP} & Math 
& Math reasoning over text and tabular data & Accuracy \\

\href{https://github.com/SalesforceAIResearch/MathHay}{MathHay} & Math (LC) & Long-context math reasoning (multi-step) & Accuracy\\

\midrule

\href{https://huggingface.co/datasets/nguha/legalbench}{LegalBench} & Law & Legal reasoning across six tasks & Accuracy, F1, and ROUGE \\

\href{https://huggingface.co/datasets/coastalcph/lex_glue}{LexGLUE} & Law & Legal language understanding & Accuracy \& F1 \\

\href{https://huggingface.co/datasets/joelniklaus/lextreme}{LEXTREME} & Law & Multilingual/multitask legal understanding & Accuracy \& F1 \\

\href{https://github.com/open-compass/LawBench}{LawBench} & Law & Chinese legal reasoning & Accuracy \& F1 \\

\href{https://github.com/Dai-shen/LAiW}{LAiW} & Law & Chinese legal tasks & Accuracy \& F1 \\

\href{https://huggingface.co/datasets/CSHaitao/LexEval}{LexEval} & Law & Chinese legal understanding and reasoning & Accuracy \\

\href{https://arxiv.org/abs/2412.14556}{CitaLaw} & Law 
& Legal answer generation & MRR, Recall, and ROUGE\\

\href{https://github.com/CSHaitao/LegalAgentBench}{LegalAgentBench} & Law & LLM agent legal task-solving & Success Rate \& LLM Judge Score \\

\href{https://github.com/vr18ub/court_view_generation}{SCALE} & Law (LC) & Multilingual/long-document legal reasoning & ROUGE \& METEOR \\

\midrule[0.5pt]
\bottomrule[1pt]
\end{tabular}
}
\end{table}

\subsubsection{Math Domain Evaluation Benchmarks}

Mathematical reasoning tasks serve as a rigorous benchmark for evaluating the generalization, compositionality, and step-by-step problem-solving capabilities of LLMs~\citep{ahn2024large}. In federated settings, math-specific evaluations are especially valuable for applications such as privacy-preserving intelligent tutoring systems, on-device educational tools, and personalized STEM learning assistants. These tasks present unique challenges for FedLLM, as they require precise multi-step reasoning, symbolic computation, and logical consistency—often under strict memory, computation, and communication constraints. As such, math benchmarks are instrumental in assessing a model's ability to perform structured reasoning in resource-constrained and heterogeneous environments.

\noindent\textbf{(i) Primary math reasoning across educational levels.} Benchmarks in this group assess basic-to-advanced math problem solving and reasoning:
GSM8K~\citep{cobbe2021gsm8k} focuses on grade school arithmetic word problems, serving as a foundation for reasoning evaluation. Competition MATH~\citep{math} extends this to competition-level questions in algebra, geometry, and calculus, emphasizing step-by-step derivation accuracy. MathOdyssey~\citep{fang2024mathodyssey} evaluates reasoning across high school, university, and Olympiad levels, providing a broad difficulty spectrum.
MathBench~\citep{liu2024mathbench} systematically tests both theoretical understanding and practical application across five levels, reflecting multi-tier federated education scenarios.
CHAMP~\citep{mao2024champ} introduces concept and hint annotations, useful for federated tutoring agents that may require step-wise guidance or personalization for struggling learners.
LILA~\citep{mishra2022lila} further expands evaluation to 23 math task types, offering comprehensive insight into the model’s versatility across mathematical formats.

\noindent\textbf{(ii) Formal proof and symbolic reasoning.}
Mathematics often requires formal logic and symbolic structure, which tests a model’s ability to generalize beyond pattern-matching:
MiniF2F~\citep{zheng2021minif2f} and ProofNet~\citep{azerbayev2023proofnet} evaluate formal mathematical reasoning and proof generation, with the latter using Lean 3 as a backend for correctness verification.
AlphaGeometry~\citep{trinh2024solving} blends neural and symbolic reasoning for Euclidean geometry, a domain requiring precise spatial logic and theorem synthesis—especially relevant in expert-centric or research-level FedLLM deployment.

\noindent\textbf{(iii) Visual and diagrammatic mathematical reasoning.}
Many real-world math problems include visual elements (graphs, tables, geometric diagrams), posing multi-modal reasoning challenges:
MathVerse~\citep{zhang2024mathverse} and We-Math~\citep{qiao2024we} evaluate visual reasoning using diagram interpretation, requiring fine-grained attention to layout and symbolic grounding.
U-MATH~\citep{chernyshev2024u} tests open-ended university-level questions involving visual cues, with LLM-assisted expert scoring.
TabMWP~\citep{lu2022dynamic} focuses on text–table joint reasoning, simulating practical applications like report analysis or financial tutoring in federated agents.

\noindent\textbf{(iv) Long-context mathematical reasoning.}
FedLLM deployed on real-world devices often faces scenarios where mathematical problems span multiple steps or documents:
MathHay~\citep{wang2024mathhay} evaluates reasoning across extended input chains, testing memory retention and logic consistency in multi-hop math reasoning—an important benchmark for long-context capabilities in private and offline educational settings.

Together, these benchmarks form a robust and diverse suite for evaluating the mathematical competency of FedLLM under different input formats, difficulty levels, and reasoning demands. They are especially vital for personalized STEM learning assistants and edge-based automated math tutoring. 
A comprehensive summary of these math evaluation benchmarks is provided in Table~\ref{tab_eva_benchmark_domain2}.

\subsubsection{Legal Domain Evaluation Benchmarks}

Legal AI applications impose stringent requirements on factual accuracy, contextual understanding, and logical consistency—making legal benchmarks particularly important for evaluating FedLLM designed for use in areas such as smart justice, personalized legal consultation, and privacy-preserving regulatory compliance. These benchmarks are designed to assess model capabilities across a range of legal reasoning tasks, including legal text comprehension, argument analysis, statutory interpretation, and multi-document synthesis. Moreover, they often adopt multilingual and long-context formats that reflect the real-world complexity and heterogeneity of legal documents—challenges that are amplified in federated  settings~\citep{chen2024survey}.


\noindent\textbf{(i) Foundational and legal-specific reasoning.} Benchmarks in this category assess general-purpose legal understanding, including statute interpretation, legal judgment tasks, and document comprehension:
LegalBench~\citep{guha2023legalbench} provides a suite of six legal reasoning tasks such as rule application and contract understanding, evaluating both correctness and rule-consistency.
LexGLUE~\citep{chalkidis2021lexglue} includes classic legal NLP tasks such as case classification, contract QA, and legal entailment, making it suitable for evaluating LLMs in general legal understanding scenarios.
LEXTREME~\citep{niklaus2023lextreme} expands this evaluation to 24 languages and 18 tasks, making it a critical multilingual benchmark for assessing FedLLM’s cross-lingual legal proficiency in global regulatory contexts.

\noindent\textbf{(ii) Chinese legal language understanding.}
Given the importance of regional legal systems, several benchmarks target Chinese legal capabilities, aligning well with privacy-sensitive applications deployed in Chinese jurisdictions:
LawBench~\citep{fei2023lawbench} evaluates Chinese legal LLMs across three cognitive levels—retention, understanding, and application—via 20 legal tasks.
LAiW~\citep{dai2023laiw} offers a fine-grained assessment framework covering fundamental to advanced legal challenges.
LexEval~\citep{li2024lexeval} structures evaluation around a taxonomy of legal cognitive abilities, including memory, reasoning, and application. These benchmarks support federated personalization and regional adaptation of legal agents.

\noindent\textbf{(iii) Legal citation and generation with formal grounding.}
Some legal tasks require not only correct answers but also proper justification grounded in laws and precedents:
CitaLaw~\citep{zhang2024citalaw} focuses on generating legally sound responses with accurate citations to statutes and precedent cases. It uses metrics like syllogism alignment and legal consistency, making it particularly relevant for FedLLM agents operating in jurisdictions with citation and traceability requirements.
\noindent\textbf{(iv) Legal agents and dynamic task-solving.} FedLLM may increasingly support legal assistants or decision-support agents that require multi-step, tool-integrated reasoning:
LegalAgentBench~\citep{li2024legalagentbench} provides a novel benchmark for evaluating LLM agents on complex, multi-turn legal scenarios. 
It incorporates intermediate progress tracking and task success scoring, which are useful for evaluating FedLLM performance in decentralized and asynchronous workflows.

\noindent\textbf{(v) Long-context and multilingual legal reasoning.} Legal documents are often lengthy, hierarchical, and span multiple statutes or precedents:
SCALE~\citep{rasiah2023scale} is designed to evaluate LLMs on long-context legal documents, legal multilingualism, and cross-document reasoning. Its inclusion of multi-task legal scenarios and code-level legal analysis makes it especially relevant for edge-deployed legal assistants in enterprise or government use cases.
Together, these benchmarks provide a rich and diverse framework for evaluating FedLLM in the legal domain—where privacy, jurisdictional customization, long-context understanding, and reasoning fidelity are all critical to real-world deployment. 
A comprehensive summary of these legal evaluation benchmarks is provided in Table~\ref{tab_eva_benchmark_domain2}.

\section{Application}\label{sec_application}



\subsection{FedLLM for Recommendation Systems}

Recommendation systems are pivotal across domains such as e-commerce, content streaming, and personalized advertising~\citep{ko2022survey}. Traditional approaches often rely on centralized data collection, raising significant privacy concerns, particularly when handling sensitive user interactions and preferences. Federated fine-tuning offers a promising alternative by enabling collaborative learning across distributed clients while preserving data privacy.

\citet{zhao2024llm} propose FELLRec, a federated framework for LLM-based recommendation, to tackle the challenges of client performance imbalance and high resource costs. Specifically, FELLRec employs dynamic parameter aggregation and adaptive learning speeds to ensure balanced performance across clients. Additionally, it selectively retains sensitive LLM layers on the client side while offloading other layers to the server, effectively preserving privacy and optimizing resource usage.
Similarly, \citet{yuan2024fellas} introduce FELLAS, a federated sequential recommendation framework that leverages LLMs as external services to enhance sequential recommendation. FELLAS enriches item embeddings via LLM-assisted textual representation while ensuring privacy protection through $d_{x}$-privacy-compliant sequence perturbation. 
Beyond privacy protection, FL also facilitates reinforcement learning from human feedback for LLM-based recommendation systems. \citet{wu2024client} propose FedBis and FedBiscuit, two frameworks designed to enable privacy-preserving federated RLHF. FedBis collaboratively trains a binary selector to filter sensitive preference data, while FedBiscuit clusters clients to train multiple selectors, ensuring better alignment with human preferences while maintaining privacy.
Another framework, GPT-FedRec, proposed by \citet{zeng2024federated}, integrates ChatGPT with a hybrid Retrieval-Augmented Generation (RAG) mechanism to address data sparsity, heterogeneity, and LLM-specific challenges in federated recommendation. GPT-FedRec employs hybrid retrieval techniques to extract user patterns and item features, then refines recommendations through LLM-generated prompts. By leveraging RAG, the framework effectively mitigates hallucination in LLM-generated content and enhances the overall recommendation quality.

In summary, as LLMs become increasingly integrated into recommendation systems, federated fine-tuning has emerged as a powerful method to fully leverage the capabilities of LLMs while ensuring user data privacy. These advancements demonstrate the potential of FedLLM to support high-quality, privacy-aware recommendation in real-world applications.

\subsection{FedLLM for Biomedical Research}

In the biomedical domain, direct data transmission and centralized model fine-tuning pose significant risks to user and patient privacy. Federated fine-tuning provides a privacy-preserving paradigm that enables collaborative model adaptation across decentralized medical datasets without exposing sensitive information. 
\citet{ali2025fine} explore the use of various FL techniques to fine-tune time-series LLMs on electrocardiogram and impedance cardiography data, enabling privacy-preserving physiological signal analysis.
\citet{naseer2024probing} conduct a systematic investigation into the application of federated PEFT strategies for fine-tuning vision transformers in medical image classification tasks. 
\citet{puppala2024scan} present a FL-based GPT chatbot designed for personalized healthcare information retrieval. The system aggregates and curates information from diverse sources while ensuring privacy and security through decentralized training. Users receive real-time, personalized insights via an intuitive interface, supported by advanced text parsing, metadata enrichment, and question-answering capabilities. This framework marks a key advancement in patient-centric AI applications.

\citet{sarwar2025fedmentalcare} introduces FedMentalCare, a privacy-preserving framework that integrates FL and LoRA to fine-tune LLMs for mental health analysis. Their study explores the impact of client data volumes and model architectures (e.g., MobileBERT, MiniLM) in FL settings, ensuring scalability, data security, and computational efficiency. 
\citet{liu2024fedfms} propose FedFMS, which introduces federated foundation models for medical image segmentation. It addresses privacy challenges in medical imaging by enabling federated training without centralized data sharing. 
\citet{wang2024fedkim} introduce FEDKIM, a federated knowledge injection framework for scaling medical foundation models. It leverages lightweight local models to extract private knowledge and integrates it into a centralized model using an adaptive multitask multimodal mixture of experts module, enabling efficient cross-institution knowledge transfer.
\citet{dai2025fedata} propose FedATA, a self-supervised FL framework for medical image segmentation, integrating masked self-distillation with adaptive attention to enhance pre-training and fine-tuning on unlabeled and limited-annotation data. Unlike traditional masked image modeling, FedATA uses latent representations as targets instead of pixels, improving feature learning. Additionally, its adaptive attention aggregation with personalized FL captures institution-specific representations, boosting model generalization and local fine-tuning performance.

In summary, LLMs have become increasingly influential in biomedical research. However, due to the highly sensitive nature of user data in this domain, federated fine-tuning has emerged as a pivotal approach for enabling large-scale AI applications in biomedicine and healthcare while ensuring data privacy and security.

\subsection{FedLLM for Finance}

The financial sector heavily relies on data-driven models for risk assessment, fraud detection, algorithmic trading, and personalized financial services. However, financial data is often highly sensitive, heavily regulated, and distributed across multiple institutions, making centralized model training infeasible due to privacy concerns and compliance constraints. Federated fine-tuning presents a promising solution by enabling collaborative learning across financial institutions without exposing raw data.

\citet{ye2024openfedllm} introduce OpenFedLLM, a federated fine-tuning framework designed to train LLMs on decentralized private data while ensuring data privacy. In the financial domain, FL-tuned LLMs significantly outperform locally trained models and even surpass GPT-4, demonstrating the potential of FL to enhance LLM performance without compromising sensitive financial data. This study underscores the value of federated fine-tuning in leveraging distributed financial data to develop more accurate, robust, and privacy-preserving LLMs for financial applications.
\citet{shabani2024harnessing} explore the use of FL for fine-tuning LLMs in finance, enhancing efficiency and privacy while addressing data scarcity and distribution challenges. Their findings show that FL achieves performance comparable to centralized fine-tuning with significantly lower computational costs and training time, making it ideal for resource-constrained environments. This approach preserves data privacy while enabling the development of more accurate and robust financial LLMs.
Similarly, \citet{zeng2024fine} investigate fine-tuning financial LLMs using LoRA and deploying them on edge devices, demonstrating FL’s potential to improve both model efficiency and performance in financial applications. Their study highlights significant gains in reasoning capabilities and cost-effectiveness, offering valuable insights into leveraging FL and LLMs for private and vertically specialized financial domains.

In summary, federated fine-tuning plays a crucial role in the financial sector by enabling collaborative model training across institutions while ensuring strict data privacy compliance. This approach allows financial organizations to leverage vast, decentralized datasets for fine-tuning, improving model accuracy  without exposing sensitive financial information. As financial markets grow more complex and globally interconnected, FedLLM presents a scalable and secure pathway toward next-generation AI-driven financial infrastructure.


\section{Open Challenges and Future Directions}\label{sec_outlook}

\paragraph{Model Security of FedLLM.} As federated fine-tuning gains momentum, ensuring model security has become a critical concern. In FedLLM, pre-trained models, whether proprietary or open-source, must be transmitted to distributed clients for local fine-tuning, inherently increasing the risk of intellectual property (IP) leakage and system vulnerabilities. Model security in this context involves two key aspects: protecting the IP of high-value models and ensuring the secure deployment of open-source models on edge devices.

First, the financial and strategic value of pre-trained LLMs makes IP protection in FedLLM deployment especially pressing. For example, training models like Gemini Ultra~\citep{gemma} and GPT-4~\citep{GPT4} is estimated to cost $\$191$ million and $\$78$ million, respectively. These models are typically developed by commercial entities under strict licensing and infrastructure control. However, in FedLLM settings, where the full model is often shared with clients in a white-box fashion, it becomes feasible for malicious participants to reverse-engineer or clone the model. This undermines the original developers’ competitive advantage and deters participation from commercial model providers. Addressing this challenge requires the development of model watermarking~\citep{pan2024markllm}, encrypted model delivery, or inference-obfuscation protocols that allow clients to fine-tune and use the model without revealing sensitive architectural or parameter details.

Second, while open-source LLMs (e.g., DeepSeek~\citep{bi2024deepseek}, Qwen~\citep{bai2023qwen}) are widely adopted in FedLLM due to their accessibility and flexibility, they present new vectors for security threats in federated deployments. In practice, most clients, especially those with limited machine learning or systems expertise, may lack the capabilities to deploy these models securely. 
For instance, as reported by the technology news outlet AIbase\footnote{https://www.aibase.com/news/15909}, frameworks like Ollama have been found to expose users to data leakage and unauthorized resource usage due to insecure default configurations. In a federated setup, such vulnerabilities are amplified: a single compromised client can leak locally fine-tuned training data, or propagate adversarial backdoors to the global model. The consequences are particularly severe in sensitive domains like healthcare and finance, where breaches may result in the disclosure of protected health information (PHI) or proprietary trading strategies.

To mitigate these risks, future FedLLM research should prioritize the integration of secure model deployment practices into the federated fine-tuning pipeline. Techniques such as confidential computing for secure execution on edge devices, encrypted model delivery, and runtime access control should be incorporated to prevent unauthorized access and tampering. By embedding such security mechanisms into the FedLLM lifecycle, both commercial and open-source models can be safeguarded against misuse, thereby promoting broader adoption in high-stakes domains such as healthcare, finance, and critical infrastructure.

\paragraph{LLM and SLM Collaboration.}
A key future direction for FedLLM lies in enabling efficient collaboration between LLMs and small language models (SLMs) to address the performance–privacy–efficiency trade-offs inherent to federated settings. While LLMs offer superior reasoning and multi-modal capabilities, their large size and resource demands make them impractical for direct deployment on edge devices. Conversely, SLMs such as Gemini Nano~\citep{team2023gemini} and Phi-3~\citep{abdin2024phi} provide lightweight alternatives with better deployment efficiency but limited generalization and task transferability.

To reconcile these limitations, emerging FedLLM architectures can adopt a hybrid model paradigm: deploying SLMs at the edge for privacy-sensitive inference and lightweight tasks, while offloading complex reasoning or orchestration to cloud-hosted LLMs. This collaborative strategy not only reduces the communication and computation burden at the client side but also enhances regulatory compliance by ensuring sensitive data never leaves the local device. Within this architecture, edge SLMs can perform initial text generation or instruction parsing, while LLMs handle tool selection, global coordination, or cross-domain alignment.

However, realizing this collaboration raises several open challenges in FedLLM: 1) minimizing latency and bandwidth overhead introduced by frequent SLM–LLM interactions; 2) preserving consistency and alignment between local SLM outputs and global LLM behavior; and 3) dynamically adapting task delegation strategies based on client heterogeneity, model confidence, and task complexity.
Future FedLLM research should design decentralized orchestration protocols for efficient SLM–LLM coordination, and introduce privacy-preserving metadata exchange mechanisms to protect tool usage logs and inference traces. By enabling seamless SLM–LLM collaboration under federated environments, this hybrid architecture can unlock new levels of scalability, efficiency, and privacy in real-world FedLLM deployments.

\paragraph{Multi-Modal FedLLM.}
While existing FedLLM research has primarily focused on text-based tasks, emerging real-world applications increasingly require multi-modal capabilities, such as integrating visual, speech, and sensor modalities~\citep{zhang2024mm}. Large multi-modal models (LMMs), including GPT-4V~\citep{yang2023dawn} and LLaVA~\citep{liu2023visual}, have demonstrated strong performance in centralized settings. However, extending these models to federated environments presents several unresolved challenges.

A primary difficulty lies in modality heterogeneity across clients~\citep{peng2024fedmm, ouyang2023harmony}. Devices may possess different input types—for example, some may only have textual data, while others may hold image–text pairs—leading to modality imbalance, where certain modalities are underrepresented in the training process~\citep{fan2024overcome}. This imbalance can degrade the model’s generalization across modalities. Additionally, achieving cross-modal alignment~\citep{gao2024softclip}—the model’s ability to relate and reason across different modalities—becomes more difficult in federated setups, where paired data (e.g., image–caption pairs) cannot be shared centrally. Moreover, the high computational demands of LMMs further constrain their deployment and fine-tuning on edge devices with limited memory and processing capabilities~\citep{liang2024survey, jin2024efficient}.

To address these challenges, future research should develop modular and flexible tuning frameworks that allow each modality to be fine-tuned independently on client devices. This decoupling enables efficient local adaptation without requiring all modalities to be present on each client. Furthermore, modality-aware aggregation protocols—which weight client contributions based on modality type, data quality, and semantic consistency—can help mitigate imbalance and enhance global model performance. Promising directions also include federated cross-modal contrastive learning~\citep{yu2023multimodal}, which can improve multi-modal alignment without requiring raw data exchange. Finally, to facilitate deployment in edge-centric applications such as smart healthcare, assistive robotics, and wearable systems, it is essential to design lightweight multi-modal architectures through techniques like knowledge distillation~\citep{cai2024llava} or dynamic subnetwork activation~\citep{alam2022fedrolex} that strike a balance between accuracy and resource efficiency.

\paragraph{Continual Learning in FedLLM.}
In dynamic federated environments, client data distributions and task objectives evolve over time, necessitating continual learning capabilities in FedLLM systems~\citep{yoon2021federated, wang2024federated}. Unlike traditional FL settings with fixed tasks and static datasets, real-world deployments require models to incrementally incorporate new knowledge without retraining from scratch. However, continual fine-tuning of LLMs introduces several unique challenges. The sheer size and overparameterization of LLMs make them prone to catastrophic forgetting during incremental updates~\citep{huang2024mitigating}, especially when client participation is sparse or irregular. Moreover, repeated retraining across rounds is computationally expensive and often infeasible on edge devices with limited hardware resources.

To overcome these limitations, future research should investigate parameter-efficient continual learning strategies that enable local knowledge retention while supporting scalable global updates. Techniques such as Elastic Weight Consolidation (EWC)~\citep{kirkpatrick2017overcoming}, PEFT-based modular updates, and rehearsal methods using compressed memory buffers~\citep{tiwari2022gcr} are promising in mitigating forgetting without incurring prohibitive overhead. In addition, the design of lifelong personalization protocols—capable of adapting to each client’s evolving task distribution under Non-IID and intermittent data availability—remains an open research frontier. Developing such protocols requires balancing communication efficiency, privacy preservation, and model stability across heterogeneous learning trajectories.

Ultimately, enabling continual adaptation in FedLLM will be essential for long-term deployment in dynamic real-world scenarios, such as personalized healthcare, evolving legal compliance systems, or lifelong learning assistants. This calls for a shift from round-based static fine-tuning to streaming, task-aware federated adaptation frameworks that can incrementally evolve with users and environments.

\paragraph{Memory-Efficient FedLLM.}

Memory efficiency remains one of the most fundamental and restrictive bottlenecks in the deployment of FedLLM—often resulting in a binary feasibility condition: a client device either meets the memory requirements to participate in training or is entirely excluded~\citep{wu2025breaking}. Unlike other challenges such as communication overhead or data heterogeneity, which degrade performance but still permit participation, memory limitations can preclude participation altogether, particularly for edge devices with constrained hardware capabilities. Although PEFT techniques like LoRA substantially reduce the number of trainable parameters, they fall short of fully mitigating memory pressure. For example, fine-tuning LLaMA2-13B with LoRA still demands over 50 GB of peak memory—an order of magnitude beyond the capacity of most mobile phones, IoT devices, or embedded systems~\citep{xu2024fwdllm, tian2024breaking}.

Addressing this limitation requires innovation at both the algorithmic and system levels. On the algorithmic side, emerging approaches such as dynamic layer-wise adaptation~\citep{pan2024lisa}, quantization-aware PEFT (e.g., QLoRA)~\citep{dettmers2023qlora}, and structured model pruning~\citep{wang2019structured, ma2023llm} offer promising pathways for reducing memory footprints during local fine-tuning. Complementing these are system-level solutions such as gradient checkpointing and accumulation~\citep{gim2022memory}, runtime memory-aware schedulers, and cloud-edge hybrid training architectures with selective computation offloading~\citep{kumar2013survey}, all of which aim to stretch the effective memory capacity of participating devices.
To fully unlock the potential of FedLLM at scale, future research should explore co-designed frameworks that jointly optimize algorithmic efficiency and system-level deployment. Such holistic solutions can harmonize memory, computation, and communication trade-offs in real time—enabling resource-adaptive, privacy-preserving model customization across highly diverse client ecosystems.


Overcoming the memory barrier would expand the pool of eligible participants to include billions of low-memory edge devices that are currently sidelined from federated training. This not only enhances the inclusiveness, representativeness, and scalability of the FedLLM framework, but also opens the door to real-world deployments in settings such as home automation, wearables, and low-power industrial IoT platforms.



\section{Conclusion}\label{sec_conclusion}

To the best of our knowledge, this is the \emph{first} comprehensive survey dedicated to the federated fine-tuning of LLMs. We begin by introducing foundational background knowledge and identifying four core challenges through empirical analysis, which reveal the fundamental limitations that federated fine-tuning must overcome. 
We then review the latest relevant research papers, systematically organizing recent advances in parameter-efficient federated fine-tuning techniques.
These approaches are categorized based on their methodologies, with detailed discussions on how each class of methods addresses the identified challenges.
Furthermore, we present a comprehensive evaluation framework encompassing both fine-tuning datasets and evaluation benchmarks across different domains, offering a holistic framework for assessing FedLLM performance. Beyond methodological contributions, we highlight practical applications of FedLLM across domains.
Finally, we outline promising future directions in this rapidly evolving field. While notable progress has been achieved, several pressing challenges remain open. Addressing these issues will be essential to unlocking the full potential of FedLLM and enabling its widespread deployment in practical, privacy-sensitive applications.

\section*{Acknowledgements}

This work is supported in part by the Science and Technology Development Fund of Macau (0107/2024/RIA2), Joint Science and Technology Research Project with Hong Kong and Macau in Key Areas of Nansha District’s Science and Technology Plan (EF2024-00180-IOTSC), and the Multi-Year Research Grant of University of Macau (MYRG-GRG2023-00211-IOTSC-UMDF and MYRG-GRG2024-00180-IOTSC).

\bibliography{ref}

\begin{thebibliography}{417}
\providecommand{\natexlab}[1]{#1}
\providecommand{\url}[1]{\texttt{#1}}
\expandafter\ifx\csname urlstyle\endcsname\relax
  \providecommand{\doi}[1]{doi: #1}\else
  \providecommand{\doi}{doi: \begingroup \urlstyle{rm}\Url}\fi

\bibitem[Abdin et~al.(2024)Abdin, Aneja, Awadalla, Awadallah, Awan, Bach, Bahree, Bakhtiari, Bao, Behl, et~al.]{abdin2024phi}
Marah Abdin, Jyoti Aneja, Hany Awadalla, Ahmed Awadallah, Ammar~Ahmad Awan, Nguyen Bach, Amit Bahree, Arash Bakhtiari, Jianmin Bao, Harkirat Behl, et~al.
\newblock Phi-3 technical report: A highly capable language model locally on your phone.
\newblock \emph{arXiv preprint arXiv:2404.14219}, 2024.

\bibitem[Adams et~al.(2024)Adams, Busch, Han, Excoffier, Ortala, L{\"o}ser, Aerts, Kather, Truhn, and Bressem]{adams2024longhealth}
Lisa Adams, Felix Busch, Tianyu Han, Jean-Baptiste Excoffier, Matthieu Ortala, Alexander L{\"o}ser, Hugo~JWL Aerts, Jakob~Nikolas Kather, Daniel Truhn, and Keno Bressem.
\newblock Longhealth: A question answering benchmark with long clinical documents.
\newblock \emph{arXiv preprint arXiv:2401.14490}, 2024.

\bibitem[Ahmad et~al.(2023)Ahmad, Liu, Khan, Gan, and Huang]{ahmad2023prompt}
Khwaja~Mutahir Ahmad, Qiao Liu, Abdullah~Aman Khan, Yanglei Gan, and Changhao Huang.
\newblock Prompt-enhanced federated learning for aspect-based sentiment analysis.
\newblock In \emph{2023 International Conference on Intelligent Communication and Computer Engineering (ICICCE)}, pp.\  81--87. IEEE, 2023.

\bibitem[Ahn et~al.(2024)Ahn, Verma, Lou, Liu, Zhang, and Yin]{ahn2024large}
Janice Ahn, Rishu Verma, Renze Lou, Di~Liu, Rui Zhang, and Wenpeng Yin.
\newblock Large language models for mathematical reasoning: Progresses and challenges.
\newblock \emph{arXiv preprint arXiv:2402.00157}, 2024.

\bibitem[Alam et~al.(2022)Alam, Liu, Yan, and Zhang]{alam2022fedrolex}
Samiul Alam, Luyang Liu, Ming Yan, and Mi~Zhang.
\newblock Fedrolex: Model-heterogeneous federated learning with rolling sub-model extraction.
\newblock \emph{Advances in neural information processing systems}, 35:\penalty0 29677--29690, 2022.

\bibitem[Albalak et~al.(2025)Albalak, Phung, Lile, Rafailov, Gandhi, Castricato, Singh, Blagden, Xiang, Mahan, and Haber]{albalak2025bigmathlargescalehighqualitymath}
Alon Albalak, Duy Phung, Nathan Lile, Rafael Rafailov, Kanishk Gandhi, Louis Castricato, Anikait Singh, Chase Blagden, Violet Xiang, Dakota Mahan, and Nick Haber.
\newblock Big-math: A large-scale, high-quality math dataset for reinforcement learning in language models, 2025.
\newblock URL \url{https://arxiv.org/abs/2502.17387}.

\bibitem[Ali et~al.(2025)Ali, Lisle, Moore, Barkouki, Kirkwood, and Brattain]{ali2025fine}
Mahad Ali, Curtis Lisle, Patrick~W Moore, Tammer Barkouki, Brian~J Kirkwood, and Laura~J Brattain.
\newblock Fine-tuning foundation models with federated learning for privacy preserving medical time series forecasting.
\newblock \emph{arXiv preprint arXiv:2502.09744}, 2025.

\bibitem[Allal et~al.(2025)Allal, Tunstall, Lozhkov, Bakouch, Penedo, and Kydlicek]{allal2025open}
Loubna~Ben Allal, Lewis Tunstall, Anton Lozhkov, Elie Bakouch, Guilherme Penedo, and Gabriel Mart{\'\i}n Bl{\'a}zquez~Hynek Kydlicek.
\newblock Open r1: Evaluating llms on uncontaminated math competitions, 2025.

\bibitem[Almanifi et~al.(2023)Almanifi, Chow, Tham, Chuah, and Kanesan]{almanifi2023communication}
Omair Rashed~Abdulwareth Almanifi, Chee-Onn Chow, Mau-Luen Tham, Joon~Huang Chuah, and Jeevan Kanesan.
\newblock Communication and computation efficiency in federated learning: A survey.
\newblock \emph{Internet of Things}, 22:\penalty0 100742, 2023.

\bibitem[An et~al.(2023)An, Gong, Zhong, Zhao, Li, Zhang, Kong, and Qiu]{an2023eval}
Chenxin An, Shansan Gong, Ming Zhong, Xingjian Zhao, Mukai Li, Jun Zhang, Lingpeng Kong, and Xipeng Qiu.
\newblock L-eval: Instituting standardized evaluation for long context language models.
\newblock \emph{arXiv preprint arXiv:2307.11088}, 2023.

\bibitem[Austin et~al.(2021)Austin, Odena, Nye, Bosma, Michalewski, Dohan, Jiang, Cai, Terry, Le, et~al.]{austin2021program}
Jacob Austin, Augustus Odena, Maxwell Nye, Maarten Bosma, Henryk Michalewski, David Dohan, Ellen Jiang, Carrie Cai, Michael Terry, Quoc Le, et~al.
\newblock Program synthesis with large language models.
\newblock \emph{arXiv preprint arXiv:2108.07732}, 2021.

\bibitem[Azerbayev et~al.(2023)Azerbayev, Piotrowski, Schoelkopf, Ayers, Radev, and Avigad]{azerbayev2023proofnet}
Zhangir Azerbayev, Bartosz Piotrowski, Hailey Schoelkopf, Edward~W Ayers, Dragomir Radev, and Jeremy Avigad.
\newblock Proofnet: Autoformalizing and formally proving undergraduate-level mathematics.
\newblock \emph{arXiv preprint arXiv:2302.12433}, 2023.

\bibitem[Babakniya et~al.(2023)Babakniya, Elkordy, Ezzeldin, Liu, Song, El-Khamy, and Avestimehr]{babakniya2023slora}
Sara Babakniya, Ahmed~Roushdy Elkordy, Yahya~H Ezzeldin, Qingfeng Liu, Kee-Bong Song, Mostafa El-Khamy, and Salman Avestimehr.
\newblock Slora: Federated parameter efficient fine-tuning of language models.
\newblock \emph{arXiv preprint arXiv:2308.06522}, 2023.

\bibitem[Bagwe et~al.(2023)Bagwe, Yuan, Pan, and Zhang]{bagwe2023fed}
Gaurav Bagwe, Xiaoyong Yuan, Miao Pan, and Lan Zhang.
\newblock Fed-cprompt: Contrastive prompt for rehearsal-free federated continual learning.
\newblock \emph{arXiv preprint arXiv:2307.04869}, 2023.

\bibitem[Bai et~al.(2024{\natexlab{a}})Bai, Chen, Qian, Yao, and Li]{bai2024federated}
Jiamu Bai, Daoyuan Chen, Bingchen Qian, Liuyi Yao, and Yaliang Li.
\newblock Federated fine-tuning of large language models under heterogeneous tasks and client resources.
\newblock In \emph{The Thirty-eighth Annual Conference on Neural Information Processing Systems}, 2024{\natexlab{a}}.

\bibitem[Bai et~al.(2023{\natexlab{a}})Bai, Bai, Chu, Cui, Dang, Deng, Fan, Ge, Han, Huang, et~al.]{bai2023qwen}
Jinze Bai, Shuai Bai, Yunfei Chu, Zeyu Cui, Kai Dang, Xiaodong Deng, Yang Fan, Wenbin Ge, Yu~Han, Fei Huang, et~al.
\newblock Qwen technical report.
\newblock \emph{arXiv preprint arXiv:2309.16609}, 2023{\natexlab{a}}.

\bibitem[Bai et~al.(2024{\natexlab{b}})Bai, Zhang, Guo, Li, Guo, Hou, Han, and Lu]{bai2024diprompt}
Sikai Bai, Jie Zhang, Song Guo, Shuaicheng Li, Jingcai Guo, Jun Hou, Tao Han, and Xiaocheng Lu.
\newblock Diprompt: Disentangled prompt tuning for multiple latent domain generalization in federated learning.
\newblock In \emph{Proceedings of the IEEE/CVF Conference on Computer Vision and Pattern Recognition}, pp.\  27284--27293, 2024{\natexlab{b}}.

\bibitem[Bai et~al.(2023{\natexlab{b}})Bai, Lv, Zhang, Lyu, Tang, Huang, Du, Liu, Zeng, Hou, et~al.]{bai2023longbench}
Yushi Bai, Xin Lv, Jiajie Zhang, Hongchang Lyu, Jiankai Tang, Zhidian Huang, Zhengxiao Du, Xiao Liu, Aohan Zeng, Lei Hou, et~al.
\newblock Longbench: A bilingual, multitask benchmark for long context understanding.
\newblock \emph{arXiv preprint arXiv:2308.14508}, 2023{\natexlab{b}}.

\bibitem[Bai et~al.(2023{\natexlab{c}})Bai, Ying, Cao, Lv, He, Wang, Yu, Zeng, Xiao, Lyu, et~al.]{bai2023benchmarking}
Yushi Bai, Jiahao Ying, Yixin Cao, Xin Lv, Yuze He, Xiaozhi Wang, Jifan Yu, Kaisheng Zeng, Yijia Xiao, Haozhe Lyu, et~al.
\newblock Benchmarking foundation models with language-model-as-an-examiner.
\newblock \emph{Advances in Neural Information Processing Systems}, 36:\penalty0 78142--78167, 2023{\natexlab{c}}.

\bibitem[Bao et~al.(2023)Bao, Chen, Xiao, Ren, Wu, Zhong, Peng, Huang, and Wei]{bao2023disc}
Zhijie Bao, Wei Chen, Shengze Xiao, Kuang Ren, Jiaao Wu, Cheng Zhong, Jiajie Peng, Xuanjing Huang, and Zhongyu Wei.
\newblock Disc-medllm: Bridging general large language models and real-world medical consultation.
\newblock \emph{arXiv preprint arXiv:2308.14346}, 2023.

\bibitem[Bi et~al.(2024)Bi, Chen, Chen, Chen, Dai, Deng, Ding, Dong, Du, Fu, et~al.]{bi2024deepseek}
Xiao Bi, Deli Chen, Guanting Chen, Shanhuang Chen, Damai Dai, Chengqi Deng, Honghui Ding, Kai Dong, Qiushi Du, Zhe Fu, et~al.
\newblock Deepseek llm: Scaling open-source language models with longtermism.
\newblock \emph{arXiv preprint arXiv:2401.02954}, 2024.

\bibitem[Bian et~al.(2024)Bian, Wang, Zhang, and Xu]{bian2024lora}
Jieming Bian, Lei Wang, Letian Zhang, and Jie Xu.
\newblock Lora-fair: Federated lora fine-tuning with aggregation and initialization refinement.
\newblock \emph{arXiv preprint arXiv:2411.14961}, 2024.

\bibitem[Bogomolov et~al.(2024)Bogomolov, Eliseeva, Galimzyanov, Glukhov, Shapkin, Tigina, Golubev, Kovrigin, van Deursen, Izadi, et~al.]{bogomolov2024long}
Egor Bogomolov, Aleksandra Eliseeva, Timur Galimzyanov, Evgeniy Glukhov, Anton Shapkin, Maria Tigina, Yaroslav Golubev, Alexander Kovrigin, Arie van Deursen, Maliheh Izadi, et~al.
\newblock Long code arena: a set of benchmarks for long-context code models.
\newblock \emph{arXiv preprint arXiv:2406.11612}, 2024.

\bibitem[Bu et~al.(2022)Bu, Wang, Zha, and Karypis]{bu2022differentially}
Zhiqi Bu, Yu-Xiang Wang, Sheng Zha, and George Karypis.
\newblock Differentially private bias-term only fine-tuning of foundation models.
\newblock 2022.

\bibitem[Byrd \& Polychroniadou(2020)Byrd and Polychroniadou]{byrd2020differentially}
David Byrd and Antigoni Polychroniadou.
\newblock Differentially private secure multi-party computation for federated learning in financial applications.
\newblock In \emph{Proceedings of the first ACM international conference on AI in finance}, pp.\  1--9, 2020.

\bibitem[Byun \& Lee(2024)Byun and Lee]{byun2024towards}
Yuji Byun and Jaeho Lee.
\newblock Towards federated low-rank adaptation of language models with rank heterogeneity.
\newblock \emph{arXiv preprint arXiv:2406.17477}, 2024.

\bibitem[Cai et~al.(2023{\natexlab{a}})Cai, Wu, Wang, Lin, and Xu]{cai2023efficient}
Dongqi Cai, Yaozong Wu, Shangguang Wang, Felix~Xiaozhu Lin, and Mengwei Xu.
\newblock Efficient federated learning for modern nlp.
\newblock In \emph{Proceedings of the 29th Annual International Conference on Mobile Computing and Networking}, pp.\  1--16, 2023{\natexlab{a}}.

\bibitem[Cai et~al.(2023{\natexlab{b}})Cai, Wu, Yuan, Wang, Lin, and Xu]{cai2023towards}
Dongqi Cai, Yaozong Wu, Haitao Yuan, Shangguang Wang, Felix~Xiaozhu Lin, and Mengwei Xu.
\newblock Towards practical few-shot federated nlp.
\newblock In \emph{Proceedings of the 3rd Workshop on Machine Learning and Systems}, pp.\  42--48, 2023{\natexlab{b}}.

\bibitem[Cai et~al.(2024{\natexlab{a}})Cai, Jiang, Wang, Tang, Kim, and Huang]{cai2024survey}
Weilin Cai, Juyong Jiang, Fan Wang, Jing Tang, Sunghun Kim, and Jiayi Huang.
\newblock A survey on mixture of experts.
\newblock \emph{arXiv preprint arXiv:2407.06204}, 2024{\natexlab{a}}.

\bibitem[Cai et~al.(2024{\natexlab{b}})Cai, Zhang, He, He, Tong, Gan, Wang, and Bai]{cai2024llava}
Yuxuan Cai, Jiangning Zhang, Haoyang He, Xinwei He, Ao~Tong, Zhenye Gan, Chengjie Wang, and Xiang Bai.
\newblock Llava-kd: A framework of distilling multimodal large language models.
\newblock \emph{arXiv preprint arXiv:2410.16236}, 2024{\natexlab{b}}.

\bibitem[Chai et~al.(2024)Chai, Wang, Yang, Zhang, Chen, and Yang]{chai2024survey}
Di~Chai, Leye Wang, Liu Yang, Junxue Zhang, Kai Chen, and Qiang Yang.
\newblock A survey for federated learning evaluations: Goals and measures.
\newblock \emph{IEEE Transactions on Knowledge and Data Engineering}, 2024.

\bibitem[Chalkidis et~al.(2021)Chalkidis, Jana, Hartung, Bommarito, Androutsopoulos, Katz, and Aletras]{chalkidis2021lexglue}
Ilias Chalkidis, Abhik Jana, Dirk Hartung, Michael Bommarito, Ion Androutsopoulos, Daniel~Martin Katz, and Nikolaos Aletras.
\newblock Lexglue: A benchmark dataset for legal language understanding in english.
\newblock \emph{arXiv preprint arXiv:2110.00976}, 2021.

\bibitem[Chaudhary(2023)]{chaudhary2023code}
Sahil Chaudhary.
\newblock Code alpaca: An instruction-following llama model for code generation, 2023.

\bibitem[Che et~al.(2023)Che, Liu, Zhou, Ren, Zhou, Sheng, Dai, and Dou]{che2023federated}
Tianshi Che, Ji~Liu, Yang Zhou, Jiaxiang Ren, Jiwen Zhou, Victor~S Sheng, Huaiyu Dai, and Dejing Dou.
\newblock Federated learning of large language models with parameter-efficient prompt tuning and adaptive optimization.
\newblock \emph{arXiv preprint arXiv:2310.15080}, 2023.

\bibitem[Chen et~al.(2025)Chen, Fang, Singla, and Dredze]{chen2025benchmarking}
Hanjie Chen, Zhouxiang Fang, Yash Singla, and Mark Dredze.
\newblock Benchmarking large language models on answering and explaining challenging medical questions.
\newblock In \emph{Proceedings of the 2025 Conference of the Nations of the Americas Chapter of the Association for Computational Linguistics: Human Language Technologies (Volume 1: Long Papers)}, pp.\  3563--3599, 2025.

\bibitem[Chen et~al.(2024{\natexlab{a}})Chen, Zhang, Krompass, Gu, and Tresp]{chen2024feddat}
Haokun Chen, Yao Zhang, Denis Krompass, Jindong Gu, and Volker Tresp.
\newblock Feddat: An approach for foundation model finetuning in multi-modal heterogeneous federated learning.
\newblock In \emph{Proceedings of the AAAI Conference on Artificial Intelligence}, volume~38, pp.\  11285--11293, 2024{\natexlab{a}}.

\bibitem[Chen et~al.(2024{\natexlab{b}})Chen, Zhou, Hua, Loh, Chen, Li, Zhu, and Liang]{chen2024fintextqa}
Jian Chen, Peilin Zhou, Yining Hua, Yingxin Loh, Kehui Chen, Ziyuan Li, Bing Zhu, and Junwei Liang.
\newblock Fintextqa: A dataset for long-form financial question answering.
\newblock \emph{arXiv preprint arXiv:2405.09980}, 2024{\natexlab{b}}.

\bibitem[Chen et~al.(2024{\natexlab{c}})Chen, Yan, Liu, Zhang, Xiong, and Yu]{chen2024federated}
Jingxue Chen, Hang Yan, Zhiyuan Liu, Min Zhang, Hu~Xiong, and Shui Yu.
\newblock When federated learning meets privacy-preserving computation.
\newblock \emph{ACM Computing Surveys}, 56\penalty0 (12):\penalty0 1--36, 2024{\natexlab{c}}.

\bibitem[Chen et~al.(2021{\natexlab{a}})Chen, Tworek, Jun, Yuan, Pinto, Kaplan, Edwards, Burda, Joseph, Brockman, et~al.]{chen2021evaluating}
Mark Chen, Jerry Tworek, Heewoo Jun, Qiming Yuan, Henrique Ponde De~Oliveira Pinto, Jared Kaplan, Harri Edwards, Yuri Burda, Nicholas Joseph, Greg Brockman, et~al.
\newblock Evaluating large language models trained on code.
\newblock \emph{arXiv preprint arXiv:2107.03374}, 2021{\natexlab{a}}.

\bibitem[Chen et~al.(2017)Chen, Zhang, Sharma, Yi, and Hsieh]{chen2017zoo}
Pin-Yu Chen, Huan Zhang, Yash Sharma, Jinfeng Yi, and Cho-Jui Hsieh.
\newblock Zoo: Zeroth order optimization based black-box attacks to deep neural networks without training substitute models.
\newblock In \emph{Proceedings of the 10th ACM workshop on artificial intelligence and security}, pp.\  15--26, 2017.

\bibitem[Chen et~al.(2023{\natexlab{a}})Chen, Long, Shen, and Jiang]{chen2023prompt}
Shengchao Chen, Guodong Long, Tao Shen, and Jing Jiang.
\newblock Prompt federated learning for weather forecasting: Toward foundation models on meteorological data.
\newblock \emph{arXiv preprint arXiv:2301.09152}, 2023{\natexlab{a}}.

\bibitem[Chen et~al.(2023{\natexlab{b}})Chen, Long, Shen, Jiang, and Zhang]{chen2023federated2}
Shengchao Chen, Guodong Long, Tao Shen, Jing Jiang, and Chengqi Zhang.
\newblock Federated prompt learning for weather foundation models on devices.
\newblock \emph{arXiv preprint arXiv:2305.14244}, 2023{\natexlab{b}}.

\bibitem[Chen et~al.(2024{\natexlab{d}})Chen, Ju, Dalal, Zhu, and Khisti]{chen2024robust}
Shuangyi Chen, Yue Ju, Hardik Dalal, Zhongwen Zhu, and Ashish Khisti.
\newblock Robust federated finetuning of foundation models via alternating minimization of lora.
\newblock \emph{arXiv preprint arXiv:2409.02346}, 2024{\natexlab{d}}.

\bibitem[Chen et~al.(2021{\natexlab{b}})Chen, Chen, Smiley, Shah, Borova, Langdon, Moussa, Beane, Huang, Routledge, et~al.]{chen2021finqa}
Zhiyu Chen, Wenhu Chen, Charese Smiley, Sameena Shah, Iana Borova, Dylan Langdon, Reema Moussa, Matt Beane, Ting-Hao Huang, Bryan Routledge, et~al.
\newblock Finqa: A dataset of numerical reasoning over financial data.
\newblock \emph{arXiv preprint arXiv:2109.00122}, 2021{\natexlab{b}}.

\bibitem[Chen et~al.(2024{\natexlab{e}})Chen, Ma, Zhang, Hao, Yan, Nourbakhsh, Yang, McAuley, Petzold, and Wang]{chen2024survey}
Zhiyu~Zoey Chen, Jing Ma, Xinlu Zhang, Nan Hao, An~Yan, Armineh Nourbakhsh, Xianjun Yang, Julian McAuley, Linda Petzold, and William~Yang Wang.
\newblock A survey on large language models for critical societal domains: Finance, healthcare, and law.
\newblock \emph{arXiv preprint arXiv:2405.01769}, 2024{\natexlab{e}}.

\bibitem[Chernyshev et~al.(2024)Chernyshev, Polshkov, Artemova, Myasnikov, Stepanov, Miasnikov, and Tilga]{chernyshev2024u}
Konstantin Chernyshev, Vitaliy Polshkov, Ekaterina Artemova, Alex Myasnikov, Vlad Stepanov, Alexei Miasnikov, and Sergei Tilga.
\newblock U-math: A university-level benchmark for evaluating mathematical skills in llms.
\newblock \emph{arXiv preprint arXiv:2412.03205}, 2024.

\bibitem[Cho et~al.(2024)Cho, Liu, Xu, Fahrezi, and Joshi]{cho2024heterogeneous}
Yae~Jee Cho, Luyang Liu, Zheng Xu, Aldi Fahrezi, and Gauri Joshi.
\newblock Heterogeneous lora for federated fine-tuning of on-device foundation models.
\newblock In \emph{Proceedings of the 2024 Conference on Empirical Methods in Natural Language Processing}, pp.\  12903--12913, 2024.

\bibitem[Choi et~al.(2023)Choi, Pei, Kumar, Shu, and Jurgens]{choi2023llms}
Minje Choi, Jiaxin Pei, Sagar Kumar, Chang Shu, and David Jurgens.
\newblock Do llms understand social knowledge? evaluating the sociability of large language models with socket benchmark.
\newblock \emph{arXiv preprint arXiv:2305.14938}, 2023.

\bibitem[Choudhary et~al.(2020)Choudhary, Mishra, Goswami, and Sarangapani]{choudhary2020comprehensive}
Tejalal Choudhary, Vipul Mishra, Anurag Goswami, and Jagannathan Sarangapani.
\newblock A comprehensive survey on model compression and acceleration.
\newblock \emph{Artificial Intelligence Review}, 53:\penalty0 5113--5155, 2020.

\bibitem[Churin et~al.(2024)Churin, Apishev, Tikhonova, Shevelev, Bulatov, Kuratov, Averkiev, and Fenogenova]{churin2024long}
Igor Churin, Murat Apishev, Maria Tikhonova, Denis Shevelev, Aydar Bulatov, Yuri Kuratov, Sergej Averkiev, and Alena Fenogenova.
\newblock Long input benchmark for russian analysis.
\newblock \emph{arXiv preprint arXiv:2408.02439}, 2024.

\bibitem[Clark et~al.(2018)Clark, Cowhey, Etzioni, Khot, Sabharwal, Schoenick, and Tafjord]{clark2018think}
Peter Clark, Isaac Cowhey, Oren Etzioni, Tushar Khot, Ashish Sabharwal, Carissa Schoenick, and Oyvind Tafjord.
\newblock Think you have solved question answering? try arc, the ai2 reasoning challenge.
\newblock \emph{arXiv preprint arXiv:1803.05457}, 2018.

\bibitem[Cobbe et~al.(2021)Cobbe, Kosaraju, Bavarian, Chen, Jun, Kaiser, Plappert, Tworek, Hilton, Nakano, Hesse, and Schulman]{cobbe2021gsm8k}
Karl Cobbe, Vineet Kosaraju, Mohammad Bavarian, Mark Chen, Heewoo Jun, Lukasz Kaiser, Matthias Plappert, Jerry Tworek, Jacob Hilton, Reiichiro Nakano, Christopher Hesse, and John Schulman.
\newblock Training verifiers to solve math word problems.
\newblock \emph{arXiv preprint arXiv:2110.14168}, 2021.

\bibitem[Conover et~al.(2023)Conover, Hayes, Mathur, Xie, Wan, Shah, Ghodsi, Wendell, Zaharia, and Xin]{conover2023free}
Mike Conover, Matt Hayes, Ankit Mathur, Jianwei Xie, Jun Wan, Sam Shah, Ali Ghodsi, Patrick Wendell, Matei Zaharia, and Reynold Xin.
\newblock Free dolly: Introducing the world’s first truly open instruction-tuned llm, 2023.

\bibitem[Cui et~al.(2024)Cui, Li, Wang, and Shi]{cui2024harmonizing}
Tianyu Cui, Hongxia Li, Jingya Wang, and Ye~Shi.
\newblock Harmonizing generalization and personalization in federated prompt learning.
\newblock \emph{arXiv preprint arXiv:2405.09771}, 2024.

\bibitem[Dai et~al.(2024)Dai, Pechi, Yang, Banga, and Mantri]{dai2024deniahl}
Hui Dai, Dan Pechi, Xinyi Yang, Garvit Banga, and Raghav Mantri.
\newblock Deniahl: In-context features influence llm needle-in-a-haystack abilities.
\newblock \emph{arXiv preprint arXiv:2411.19360}, 2024.

\bibitem[Dai et~al.(2025)Dai, Wu, Liu, Yu, Hu, Liu, and Geng]{dai2025fedata}
Jian Dai, Hao Wu, Huan Liu, Liheng Yu, Xing Hu, Xiao Liu, and Daoying Geng.
\newblock Fedata: Adaptive attention aggregation for federated self-supervised medical image segmentation.
\newblock \emph{Neurocomputing}, 613:\penalty0 128691, 2025.

\bibitem[Dai et~al.(2023)Dai, Feng, Huang, Jia, Xie, Zhang, Han, Tian, and Wang]{dai2023laiw}
Yongfu Dai, Duanyu Feng, Jimin Huang, Haochen Jia, Qianqian Xie, Yifang Zhang, Weiguang Han, Wei Tian, and Hao Wang.
\newblock Laiw: a chinese legal large language models benchmark.
\newblock \emph{arXiv preprint arXiv:2310.05620}, 2023.

\bibitem[Deng et~al.(2020)Deng, Li, Han, Shi, and Xie]{deng2020model}
Lei Deng, Guoqi Li, Song Han, Luping Shi, and Yuan Xie.
\newblock Model compression and hardware acceleration for neural networks: A comprehensive survey.
\newblock \emph{Proceedings of the IEEE}, 108\penalty0 (4):\penalty0 485--532, 2020.

\bibitem[Deng et~al.(2024)Deng, Thrampoulidis, and Li]{deng2024unlocking}
Wenlong Deng, Christos Thrampoulidis, and Xiaoxiao Li.
\newblock Unlocking the potential of prompt-tuning in bridging generalized and personalized federated learning.
\newblock In \emph{Proceedings of the IEEE/CVF Conference on Computer Vision and Pattern Recognition}, pp.\  6087--6097, 2024.

\bibitem[Dettmers et~al.(2023)Dettmers, Pagnoni, Holtzman, and Zettlemoyer]{dettmers2023qlora}
Tim Dettmers, Artidoro Pagnoni, Ari Holtzman, and Luke Zettlemoyer.
\newblock Qlora: Efficient finetuning of quantized llms.
\newblock \emph{Advances in neural information processing systems}, 36:\penalty0 10088--10115, 2023.

\bibitem[Devlin et~al.(2019)Devlin, Chang, Lee, and Toutanova]{BERT}
Jacob Devlin, Ming-Wei Chang, Kenton Lee, and Kristina Toutanova.
\newblock Bert: Pre-training of deep bidirectional transformers for language understanding.
\newblock In \emph{Proceedings of the 2019 Conference of the North}, Jan 2019.
\newblock \doi{10.18653/v1/n19-1423}.
\newblock URL \url{http://dx.doi.org/10.18653/v1/n19-1423}.

\bibitem[Ding et~al.(2023{\natexlab{a}})Ding, Chen, Xu, Qin, Zheng, Hu, Liu, Sun, and Zhou]{ding2023enhancing}
Ning Ding, Yulin Chen, Bokai Xu, Yujia Qin, Zhi Zheng, Shengding Hu, Zhiyuan Liu, Maosong Sun, and Bowen Zhou.
\newblock Enhancing chat language models by scaling high-quality instructional conversations.
\newblock \emph{arXiv preprint arXiv:2305.14233}, 2023{\natexlab{a}}.

\bibitem[Ding et~al.(2023{\natexlab{b}})Ding, Qin, Yang, Wei, Yang, Su, Hu, Chen, Chan, Chen, et~al.]{ding2023parameter}
Ning Ding, Yujia Qin, Guang Yang, Fuchao Wei, Zonghan Yang, Yusheng Su, Shengding Hu, Yulin Chen, Chi-Min Chan, Weize Chen, et~al.
\newblock Parameter-efficient fine-tuning of large-scale pre-trained language models.
\newblock \emph{Nature Machine Intelligence}, 5\penalty0 (3):\penalty0 220--235, 2023{\natexlab{b}}.

\bibitem[Dodge et~al.(2020)Dodge, Ilharco, Schwartz, Farhadi, Hajishirzi, and Smith]{dodge2020fine}
Jesse Dodge, Gabriel Ilharco, Roy Schwartz, Ali Farhadi, Hannaneh Hajishirzi, and Noah Smith.
\newblock Fine-tuning pretrained language models: Weight initializations, data orders, and early stopping.
\newblock \emph{arXiv preprint arXiv:2002.06305}, 2020.

\bibitem[Dong et~al.(2023{\natexlab{a}})Dong, Xie, Ding, Shen, and Li]{dong2023tunable}
Chenhe Dong, Yuexiang Xie, Bolin Ding, Ying Shen, and Yaliang Li.
\newblock Tunable soft prompts are messengers in federated learning.
\newblock \emph{arXiv preprint arXiv:2311.06805}, 2023{\natexlab{a}}.

\bibitem[Dong et~al.(2023{\natexlab{b}})Dong, Tang, Li, Zhao, and Wen]{dong2023bamboo}
Zican Dong, Tianyi Tang, Junyi Li, Wayne~Xin Zhao, and Ji-Rong Wen.
\newblock Bamboo: A comprehensive benchmark for evaluating long text modeling capacities of large language models.
\newblock \emph{arXiv preprint arXiv:2309.13345}, 2023{\natexlab{b}}.

\bibitem[Du et~al.(2023)Du, Liu, Wang, Wang, Liu, Chen, Feng, Sha, Peng, and Lou]{du2023classeval}
Xueying Du, Mingwei Liu, Kaixin Wang, Hanlin Wang, Junwei Liu, Yixuan Chen, Jiayi Feng, Chaofeng Sha, Xin Peng, and Yiling Lou.
\newblock Classeval: A manually-crafted benchmark for evaluating llms on class-level code generation.
\newblock \emph{arXiv preprint arXiv:2308.01861}, 2023.

\bibitem[Du et~al.(2024)Du, Zhang, Yue, Huang, Zhang, Xu, Xu, and Chen]{du2024communication}
Yichao Du, Zhirui Zhang, Linan Yue, Xu~Huang, Yuqing Zhang, Tong Xu, Linli Xu, and Enhong Chen.
\newblock Communication-efficient personalized federated learning for speech-to-text tasks.
\newblock In \emph{ICASSP 2024-2024 IEEE International Conference on Acoustics, Speech and Signal Processing (ICASSP)}, pp.\  10001--10005. IEEE, 2024.

\bibitem[Dua et~al.(2019)Dua, Wang, Dasigi, Stanovsky, Singh, and Gardner]{dua2019drop}
Dheeru Dua, Yizhong Wang, Pradeep Dasigi, Gabriel Stanovsky, Sameer Singh, and Matt Gardner.
\newblock {DROP}: A reading comprehension benchmark requiring discrete reasoning over paragraphs.
\newblock In \emph{Proc. of NAACL}, 2019.

\bibitem[Dubey et~al.(2024)Dubey, Jauhri, Pandey, Kadian, Al-Dahle, Letman, Mathur, Schelten, Yang, Fan, et~al.]{dubey2024llama}
Abhimanyu Dubey, Abhinav Jauhri, Abhinav Pandey, Abhishek Kadian, Ahmad Al-Dahle, Aiesha Letman, Akhil Mathur, Alan Schelten, Amy Yang, Angela Fan, et~al.
\newblock The llama 3 herd of models.
\newblock \emph{arXiv preprint arXiv:2407.21783}, 2024.

\bibitem[Dubois et~al.(2023)Dubois, Li, Taori, Zhang, Gulrajani, Ba, Guestrin, Liang, and Hashimoto]{dubois2023alpacafarm}
Yann Dubois, Chen~Xuechen Li, Rohan Taori, Tianyi Zhang, Ishaan Gulrajani, Jimmy Ba, Carlos Guestrin, Percy~S Liang, and Tatsunori~B Hashimoto.
\newblock Alpacafarm: A simulation framework for methods that learn from human feedback.
\newblock \emph{Advances in Neural Information Processing Systems}, 36:\penalty0 30039--30069, 2023.

\bibitem[Fan et~al.(2024{\natexlab{a}})Fan, Sun, Xue, Zhang, Zhang, and Ruan]{fan2024medodyssey}
Yongqi Fan, Hongli Sun, Kui Xue, Xiaofan Zhang, Shaoting Zhang, and Tong Ruan.
\newblock Medodyssey: A medical domain benchmark for long context evaluation up to 200k tokens.
\newblock \emph{arXiv preprint arXiv:2406.15019}, 2024{\natexlab{a}}.

\bibitem[Fan et~al.(2024{\natexlab{b}})Fan, Xu, Wang, Huo, Chen, and Guo]{fan2024overcome}
Yunfeng Fan, Wenchao Xu, Haozhao Wang, Fushuo Huo, Jinyu Chen, and Song Guo.
\newblock Overcome modal bias in multi-modal federated learning via balanced modality selection.
\newblock In \emph{European Conference on Computer Vision}, pp.\  178--195. Springer, 2024{\natexlab{b}}.

\bibitem[Fang et~al.(2024{\natexlab{a}})Fang, Wan, Lu, Xing, and Zou]{fang2024mathodyssey}
Meng Fang, Xiangpeng Wan, Fei Lu, Fei Xing, and Kai Zou.
\newblock Mathodyssey: Benchmarking mathematical problem-solving skills in large language models using odyssey math data.
\newblock \emph{arXiv preprint arXiv:2406.18321}, 2024{\natexlab{a}}.

\bibitem[Fang et~al.(2024{\natexlab{b}})Fang, Lin, Chen, Chen, Gao, and Fang]{fang2024automated}
Zihan Fang, Zheng Lin, Zhe Chen, Xianhao Chen, Yue Gao, and Yuguang Fang.
\newblock Automated federated pipeline for parameter-efficient fine-tuning of large language models.
\newblock \emph{arXiv preprint arXiv:2404.06448}, 2024{\natexlab{b}}.

\bibitem[Fei et~al.(2023)Fei, Shen, Zhu, Zhou, Han, Zhang, Chen, Shen, and Ge]{fei2023lawbench}
Zhiwei Fei, Xiaoyu Shen, Dawei Zhu, Fengzhe Zhou, Zhuo Han, Songyang Zhang, Kai Chen, Zongwen Shen, and Jidong Ge.
\newblock Lawbench: Benchmarking legal knowledge of large language models.
\newblock \emph{arXiv preprint arXiv:2309.16289}, 2023.

\bibitem[Feng et~al.(2023{\natexlab{a}})Feng, Li, Xu, Liu, Fu, and Zuo]{feng2023learning}
Chun-Mei Feng, Bangjun Li, Xinxing Xu, Yong Liu, Huazhu Fu, and Wangmeng Zuo.
\newblock Learning federated visual prompt in null space for mri reconstruction.
\newblock In \emph{Proceedings of the IEEE/CVF Conference on Computer Vision and Pattern Recognition}, pp.\  8064--8073, 2023{\natexlab{a}}.

\bibitem[Feng et~al.(2023{\natexlab{b}})Feng, Bose, Zhang, Hebbar, Ramakrishna, Gupta, Zhang, Avestimehr, and Narayanan]{feng2023fedmultimodal}
Tiantian Feng, Digbalay Bose, Tuo Zhang, Rajat Hebbar, Anil Ramakrishna, Rahul Gupta, Mi~Zhang, Salman Avestimehr, and Shrikanth Narayanan.
\newblock Fedmultimodal: A benchmark for multimodal federated learning.
\newblock In \emph{Proceedings of the 29th ACM SIGKDD Conference on Knowledge Discovery and Data Mining}, pp.\  4035--4045, 2023{\natexlab{b}}.

\bibitem[Feng et~al.(2024{\natexlab{a}})Feng, Feng, Du, Kan, and Qin]{feng2024adapter}
Xiachong Feng, Xiaocheng Feng, Xiyuan Du, Min-Yen Kan, and Bing Qin.
\newblock Adapter-based selective knowledge distillation for federated multi-domain meeting summarization.
\newblock \emph{IEEE/ACM Transactions on Audio, Speech, and Language Processing}, 2024{\natexlab{a}}.

\bibitem[Feng et~al.(2024{\natexlab{b}})Feng, Tian, Zhu, Han, Luo, Zhang, and Song]{feng2024cp}
Yu~Feng, Zhen Tian, Yifan Zhu, Zongfu Han, Haoran Luo, Guangwei Zhang, and Meina Song.
\newblock Cp-prompt: Composition-based cross-modal prompting for domain-incremental continual learning.
\newblock In \emph{Proceedings of the 32nd ACM International Conference on Multimedia}, pp.\  2729--2738, 2024{\natexlab{b}}.

\bibitem[Flowers(2025)]{josephgflowers2025financeinstruct}
Joseph~G. Flowers.
\newblock Finance-instruct-500k, 2025.
\newblock URL \url{https://huggingface.co/datasets/Josephgflowers/Finance-Instruct-500k}.

\bibitem[Frohberg \& Binder(2021)Frohberg and Binder]{frohberg2021crass}
J{\"o}rg Frohberg and Frank Binder.
\newblock Crass: A novel data set and benchmark to test counterfactual reasoning of large language models.
\newblock \emph{arXiv preprint arXiv:2112.11941}, 2021.

\bibitem[Fu et~al.(2022)Fu, Zhang, Dong, Chen, and Li]{fu2022federated}
Xingbo Fu, Binchi Zhang, Yushun Dong, Chen Chen, and Jundong Li.
\newblock Federated graph machine learning: A survey of concepts, techniques, and applications.
\newblock \emph{ACM SIGKDD Explorations Newsletter}, 24\penalty0 (2):\penalty0 32--47, 2022.

\bibitem[Fu et~al.(2024{\natexlab{a}})Fu, Chen, He, Wang, Zhang, Chen, and Li]{fu2024virtual}
Xingbo Fu, Zihan Chen, Yinhan He, Song Wang, Binchi Zhang, Chen Chen, and Jundong Li.
\newblock Virtual nodes can help: Tackling distribution shifts in federated graph learning.
\newblock \emph{arXiv preprint arXiv:2412.19229}, 2024{\natexlab{a}}.

\bibitem[Fu et~al.(2024{\natexlab{b}})Fu, Chen, Zhang, Chen, and Li]{fu2024federated}
Xingbo Fu, Zihan Chen, Binchi Zhang, Chen Chen, and Jundong Li.
\newblock Federated graph learning with structure proxy alignment.
\newblock In \emph{Proceedings of the 30th ACM SIGKDD Conference on Knowledge Discovery and Data Mining}, pp.\  827--838, 2024{\natexlab{b}}.

\bibitem[Fu et~al.(2024{\natexlab{c}})Fu, Wang, Dong, Zhang, Chen, and Li]{fu2024federated2}
Xingbo Fu, Song Wang, Yushun Dong, Binchi Zhang, Chen Chen, and Jundong Li.
\newblock Federated graph learning with graphless clients.
\newblock \emph{arXiv preprint arXiv:2411.08374}, 2024{\natexlab{c}}.

\bibitem[Fu et~al.(2023)Fu, Ou, Chen, Wan, Peng, and Khot]{fu2023chain}
Yao Fu, Litu Ou, Mingyu Chen, Yuhao Wan, Hao Peng, and Tushar Khot.
\newblock Chain-of-thought hub: A continuous effort to measure large language models' reasoning performance.
\newblock \emph{arXiv preprint arXiv:2305.17306}, 2023.

\bibitem[Gao et~al.(2024{\natexlab{a}})Gao, Zhao, Qiu, Wang, Yao, and Hu]{gao2024cp}
Fei Gao, Yunfeng Zhao, Chao Qiu, Xiaofei Wang, Haipeng Yao, and Qinghua Hu.
\newblock Cp 2 gfed: Cross-granular and personalized prompt-based green federated tuning for giant models.
\newblock In \emph{2024 IEEE/ACM 32nd International Symposium on Quality of Service (IWQoS)}, pp.\  1--10. IEEE, 2024{\natexlab{a}}.

\bibitem[Gao et~al.(2020)Gao, Biderman, Black, Golding, Hoppe, Foster, Phang, He, Thite, Nabeshima, et~al.]{gao2020pile}
Leo Gao, Stella Biderman, Sid Black, Laurence Golding, Travis Hoppe, Charles Foster, Jason Phang, Horace He, Anish Thite, Noa Nabeshima, et~al.
\newblock The pile: An 800gb dataset of diverse text for language modeling.
\newblock \emph{arXiv preprint arXiv:2101.00027}, 2020.

\bibitem[Gao et~al.(2024{\natexlab{b}})Gao, Hu, Ruan, Pu, and Wan]{gao2024llm}
Mingqi Gao, Xinyu Hu, Jie Ruan, Xiao Pu, and Xiaojun Wan.
\newblock Llm-based nlg evaluation: Current status and challenges.
\newblock \emph{arXiv preprint arXiv:2402.01383}, 2024{\natexlab{b}}.

\bibitem[Gao et~al.(2024{\natexlab{c}})Gao, Liu, Xu, Wu, Zhang, Li, Yang, Liu, and Sun]{gao2024softclip}
Yuting Gao, Jinfeng Liu, Zihan Xu, Tong Wu, Enwei Zhang, Ke~Li, Jie Yang, Wei Liu, and Xing Sun.
\newblock Softclip: Softer cross-modal alignment makes clip stronger.
\newblock In \emph{Proceedings of the AAAI Conference on Artificial Intelligence}, volume~38, pp.\  1860--1868, 2024{\natexlab{c}}.

\bibitem[Ghiasvand et~al.(2024)Ghiasvand, Yang, Xue, Alizadeh, Zhang, and Pedarsani]{ghiasvand2024communication}
Sajjad Ghiasvand, Yifan Yang, Zhiyu Xue, Mahnoosh Alizadeh, Zheng Zhang, and Ramtin Pedarsani.
\newblock Communication-efficient and tensorized federated fine-tuning of large language models.
\newblock \emph{arXiv preprint arXiv:2410.13097}, 2024.

\bibitem[Gim \& Ko(2022)Gim and Ko]{gim2022memory}
In~Gim and JeongGil Ko.
\newblock Memory-efficient dnn training on mobile devices.
\newblock In \emph{Proceedings of the 20th Annual International Conference on Mobile Systems, Applications and Services}, pp.\  464--476, 2022.

\bibitem[Goetz \& Tewari(2020)Goetz and Tewari]{goetz2020federated}
Jack Goetz and Ambuj Tewari.
\newblock Federated learning via synthetic data.
\newblock \emph{arXiv preprint arXiv:2008.04489}, 2020.

\bibitem[Gong et~al.(2024)Gong, Cui, Zhang, Wang, Nie, and Zhu]{gong2024federated}
Shuai Gong, Chaoran Cui, Chunyun Zhang, Wenna Wang, Xiushan Nie, and Lei Zhu.
\newblock Federated domain generalization via prompt learning and aggregation.
\newblock \emph{arXiv preprint arXiv:2411.10063}, 2024.

\bibitem[Gu et~al.(2024{\natexlab{a}})Gu, Rozi{\`e}re, Leather, Solar-Lezama, Synnaeve, and Wang]{gu2024cruxeval}
Alex Gu, Baptiste Rozi{\`e}re, Hugh Leather, Armando Solar-Lezama, Gabriel Synnaeve, and Sida~I Wang.
\newblock Cruxeval: A benchmark for code reasoning, understanding and execution.
\newblock \emph{arXiv preprint arXiv:2401.03065}, 2024{\natexlab{a}}.

\bibitem[Gu et~al.(2024{\natexlab{b}})Gu, Zhu, Ye, Zhang, Wang, Zhu, Jiang, Xiong, Li, Wu, et~al.]{gu2024xiezhi}
Zhouhong Gu, Xiaoxuan Zhu, Haoning Ye, Lin Zhang, Jianchen Wang, Yixin Zhu, Sihang Jiang, Zhuozhi Xiong, Zihan Li, Weijie Wu, et~al.
\newblock Xiezhi: An ever-updating benchmark for holistic domain knowledge evaluation.
\newblock In \emph{Proceedings of the AAAI Conference on Artificial Intelligence}, volume~38, pp.\  18099--18107, 2024{\natexlab{b}}.

\bibitem[Guha et~al.(2023)Guha, Nyarko, Ho, R{\'e}, Chilton, Chohlas-Wood, Peters, Waldon, Rockmore, Zambrano, et~al.]{guha2023legalbench}
Neel Guha, Julian Nyarko, Daniel Ho, Christopher R{\'e}, Adam Chilton, Alex Chohlas-Wood, Austin Peters, Brandon Waldon, Daniel Rockmore, Diego Zambrano, et~al.
\newblock Legalbench: A collaboratively built benchmark for measuring legal reasoning in large language models.
\newblock \emph{Advances in Neural Information Processing Systems}, 36:\penalty0 44123--44279, 2023.

\bibitem[Guo et~al.(2023{\natexlab{a}})Guo, Zhang, Wang, Jiang, Nie, Ding, Yue, and Wu]{guo2023close}
Biyang Guo, Xin Zhang, Ziyuan Wang, Minqi Jiang, Jinran Nie, Yuxuan Ding, Jianwei Yue, and Yupeng Wu.
\newblock How close is chatgpt to human experts? comparison corpus, evaluation, and detection.
\newblock \emph{arXiv preprint arXiv:2301.07597}, 2023{\natexlab{a}}.

\bibitem[Guo et~al.(2025)Guo, Yang, Zhang, Song, Zhang, Xu, Zhu, Ma, Wang, Bi, et~al.]{guo2025deepseek}
Daya Guo, Dejian Yang, Haowei Zhang, Junxiao Song, Ruoyu Zhang, Runxin Xu, Qihao Zhu, Shirong Ma, Peiyi Wang, Xiao Bi, et~al.
\newblock Deepseek-r1: Incentivizing reasoning capability in llms via reinforcement learning.
\newblock \emph{arXiv preprint arXiv:2501.12948}, 2025.

\bibitem[Guo et~al.(2024{\natexlab{a}})Guo, Lu, Yu, Nguyen, and Yin]{guo2024prompt}
Lei Guo, Ziang Lu, Junliang Yu, Quoc Viet~Hung Nguyen, and Hongzhi Yin.
\newblock Prompt-enhanced federated content representation learning for cross-domain recommendation.
\newblock In \emph{Proceedings of the ACM on Web Conference 2024}, pp.\  3139--3149, 2024{\natexlab{a}}.

\bibitem[Guo et~al.(2024{\natexlab{b}})Guo, Zeng, Wang, Fan, Wang, and Qu]{guo2024selective}
Pengxin Guo, Shuang Zeng, Yanran Wang, Huijie Fan, Feifei Wang, and Liangqiong Qu.
\newblock Selective aggregation for low-rank adaptation in federated learning.
\newblock \emph{arXiv preprint arXiv:2410.01463}, 2024{\natexlab{b}}.

\bibitem[Guo et~al.(2023{\natexlab{b}})Guo, Guo, and Wang]{guo2023pfedprompt}
Tao Guo, Song Guo, and Junxiao Wang.
\newblock Pfedprompt: Learning personalized prompt for vision-language models in federated learning.
\newblock In \emph{Proceedings of the ACM Web Conference 2023}, pp.\  1364--1374, 2023{\natexlab{b}}.

\bibitem[Guo et~al.(2023{\natexlab{c}})Guo, Guo, Wang, Tang, and Xu]{guo2023promptfl}
Tao Guo, Song Guo, Junxiao Wang, Xueyang Tang, and Wenchao Xu.
\newblock Promptfl: Let federated participants cooperatively learn prompts instead of models-federated learning in age of foundation model.
\newblock \emph{IEEE Transactions on Mobile Computing}, 2023{\natexlab{c}}.

\bibitem[Guo et~al.(2024{\natexlab{c}})Guo, Guo, and Wang]{guo2024explore}
Tao Guo, Song Guo, and Junxiao Wang.
\newblock Explore and cure: Unveiling sample effectiveness with context-aware federated prompt tuning.
\newblock \emph{IEEE Transactions on Mobile Computing}, 2024{\natexlab{c}}.

\bibitem[Halbe et~al.(2023)Halbe, Smith, Tian, and Kira]{halbe2023hepco}
Shaunak Halbe, James~Seale Smith, Junjiao Tian, and Zsolt Kira.
\newblock Hepco: Data-free heterogeneous prompt consolidation for continual federated learning.
\newblock \emph{arXiv preprint arXiv:2306.09970}, 2023.

\bibitem[Han et~al.(2023)Han, Adams, Papaioannou, Grundmann, Oberhauser, L{\"o}ser, Truhn, and Bressem]{han2023medalpaca}
Tianyu Han, Lisa~C Adams, Jens-Michalis Papaioannou, Paul Grundmann, Tom Oberhauser, Alexander L{\"o}ser, Daniel Truhn, and Keno~K Bressem.
\newblock Medalpaca--an open-source collection of medical conversational ai models and training data.
\newblock \emph{arXiv preprint arXiv:2304.08247}, 2023.

\bibitem[Han et~al.(2024)Han, Gao, Liu, Zhang, and Zhang]{han2024parameter}
Zeyu Han, Chao Gao, Jinyang Liu, Jeff Zhang, and Sai~Qian Zhang.
\newblock Parameter-efficient fine-tuning for large models: A comprehensive survey.
\newblock \emph{arXiv preprint arXiv:2403.14608}, 2024.

\bibitem[He et~al.(2024)He, Zeng, Huang, Chen, Xiao, He, Zhou, Liang, and Xiao]{he2024can}
Qianyu He, Jie Zeng, Wenhao Huang, Lina Chen, Jin Xiao, Qianxi He, Xunzhe Zhou, Jiaqing Liang, and Yanghua Xiao.
\newblock Can large language models understand real-world complex instructions?
\newblock In \emph{Proceedings of the AAAI Conference on Artificial Intelligence}, volume~38, pp.\  18188--18196, 2024.

\bibitem[Hendrycks et~al.(2020)Hendrycks, Burns, Basart, Zou, Mazeika, Song, and Steinhardt]{hendrycks2020measuring}
Dan Hendrycks, Collin Burns, Steven Basart, Andy Zou, Mantas Mazeika, Dawn Song, and Jacob Steinhardt.
\newblock Measuring massive multitask language understanding.
\newblock \emph{arXiv preprint arXiv:2009.03300}, 2020.

\bibitem[Hendrycks et~al.(2021{\natexlab{a}})Hendrycks, Basart, Kadavath, Mazeika, Arora, Guo, Burns, Puranik, He, Song, et~al.]{hendrycks2021measuring2}
Dan Hendrycks, Steven Basart, Saurav Kadavath, Mantas Mazeika, Akul Arora, Ethan Guo, Collin Burns, Samir Puranik, Horace He, Dawn Song, et~al.
\newblock Measuring coding challenge competence with apps.
\newblock \emph{arXiv preprint arXiv:2105.09938}, 2021{\natexlab{a}}.

\bibitem[Hendrycks et~al.(2021{\natexlab{b}})Hendrycks, Burns, Chen, and Ball]{hendrycks2021cuad}
Dan Hendrycks, Collin Burns, Anya Chen, and Spencer Ball.
\newblock Cuad: an expert-annotated nlp dataset for legal contract review.
\newblock \emph{arXiv preprint arXiv:2103.06268}, 2021{\natexlab{b}}.

\bibitem[Hendrycks et~al.(2021{\natexlab{c}})Hendrycks, Burns, Kadavath, Arora, Basart, Tang, Song, and Steinhardt]{math}
Dan Hendrycks, Collin Burns, Saurav Kadavath, Akul Arora, Steven Basart, Eric Tang, Dawn Song, and Jacob Steinhardt.
\newblock Measuring mathematical problem solving with the math dataset.
\newblock In \emph{Thirty-fifth Conference on Neural Information Processing Systems Datasets and Benchmarks Track (Round 2)}, 2021{\natexlab{c}}.

\bibitem[Hirano(2024)]{hirano2024construction}
Masanori Hirano.
\newblock Construction of a japanese financial benchmark for large language models.
\newblock \emph{arXiv preprint arXiv:2403.15062}, 2024.

\bibitem[Hou et~al.(2023)Hou, Zhang, and Callison-Burch]{hou2023choice}
Zhaoyi~Joey Hou, Li~Zhang, and Chris Callison-Burch.
\newblock Choice-75: A dataset on decision branching in script learning.
\newblock \emph{arXiv preprint arXiv:2309.11737}, 2023.

\bibitem[Houlsby et~al.(2019)Houlsby, Giurgiu, Jastrzebski, Morrone, De~Laroussilhe, Gesmundo, Attariyan, and Gelly]{houlsby2019parameter}
Neil Houlsby, Andrei Giurgiu, Stanislaw Jastrzebski, Bruna Morrone, Quentin De~Laroussilhe, Andrea Gesmundo, Mona Attariyan, and Sylvain Gelly.
\newblock Parameter-efficient transfer learning for nlp.
\newblock In \emph{International conference on machine learning}, pp.\  2790--2799. PMLR, 2019.

\bibitem[Hu et~al.(2021)Hu, Shen, Wallis, Allen-Zhu, Li, Wang, Wang, and Chen]{hu2021lora}
Edward~J Hu, Yelong Shen, Phillip Wallis, Zeyuan Allen-Zhu, Yuanzhi Li, Shean Wang, Lu~Wang, and Weizhu Chen.
\newblock Lora: Low-rank adaptation of large language models.
\newblock \emph{arXiv preprint arXiv:2106.09685}, 2021.

\bibitem[Huang et~al.(2024{\natexlab{a}})Huang, Cui, Wang, Yang, Liao, Song, Yao, and Su]{huang2024mitigating}
Jianheng Huang, Leyang Cui, Ante Wang, Chengyi Yang, Xinting Liao, Linfeng Song, Junfeng Yao, and Jinsong Su.
\newblock Mitigating catastrophic forgetting in large language models with self-synthesized rehearsal.
\newblock \emph{arXiv preprint arXiv:2403.01244}, 2024{\natexlab{a}}.

\bibitem[Huang et~al.(2025)Huang, Yu, Ma, Zhong, Feng, Wang, Chen, Peng, Feng, Qin, et~al.]{huang2025survey}
Lei Huang, Weijiang Yu, Weitao Ma, Weihong Zhong, Zhangyin Feng, Haotian Wang, Qianglong Chen, Weihua Peng, Xiaocheng Feng, Bing Qin, et~al.
\newblock A survey on hallucination in large language models: Principles, taxonomy, challenges, and open questions.
\newblock \emph{ACM Transactions on Information Systems}, 43\penalty0 (2):\penalty0 1--55, 2025.

\bibitem[Huang et~al.(2023{\natexlab{a}})Huang, Tao, Zhang, An, Jiang, Chen, Wu, and Feng]{huang2023lawyer}
Quzhe Huang, Mingxu Tao, Chen Zhang, Zhenwei An, Cong Jiang, Zhibin Chen, Zirui Wu, and Yansong Feng.
\newblock Lawyer llama technical report.
\newblock \emph{arXiv preprint arXiv:2305.15062}, 2023{\natexlab{a}}.

\bibitem[Huang et~al.(2024{\natexlab{b}})Huang, Ye, Shi, Wan, Li, Du, and Yang]{huang2024federated}
Wenke Huang, Mang Ye, Zekun Shi, Guancheng Wan, He~Li, Bo~Du, and Qiang Yang.
\newblock Federated learning for generalization, robustness, fairness: A survey and benchmark.
\newblock \emph{IEEE Transactions on Pattern Analysis and Machine Intelligence}, 2024{\natexlab{b}}.

\bibitem[Huang et~al.(2023{\natexlab{b}})Huang, Bai, Zhu, Zhang, Zhang, Su, Liu, Lv, Zhang, Fu, et~al.]{huang2023c}
Yuzhen Huang, Yuzhuo Bai, Zhihao Zhu, Junlei Zhang, Jinghan Zhang, Tangjun Su, Junteng Liu, Chuancheng Lv, Yikai Zhang, Yao Fu, et~al.
\newblock C-eval: A multi-level multi-discipline chinese evaluation suite for foundation models.
\newblock \emph{Advances in Neural Information Processing Systems}, 36:\penalty0 62991--63010, 2023{\natexlab{b}}.

\bibitem[Husain et~al.(2019)Husain, Wu, Gazit, Allamanis, and Brockschmidt]{husain2019codesearchnet}
Hamel Husain, Ho-Hsiang Wu, Tiferet Gazit, Miltiadis Allamanis, and Marc Brockschmidt.
\newblock {CodeSearchNet} challenge: Evaluating the state of semantic code search.
\newblock \emph{arXiv preprint arXiv:1909.09436}, 2019.

\bibitem[Islam et~al.(2023)Islam, Kannappan, Kiela, Qian, Scherrer, and Vidgen]{islam2023financebench}
Pranab Islam, Anand Kannappan, Douwe Kiela, Rebecca Qian, Nino Scherrer, and Bertie Vidgen.
\newblock Financebench: A new benchmark for financial question answering.
\newblock \emph{arXiv preprint arXiv:2311.11944}, 2023.

\bibitem[Ji et~al.(2023{\natexlab{a}})Ji, Deng, Gong, Peng, Niu, Ma, and Li]{ji2023belle}
Yunjie Ji, Yong Deng, Yan Gong, Yiping Peng, Qiang Niu, Baochang Ma, and Xiangang Li.
\newblock Belle: Be everyone’s large language model engine, 2023{\natexlab{a}}.

\bibitem[Ji et~al.(2023{\natexlab{b}})Ji, Deng, Gong, Peng, Niu, Zhang, Ma, and Li]{ji2023exploring}
Yunjie Ji, Yong Deng, Yan Gong, Yiping Peng, Qiang Niu, Lei Zhang, Baochang Ma, and Xiangang Li.
\newblock Exploring the impact of instruction data scaling on large language models: An empirical study on real-world use cases.
\newblock \emph{arXiv preprint arXiv:2303.14742}, 2023{\natexlab{b}}.

\bibitem[Jiang et~al.(2023)Jiang, Liu, and Fan]{lpfl}
Jingang Jiang, Xiangyang Liu, and Chenyou Fan.
\newblock Low-parameter federated learning with large language models, 2023.

\bibitem[Jiang et~al.(2024{\natexlab{a}})Jiang, Wang, Shen, Kim, and Kim]{jiang2024survey}
Juyong Jiang, Fan Wang, Jiasi Shen, Sungju Kim, and Sunghun Kim.
\newblock A survey on large language models for code generation.
\newblock \emph{arXiv preprint arXiv:2406.00515}, 2024{\natexlab{a}}.

\bibitem[Jiang et~al.(2024{\natexlab{b}})Jiang, Ma, Wang, Yu, Yu, Wang, Ni, and Liu]{jiang2024blockchained}
Yanna Jiang, Baihe Ma, Xu~Wang, Guangsheng Yu, Ping Yu, Zhe Wang, Wei Ni, and Ren~Ping Liu.
\newblock Blockchained federated learning for internet of things: A comprehensive survey.
\newblock \emph{ACM Computing Surveys}, 56\penalty0 (10):\penalty0 1--37, 2024{\natexlab{b}}.

\bibitem[Jin et~al.(2024)Jin, Li, Liu, Gu, Wu, Jiang, He, Zhao, Tan, Gan, et~al.]{jin2024efficient}
Yizhang Jin, Jian Li, Yexin Liu, Tianjun Gu, Kai Wu, Zhengkai Jiang, Muyang He, Bo~Zhao, Xin Tan, Zhenye Gan, et~al.
\newblock Efficient multimodal large language models: A survey.
\newblock \emph{arXiv preprint arXiv:2405.10739}, 2024.

\bibitem[J{\o}rgensen et~al.(2023)J{\o}rgensen, Brandt, Hartmann, Dai, Igel, and Elliott]{jorgensen2023multifin}
Rasmus J{\o}rgensen, Oliver Brandt, Mareike Hartmann, Xiang Dai, Christian Igel, and Desmond Elliott.
\newblock Multifin: A dataset for multilingual financial nlp.
\newblock In \emph{Findings of the Association for Computational Linguistics: EACL 2023}, pp.\  894--909, 2023.

\bibitem[Karpinska et~al.(2024)Karpinska, Thai, Lo, Goyal, and Iyyer]{karpinska2024one}
Marzena Karpinska, Katherine Thai, Kyle Lo, Tanya Goyal, and Mohit Iyyer.
\newblock One thousand and one pairs: A" novel" challenge for long-context language models.
\newblock \emph{arXiv preprint arXiv:2406.16264}, 2024.

\bibitem[Kiela et~al.(2021)Kiela, Bartolo, Nie, Kaushik, Geiger, Wu, Vidgen, Prasad, Singh, Ringshia, et~al.]{kiela2021dynabench}
Douwe Kiela, Max Bartolo, Yixin Nie, Divyansh Kaushik, Atticus Geiger, Zhengxuan Wu, Bertie Vidgen, Grusha Prasad, Amanpreet Singh, Pratik Ringshia, et~al.
\newblock Dynabench: Rethinking benchmarking in nlp.
\newblock \emph{arXiv preprint arXiv:2104.14337}, 2021.

\bibitem[Kim et~al.(2023{\natexlab{a}})Kim, Yoo, and Kang]{kim2023efficient2}
Gyunyeop Kim, Joon Yoo, and Sangwoo Kang.
\newblock Efficient federated learning with pre-trained large language model using several adapter mechanisms.
\newblock \emph{Mathematics}, 11\penalty0 (21):\penalty0 4479, 2023{\natexlab{a}}.

\bibitem[Kim et~al.(2023{\natexlab{b}})Kim, Kim, Mok, Park, and Lee]{kim2023client}
Yeachan Kim, Junho Kim, Wing-Lam Mok, Jun-Hyung Park, and SangKeun Lee.
\newblock Client-customized adaptation for parameter-efficient federated learning.
\newblock In \emph{Findings of the Association for Computational Linguistics: ACL 2023}, pp.\  1159--1172, 2023{\natexlab{b}}.

\bibitem[Kim et~al.(2024)Kim, Wu, Abdulle, and Wu]{kim2024medexqa}
Yunsoo Kim, Jinge Wu, Yusuf Abdulle, and Honghan Wu.
\newblock Medexqa: Medical question answering benchmark with multiple explanations.
\newblock \emph{arXiv preprint arXiv:2406.06331}, 2024.

\bibitem[Kirkpatrick et~al.(2017)Kirkpatrick, Pascanu, Rabinowitz, Veness, Desjardins, Rusu, Milan, Quan, Ramalho, Grabska-Barwinska, et~al.]{kirkpatrick2017overcoming}
James Kirkpatrick, Razvan Pascanu, Neil Rabinowitz, Joel Veness, Guillaume Desjardins, Andrei~A Rusu, Kieran Milan, John Quan, Tiago Ramalho, Agnieszka Grabska-Barwinska, et~al.
\newblock Overcoming catastrophic forgetting in neural networks.
\newblock \emph{Proceedings of the national academy of sciences}, 114\penalty0 (13):\penalty0 3521--3526, 2017.

\bibitem[Ko et~al.(2022)Ko, Lee, Park, and Choi]{ko2022survey}
Hyeyoung Ko, Suyeon Lee, Yoonseo Park, and Anna Choi.
\newblock A survey of recommendation systems: recommendation models, techniques, and application fields.
\newblock \emph{Electronics}, 11\penalty0 (1):\penalty0 141, 2022.

\bibitem[Kocetkov et~al.(2022)Kocetkov, Li, Allal, Li, Mou, Ferrandis, Jernite, Mitchell, Hughes, Wolf, et~al.]{kocetkov2022stack}
Denis Kocetkov, Raymond Li, Loubna~Ben Allal, Jia Li, Chenghao Mou, Carlos~Mu{\~n}oz Ferrandis, Yacine Jernite, Margaret Mitchell, Sean Hughes, Thomas Wolf, et~al.
\newblock The stack: 3 tb of permissively licensed source code.
\newblock \emph{arXiv preprint arXiv:2211.15533}, 2022.

\bibitem[Koncel-Kedziorski et~al.(2023)Koncel-Kedziorski, Krumdick, Lai, Reddy, Lovering, and Tanner]{koncel2023bizbench}
Rik Koncel-Kedziorski, Michael Krumdick, Viet Lai, Varshini Reddy, Charles Lovering, and Chris Tanner.
\newblock Bizbench: A quantitative reasoning benchmark for business and finance.
\newblock \emph{arXiv preprint arXiv:2311.06602}, 2023.

\bibitem[Koo et~al.(2024)Koo, Jang, and Ok]{koo2024towards}
Jabin Koo, Minwoo Jang, and Jungseul Ok.
\newblock Towards robust and efficient federated low-rank adaptation with heterogeneous clients.
\newblock \emph{arXiv preprint arXiv:2410.22815}, 2024.

\bibitem[Kornblith et~al.(2019)Kornblith, Shlens, and Le]{kornblith2019better}
Simon Kornblith, Jonathon Shlens, and Quoc~V Le.
\newblock Do better imagenet models transfer better?
\newblock In \emph{Proceedings of the IEEE/CVF conference on computer vision and pattern recognition}, pp.\  2661--2671, 2019.

\bibitem[Kou et~al.(2024)Kou, Lin, Tang, Xu, Ye, Leng, Wang, Li, Chen, Zhu, et~al.]{kou2024pfedlvm}
Wei-Bin Kou, Qingfeng Lin, Ming Tang, Sheng Xu, Rongguang Ye, Yang Leng, Shuai Wang, Guofa Li, Zhenyu Chen, Guangxu Zhu, et~al.
\newblock pfedlvm: A large vision model (lvm)-driven and latent feature-based personalized federated learning framework in autonomous driving.
\newblock \emph{arXiv preprint arXiv:2405.04146}, 2024.

\bibitem[Kou et~al.(2025)Kou, Lin, Tang, Ye, Wang, Zhu, and Wu]{kou2025fast}
Wei-Bin Kou, Qingfeng Lin, Ming Tang, Rongguang Ye, Shuai Wang, Guangxu Zhu, and Yik-Chung Wu.
\newblock Fast-convergent and communication-alleviated heterogeneous hierarchical federated learning in autonomous driving.
\newblock \emph{IEEE Transactions on Intelligent Transportation Systems}, 2025.

\bibitem[Kuang et~al.(2023)Kuang, Qian, Li, Chen, Gao, Pan, Xie, Li, Ding, and Zhou]{fsllm}
Weirui Kuang, Bingchen Qian, Zitao Li, Daoyuan Chen, Dawei Gao, Xuchen Pan, Yuexiang Xie, Yaliang Li, Bolin Ding, and Jingren Zhou.
\newblock Federatedscope-llm: A comprehensive package for fine-tuning large language models in federated learning, 2023.

\bibitem[Kumar et~al.(2013)Kumar, Liu, Lu, and Bhargava]{kumar2013survey}
Karthik Kumar, Jibang Liu, Yung-Hsiang Lu, and Bharat Bhargava.
\newblock A survey of computation offloading for mobile systems.
\newblock \emph{Mobile networks and Applications}, 18:\penalty0 129--140, 2013.

\bibitem[Kuo et~al.(2024)Kuo, Raje, Rajesh, and Smith]{kuo2024federated}
Kevin Kuo, Arian Raje, Kousik Rajesh, and Virginia Smith.
\newblock Federated lora with sparse communication.
\newblock \emph{arXiv preprint arXiv:2406.05233}, 2024.

\bibitem[Kuratov et~al.(2024)Kuratov, Bulatov, Anokhin, Rodkin, Sorokin, Sorokin, and Burtsev]{kuratov2024babilong}
Yury Kuratov, Aydar Bulatov, Petr Anokhin, Ivan Rodkin, Dmitry Sorokin, Artyom Sorokin, and Mikhail Burtsev.
\newblock Babilong: Testing the limits of llms with long context reasoning-in-a-haystack.
\newblock \emph{Advances in Neural Information Processing Systems}, 37:\penalty0 106519--106554, 2024.

\bibitem[Lai et~al.(2023)Lai, Li, Wang, Zhang, Zhong, Zettlemoyer, Yih, Fried, Wang, and Yu]{lai2023ds}
Yuhang Lai, Chengxi Li, Yiming Wang, Tianyi Zhang, Ruiqi Zhong, Luke Zettlemoyer, Wen-tau Yih, Daniel Fried, Sida Wang, and Tao Yu.
\newblock Ds-1000: A natural and reliable benchmark for data science code generation.
\newblock In \emph{International Conference on Machine Learning}, pp.\  18319--18345. PMLR, 2023.

\bibitem[Lee et~al.(2024{\natexlab{a}})Lee, Chen, Dai, Dua, Sachan, Boratko, Luan, Arnold, Perot, Dalmia, et~al.]{lee2024can}
Jinhyuk Lee, Anthony Chen, Zhuyun Dai, Dheeru Dua, Devendra~Singh Sachan, Michael Boratko, Yi~Luan, S{\'e}bastien~MR Arnold, Vincent Perot, Siddharth Dalmia, et~al.
\newblock Can long-context language models subsume retrieval, rag, sql, and more?
\newblock \emph{arXiv preprint arXiv:2406.13121}, 2024{\natexlab{a}}.

\bibitem[Lee et~al.(2024{\natexlab{b}})Lee, Yoon, Jang, Lee, Song, Kim, and Kang]{lee2024ethic}
Taewhoo Lee, Chanwoong Yoon, Kyochul Jang, Donghyeon Lee, Minju Song, Hyunjae Kim, and Jaewoo Kang.
\newblock Ethic: Evaluating large language models on long-context tasks with high information coverage.
\newblock \emph{arXiv preprint arXiv:2410.16848}, 2024{\natexlab{b}}.

\bibitem[Lei et~al.(2023)Lei, Li, Cheng, Ding, and Jiang]{lei2023cfbenchmark}
Yang Lei, Jiangtong Li, Dawei Cheng, Zhijun Ding, and Changjun Jiang.
\newblock Cfbenchmark: Chinese financial assistant benchmark for large language model.
\newblock \emph{arXiv preprint arXiv:2311.05812}, 2023.

\bibitem[Lester et~al.(2021)Lester, Al-Rfou, and Constant]{lester2021power}
Brian Lester, Rami Al-Rfou, and Noah Constant.
\newblock The power of scale for parameter-efficient prompt tuning.
\newblock \emph{arXiv preprint arXiv:2104.08691}, 2021.

\bibitem[Li et~al.(2023{\natexlab{a}})Li, Shao, Xie, Sheng, Zheng, Gonzalez, Stoica, Ma, and Zhang]{li2023long}
Dacheng Li, Rulin Shao, Anze Xie, Ying Sheng, Lianmin Zheng, Joseph Gonzalez, Ion Stoica, Xuezhe Ma, and Hao Zhang.
\newblock How long can context length of open-source llms truly promise?
\newblock In \emph{NeurIPS 2023 Workshop on Instruction Tuning and Instruction Following}, 2023{\natexlab{a}}.

\bibitem[Li et~al.(2023{\natexlab{b}})Li, Wu, Sun, Shen, Wu, and Tao]{li2023visual}
Guanghao Li, Wansen Wu, Yan Sun, Li~Shen, Baoyuan Wu, and Dacheng Tao.
\newblock Visual prompt based personalized federated learning.
\newblock \emph{arXiv preprint arXiv:2303.08678}, 2023{\natexlab{b}}.

\bibitem[Li et~al.(2024{\natexlab{a}})Li, Chen, Yang, Ai, Jia, Liu, Lin, Wu, Yuan, Hu, et~al.]{li2024legalagentbench}
Haitao Li, Junjie Chen, Jingli Yang, Qingyao Ai, Wei Jia, Youfeng Liu, Kai Lin, Yueyue Wu, Guozhi Yuan, Yiran Hu, et~al.
\newblock Legalagentbench: Evaluating llm agents in legal domain.
\newblock \emph{arXiv preprint arXiv:2412.17259}, 2024{\natexlab{a}}.

\bibitem[Li et~al.(2024{\natexlab{b}})Li, Chen, Ai, Wu, Zhang, and Liu]{li2024lexeval}
Haitao Li, You Chen, Qingyao Ai, Yueyue Wu, Ruizhe Zhang, and Yiqun Liu.
\newblock Lexeval: A comprehensive chinese legal benchmark for evaluating large language models.
\newblock \emph{arXiv preprint arXiv:2409.20288}, 2024{\natexlab{b}}.

\bibitem[Li et~al.(2022{\natexlab{a}})Li, Wang, Wu, and Zhang]{li2022federated}
Heju Li, Rui Wang, Jun Wu, and Wei Zhang.
\newblock Federated edge learning via reconfigurable intelligent surface with one-bit quantization.
\newblock In \emph{GLOBECOM 2022-2022 IEEE Global Communications Conference}, pp.\  1055--1060. IEEE, 2022{\natexlab{a}}.

\bibitem[Li et~al.(2022{\natexlab{b}})Li, Wang, Zhang, and Wu]{li2022one}
Heju Li, Rui Wang, Wei Zhang, and Jun Wu.
\newblock One bit aggregation for federated edge learning with reconfigurable intelligent surface: Analysis and optimization.
\newblock \emph{IEEE Transactions on Wireless Communications}, 22\penalty0 (2):\penalty0 872--888, 2022{\natexlab{b}}.

\bibitem[Li et~al.(2023{\natexlab{c}})Li, Wang, Wu, Zhang, and Soto]{li2023reconfigurable}
Heju Li, Rui Wang, Jun Wu, Wei Zhang, and Ismael Soto.
\newblock Reconfigurable intelligent surface empowered federated edge learning with statistical csi.
\newblock \emph{IEEE Transactions on Wireless Communications}, 23\penalty0 (6):\penalty0 6595--6608, 2023{\natexlab{c}}.

\bibitem[Li et~al.(2025)Li, Wang, Jiang, and Liu]{li2025star}
Heju Li, Rui Wang, Mingyang Jiang, and Jianquan Liu.
\newblock Star-ris empowered heterogeneous federated edge learning with flexible aggregation.
\newblock \emph{IEEE Internet of Things Journal}, 2025.

\bibitem[Li et~al.(2024{\natexlab{c}})Li, Huang, Wang, and Shi]{li2024global}
Hongxia Li, Wei Huang, Jingya Wang, and Ye~Shi.
\newblock Global and local prompts cooperation via optimal transport for federated learning.
\newblock In \emph{Proceedings of the IEEE/CVF Conference on Computer Vision and Pattern Recognition}, pp.\  12151--12161, 2024{\natexlab{c}}.

\bibitem[Li et~al.(2024{\natexlab{d}})Li, Lu, Fei, Luo, Dai, Xia, Jin, Gan, Qi, Fu, et~al.]{li2024survey}
Jian Li, Weiheng Lu, Hao Fei, Meng Luo, Ming Dai, Min Xia, Yizhang Jin, Zhenye Gan, Ding Qi, Chaoyou Fu, et~al.
\newblock A survey on benchmarks of multimodal large language models.
\newblock \emph{arXiv preprint arXiv:2408.08632}, 2024{\natexlab{d}}.

\bibitem[Li et~al.(2023{\natexlab{d}})Li, Wang, Wu, Zhang, Xu, Fu, Tiwari, Wan, and Wang]{li2023huatuo}
Jianquan Li, Xidong Wang, Xiangbo Wu, Zhiyi Zhang, Xiaolong Xu, Jie Fu, Prayag Tiwari, Xiang Wan, and Benyou Wang.
\newblock Huatuo-26m, a large-scale chinese medical qa dataset.
\newblock \emph{arXiv preprint arXiv:2305.01526}, 2023{\natexlab{d}}.

\bibitem[Li et~al.(2023{\natexlab{e}})Li, Wang, Zheng, and Zhang]{li2023loogle}
Jiaqi Li, Mengmeng Wang, Zilong Zheng, and Muhan Zhang.
\newblock Loogle: Can long-context language models understand long contexts?
\newblock \emph{arXiv preprint arXiv:2311.04939}, 2023{\natexlab{e}}.

\bibitem[Li et~al.(2023{\natexlab{f}})Li, Hui, Qu, Yang, Li, Li, Wang, Qin, Geng, Huo, et~al.]{li2023can}
Jinyang Li, Binyuan Hui, Ge~Qu, Jiaxi Yang, Binhua Li, Bowen Li, Bailin Wang, Bowen Qin, Ruiying Geng, Nan Huo, et~al.
\newblock Can llm already serve as a database interface? a big bench for large-scale database grounded text-to-sqls.
\newblock \emph{Advances in Neural Information Processing Systems}, 36:\penalty0 42330--42357, 2023{\natexlab{f}}.

\bibitem[Li et~al.(2024{\natexlab{e}})Li, Ye, Fang, Zhao, Chan, Ngai, and Voigt]{li2024synergizing}
Shenghui Li, Fanghua Ye, Meng Fang, Jiaxu Zhao, Yun-Hin Chan, Edith C-H Ngai, and Thiemo Voigt.
\newblock Synergizing foundation models and federated learning: A survey.
\newblock \emph{arXiv preprint arXiv:2406.12844}, 2024{\natexlab{e}}.

\bibitem[Li et~al.(2023{\natexlab{g}})Li, Tian, Tam, Ma, and Li]{li2023breaking}
Shitian Li, Chunlin Tian, Kahou Tam, Rui Ma, and Li~Li.
\newblock Breaking on-device training memory wall: A systematic survey.
\newblock \emph{arXiv preprint arXiv:2306.10388}, 2023{\natexlab{g}}.

\bibitem[Li et~al.(2020)Li, Sahu, Zaheer, Sanjabi, Talwalkar, and Smith]{li2020federated}
Tian Li, Anit~Kumar Sahu, Manzil Zaheer, Maziar Sanjabi, Ameet Talwalkar, and Virginia Smith.
\newblock Federated optimization in heterogeneous networks.
\newblock \emph{Proceedings of Machine learning and systems}, 2:\penalty0 429--450, 2020.

\bibitem[Li et~al.(2024{\natexlab{f}})Li, Zhang, Do, Yue, and Chen]{li2024long}
Tianle Li, Ge~Zhang, Quy~Duc Do, Xiang Yue, and Wenhu Chen.
\newblock Long-context llms struggle with long in-context learning.
\newblock \emph{arXiv preprint arXiv:2404.02060}, 2024{\natexlab{f}}.

\bibitem[Li et~al.(2023{\natexlab{h}})Li, Wang, Ding, and Chen]{li2023large}
Yinheng Li, Shaofei Wang, Han Ding, and Hang Chen.
\newblock Large language models in finance: A survey.
\newblock In \emph{Proceedings of the fourth ACM international conference on AI in finance}, pp.\  374--382, 2023{\natexlab{h}}.

\bibitem[Li et~al.(2022{\natexlab{c}})Li, Choi, Chung, Kushman, Schrittwieser, Leblond, Eccles, Keeling, Gimeno, Dal~Lago, Hubert, Choy, de~Masson~d'Autume, Babuschkin, Chen, Huang, Welbl, Gowal, Cherepanov, Molloy, Mankowitz, Sutherland~Robson, Kohli, de~Freitas, Kavukcuoglu, and Vinyals]{li2022competition}
Yujia Li, David Choi, Junyoung Chung, Nate Kushman, Julian Schrittwieser, R{\'e}mi Leblond, Tom Eccles, James Keeling, Felix Gimeno, Agustin Dal~Lago, Thomas Hubert, Peter Choy, Cyprien de~Masson~d'Autume, Igor Babuschkin, Xinyun Chen, Po-Sen Huang, Johannes Welbl, Sven Gowal, Alexey Cherepanov, James Molloy, Daniel Mankowitz, Esme Sutherland~Robson, Pushmeet Kohli, Nando de~Freitas, Koray Kavukcuoglu, and Oriol Vinyals.
\newblock Competition-level code generation with alphacode.
\newblock \emph{arXiv preprint arXiv:2203.07814}, 2022{\natexlab{c}}.

\bibitem[Li et~al.(2023{\natexlab{i}})Li, Li, Zhang, Dan, Jiang, and Zhang]{li2023chatdoctor}
Yunxiang Li, Zihan Li, Kai Zhang, Ruilong Dan, Steve Jiang, and You Zhang.
\newblock Chatdoctor: A medical chat model fine-tuned on a large language model meta-ai (llama) using medical domain knowledge.
\newblock \emph{Cureus}, 15\penalty0 (6), 2023{\natexlab{i}}.

\bibitem[Lian et~al.(2023)Lian, Goodson, Pentland, Cook, Vong, and “Teknium”]{lian2023openorca}
Wing Lian, Bleys Goodson, Eugene Pentland, Austin Cook, Chanvichet Vong, and “Teknium”.
\newblock Openorca: An open dataset of gpt augmented flan reasoning traces, 2023.

\bibitem[Liang et~al.(2022)Liang, Bommasani, Lee, Tsipras, Soylu, Yasunaga, Zhang, Narayanan, Wu, Kumar, et~al.]{liang2022holistic}
Percy Liang, Rishi Bommasani, Tony Lee, Dimitris Tsipras, Dilara Soylu, Michihiro Yasunaga, Yian Zhang, Deepak Narayanan, Yuhuai Wu, Ananya Kumar, et~al.
\newblock Holistic evaluation of language models.
\newblock \emph{arXiv preprint arXiv:2211.09110}, 2022.

\bibitem[Liang et~al.(2024)Liang, Xu, Hong, Shang, Wang, Fu, and Liu]{liang2024survey}
Zijing Liang, Yanjie Xu, Yifan Hong, Penghui Shang, Qi~Wang, Qiang Fu, and Ke~Liu.
\newblock A survey of multimodel large language models.
\newblock In \emph{Proceedings of the 3rd International Conference on Computer, Artificial Intelligence and Control Engineering}, pp.\  405--409, 2024.

\bibitem[Lin et~al.(2024)Lin, Liu, Wu, Cheng, Cai, Wong, and Tang]{lin2024fedlppa}
Li~Lin, Yixiang Liu, Jiewei Wu, Pujin Cheng, Zhiyuan Cai, Kenneth~KY Wong, and Xiaoying Tang.
\newblock Fedlppa: Learning personalized prompt and aggregation for federated weakly-supervised medical image segmentation.
\newblock \emph{arXiv preprint arXiv:2402.17502}, 2024.

\bibitem[Lin et~al.(2021)Lin, Hilton, and Evans]{lin2021truthfulqa}
Stephanie Lin, Jacob Hilton, and Owain Evans.
\newblock Truthfulqa: Measuring how models mimic human falsehoods.
\newblock \emph{arXiv preprint arXiv:2109.07958}, 2021.

\bibitem[Lin et~al.(2023)Lin, Sun, Shi, Wang, Huang, Shen, and Tao]{lin2023efficient}
Zihao Lin, Yan Sun, Yifan Shi, Xueqian Wang, Lifu Huang, Li~Shen, and Dacheng Tao.
\newblock Efficient federated prompt tuning for black-box large pre-trained models.
\newblock \emph{arXiv preprint arXiv:2310.03123}, 2023.

\bibitem[Ling et~al.(2025)Ling, Liu, Yan, Yang, Lin, Fan, Shen, Du, and Chen]{ling2025longreason}
Zhan Ling, Kang Liu, Kai Yan, Yifan Yang, Weijian Lin, Ting-Han Fan, Lingfeng Shen, Zhengyin Du, and Jiecao Chen.
\newblock Longreason: A synthetic long-context reasoning benchmark via context expansion.
\newblock \emph{arXiv preprint arXiv:2501.15089}, 2025.

\bibitem[Liu et~al.(2023{\natexlab{a}})Liu, Jin, Ren, Yu, Dong, Peng, Zhang, Peng, Zhang, Lyu, et~al.]{liu2023m3ke}
Chuang Liu, Renren Jin, Yuqi Ren, Linhao Yu, Tianyu Dong, Xiaohan Peng, Shuting Zhang, Jianxiang Peng, Peiyi Zhang, Qingqing Lyu, et~al.
\newblock M3ke: A massive multi-level multi-subject knowledge evaluation benchmark for chinese large language models.
\newblock \emph{arXiv preprint arXiv:2305.10263}, 2023{\natexlab{a}}.

\bibitem[Liu et~al.(2023{\natexlab{b}})Liu, Li, Wu, and Lee]{liu2023visual}
Haotian Liu, Chunyuan Li, Qingyang Wu, and Yong~Jae Lee.
\newblock Visual instruction tuning.
\newblock \emph{Advances in neural information processing systems}, 36:\penalty0 34892--34916, 2023{\natexlab{b}}.

\bibitem[Liu et~al.(2024{\natexlab{a}})Liu, Zheng, Qiao, Duan, Fei, Zhou, Zhang, Zhang, Lin, and Chen]{liu2024mathbench}
Hongwei Liu, Zilong Zheng, Yuxuan Qiao, Haodong Duan, Zhiwei Fei, Fengzhe Zhou, Wenwei Zhang, Songyang Zhang, Dahua Lin, and Kai Chen.
\newblock Mathbench: Evaluating the theory and application proficiency of llms with a hierarchical mathematics benchmark.
\newblock \emph{arXiv preprint arXiv:2405.12209}, 2024{\natexlab{a}}.

\bibitem[Liu et~al.(2023{\natexlab{c}})Liu, Zhan, Zhang, Luo, Chen, Wei, and Xu]{liu2023federated1}
Jiale Liu, Yu-Wei Zhan, Chong-Yu Zhang, Xin Luo, Zhen-Duo Chen, Yinwei Wei, and Xin-Shun Xu.
\newblock Federated class-incremental learning with prompting.
\newblock \emph{arXiv preprint arXiv:2310.08948}, 2023{\natexlab{c}}.

\bibitem[Liu et~al.(2024{\natexlab{b}})Liu, Tian, Daita, Wei, Ding, Wang, Yang, and Zhang]{liu2024repoqa}
Jiawei Liu, Jia~Le Tian, Vijay Daita, Yuxiang Wei, Yifeng Ding, Yuhan~Katherine Wang, Jun Yang, and Lingming Zhang.
\newblock Repoqa: Evaluating long context code understanding.
\newblock \emph{arXiv preprint arXiv:2406.06025}, 2024{\natexlab{b}}.

\bibitem[Liu et~al.(2023{\natexlab{d}})Liu, Zhou, Hua, Chong, Tian, Liu, Wang, You, Guo, Zhu, et~al.]{liu2023benchmarking}
Junling Liu, Peilin Zhou, Yining Hua, Dading Chong, Zhongyu Tian, Andrew Liu, Helin Wang, Chenyu You, Zhenhua Guo, Lei Zhu, et~al.
\newblock Benchmarking large language models on cmexam-a comprehensive chinese medical exam dataset.
\newblock \emph{Advances in Neural Information Processing Systems}, 36:\penalty0 52430--52452, 2023{\natexlab{d}}.

\bibitem[Liu et~al.(2023{\natexlab{e}})Liu, Pang, and Fan]{liu2023federated}
Xiangyang Liu, Tianqi Pang, and Chenyou Fan.
\newblock Federated prompting and chain-of-thought reasoning for improving llms answering.
\newblock In \emph{International Conference on Knowledge Science, Engineering and Management}, pp.\  3--11. Springer, 2023{\natexlab{e}}.

\bibitem[Liu et~al.(2023{\natexlab{f}})Liu, Bi, Li, Chen, Yang, and Sun]{liu2023communication}
Yi~Liu, Xiaohan Bi, Lei Li, Sishuo Chen, Wenkai Yang, and Xu~Sun.
\newblock Communication efficient federated learning for multilingual neural machine translation with adapter.
\newblock \emph{arXiv preprint arXiv:2305.12449}, 2023{\natexlab{f}}.

\bibitem[Liu et~al.(2024{\natexlab{c}})Liu, Luo, and Zhu]{liu2024fedfms}
Yuxi Liu, Guibo Luo, and Yuesheng Zhu.
\newblock Fedfms: Exploring federated foundation models for medical image segmentation.
\newblock In \emph{International Conference on Medical Image Computing and Computer-Assisted Intervention}, pp.\  283--293. Springer, 2024{\natexlab{c}}.

\bibitem[Lou et~al.(2024)Lou, Zhang, and Yin]{lou2024large}
Renze Lou, Kai Zhang, and Wenpeng Yin.
\newblock Large language model instruction following: A survey of progresses and challenges.
\newblock \emph{Computational Linguistics}, 50\penalty0 (3):\penalty0 1053--1095, 2024.

\bibitem[Lu et~al.(2023{\natexlab{a}})Lu, Wu, Liang, Xu, He, Geng, Han, Xin, and Xiao]{lu2023bbt}
Dakuan Lu, Hengkui Wu, Jiaqing Liang, Yipei Xu, Qianyu He, Yipeng Geng, Mengkun Han, Yingsi Xin, and Yanghua Xiao.
\newblock Bbt-fin: Comprehensive construction of chinese financial domain pre-trained language model, corpus and benchmark.
\newblock \emph{arXiv preprint arXiv:2302.09432}, 2023{\natexlab{a}}.

\bibitem[Lu et~al.(2022{\natexlab{a}})Lu, Mishra, Xia, Qiu, Chang, Zhu, Tafjord, Clark, and Kalyan]{lu2022learn}
Pan Lu, Swaroop Mishra, Tanglin Xia, Liang Qiu, Kai-Wei Chang, Song-Chun Zhu, Oyvind Tafjord, Peter Clark, and Ashwin Kalyan.
\newblock Learn to explain: Multimodal reasoning via thought chains for science question answering.
\newblock \emph{Advances in Neural Information Processing Systems}, 35:\penalty0 2507--2521, 2022{\natexlab{a}}.

\bibitem[Lu et~al.(2022{\natexlab{b}})Lu, Qiu, Chang, Wu, Zhu, Rajpurohit, Clark, and Kalyan]{lu2022dynamic}
Pan Lu, Liang Qiu, Kai-Wei Chang, Ying~Nian Wu, Song-Chun Zhu, Tanmay Rajpurohit, Peter Clark, and Ashwin Kalyan.
\newblock Dynamic prompt learning via policy gradient for semi-structured mathematical reasoning.
\newblock \emph{arXiv preprint arXiv:2209.14610}, 2022{\natexlab{b}}.

\bibitem[Lu et~al.(2021)Lu, Guo, Ren, Huang, Svyatkovskiy, Blanco, Clement, Drain, Jiang, Tang, et~al.]{lu2021codexglue}
Shuai Lu, Daya Guo, Shuo Ren, Junjie Huang, Alexey Svyatkovskiy, Ambrosio Blanco, Colin Clement, Dawn Drain, Daxin Jiang, Duyu Tang, et~al.
\newblock Codexglue: A machine learning benchmark dataset for code understanding and generation.
\newblock \emph{arXiv preprint arXiv:2102.04664}, 2021.

\bibitem[Lu et~al.(2023{\natexlab{b}})Lu, Hu, Wang, and Xie]{lu2023fedclip}
Wang Lu, Xixu Hu, Jindong Wang, and Xing Xie.
\newblock Fedclip: Fast generalization and personalization for clip in federated learning.
\newblock \emph{arXiv preprint arXiv:2302.13485}, 2023{\natexlab{b}}.

\bibitem[Luo et~al.(2024)Luo, Chen, and Wu]{luo2024mixture}
Jun Luo, Chen Chen, and Shandong Wu.
\newblock Mixture of experts made personalized: Federated prompt learning for vision-language models.
\newblock \emph{arXiv preprint arXiv:2410.10114}, 2024.

\bibitem[Ma et~al.(2024{\natexlab{a}})Ma, Tian, Li, and Xu]{ma2024fedmg}
Jialiang Ma, Chunlin Tian, Li~Li, and Chengzhong Xu.
\newblock Fedmg: A federated multi-global optimization framework for autonomous driving control.
\newblock In \emph{2024 IEEE/ACM 32nd International Symposium on Quality of Service (IWQoS)}, pp.\  1--10. IEEE, 2024{\natexlab{a}}.

\bibitem[Ma et~al.(2023)Ma, Fang, and Wang]{ma2023llm}
Xinyin Ma, Gongfan Fang, and Xinchao Wang.
\newblock Llm-pruner: On the structural pruning of large language models.
\newblock \emph{Advances in neural information processing systems}, 36:\penalty0 21702--21720, 2023.

\bibitem[Ma et~al.(2024{\natexlab{b}})Ma, Cheng, Wang, Zhong, Xu, and Wang]{ma2024fedhpl}
Yuting Ma, Lechao Cheng, Yaxiong Wang, Zhun Zhong, Xiaohua Xu, and Meng Wang.
\newblock Fedhpl: Efficient heterogeneous federated learning with prompt tuning and logit distillation.
\newblock \emph{arXiv preprint arXiv:2405.17267}, 2024{\natexlab{b}}.

\bibitem[Maekawa et~al.(2024)Maekawa, Iso, and Bhutani]{maekawa2024holistic}
Seiji Maekawa, Hayate Iso, and Nikita Bhutani.
\newblock Holistic reasoning with long-context lms: A benchmark for database operations on massive textual data.
\newblock \emph{arXiv preprint arXiv:2410.11996}, 2024.

\bibitem[Maia et~al.(2018)Maia, Handschuh, Freitas, Davis, McDermott, Zarrouk, and Balahur]{maia201818}
Macedo Maia, Siegfried Handschuh, Andr{\'e} Freitas, Brian Davis, Ross McDermott, Manel Zarrouk, and Alexandra Balahur.
\newblock Www'18 open challenge: financial opinion mining and question answering.
\newblock In \emph{Companion proceedings of the the web conference 2018}, pp.\  1941--1942, 2018.

\bibitem[Malladi et~al.(2023)Malladi, Gao, Nichani, Damian, Lee, Chen, and Arora]{malladi2023fine}
Sadhika Malladi, Tianyu Gao, Eshaan Nichani, Alex Damian, Jason~D Lee, Danqi Chen, and Sanjeev Arora.
\newblock Fine-tuning language models with just forward passes.
\newblock \emph{Advances in Neural Information Processing Systems}, 36:\penalty0 53038--53075, 2023.

\bibitem[Malo et~al.(2014)Malo, Sinha, Korhonen, Wallenius, and Takala]{malo2014good}
Pekka Malo, Ankur Sinha, Pekka Korhonen, Jyrki Wallenius, and Pyry Takala.
\newblock Good debt or bad debt: Detecting semantic orientations in economic texts.
\newblock \emph{Journal of the Association for Information Science and Technology}, 65\penalty0 (4):\penalty0 782--796, 2014.

\bibitem[Mao et~al.(2024)Mao, Kim, and Zhou]{mao2024champ}
Yujun Mao, Yoon Kim, and Yilun Zhou.
\newblock Champ: A competition-level dataset for fine-grained analyses of llms' mathematical reasoning capabilities.
\newblock \emph{arXiv preprint arXiv:2401.06961}, 2024.

\bibitem[McMahan et~al.(2017)McMahan, Moore, Ramage, Hampson, and y~Arcas]{mcmahan2017communication}
Brendan McMahan, Eider Moore, Daniel Ramage, Seth Hampson, and Blaise~Aguera y~Arcas.
\newblock Communication-efficient learning of deep networks from decentralized data.
\newblock In \emph{Artificial intelligence and statistics}, pp.\  1273--1282. PMLR, 2017.

\bibitem[Mesnard et~al.(2024)Mesnard, Hardin, Dadashi, Bhupatiraju, Pathak, Sifre, Rivi{\`{e}}re, Kale, Love, Tafti, Hussenot, Chowdhery, Roberts, Barua, Botev, Castro{-}Ros, Slone, H{\'{e}}liou, Tacchetti, Bulanova, Paterson, Tsai, Shahriari, Lan, Choquette{-}Choo, Crepy, Cer, Ippolito, Reid, Buchatskaya, Ni, Noland, Yan, Tucker, Muraru, Rozhdestvenskiy, Michalewski, Tenney, Grishchenko, Austin, Keeling, Labanowski, Lespiau, Stanway, Brennan, Chen, Ferret, Chiu, and et~al.]{gemma}
Thomas Mesnard, Cassidy Hardin, Robert Dadashi, Surya Bhupatiraju, Shreya Pathak, Laurent Sifre, Morgane Rivi{\`{e}}re, Mihir~Sanjay Kale, Juliette Love, Pouya Tafti, L{\'{e}}onard Hussenot, Aakanksha Chowdhery, Adam Roberts, Aditya Barua, Alex Botev, Alex Castro{-}Ros, Ambrose Slone, Am{\'{e}}lie H{\'{e}}liou, Andrea Tacchetti, Anna Bulanova, Antonia Paterson, Beth Tsai, Bobak Shahriari, Charline~Le Lan, Christopher~A. Choquette{-}Choo, Cl{\'{e}}ment Crepy, Daniel Cer, Daphne Ippolito, David Reid, Elena Buchatskaya, Eric Ni, Eric Noland, Geng Yan, George Tucker, George{-}Christian Muraru, Grigory Rozhdestvenskiy, Henryk Michalewski, Ian Tenney, Ivan Grishchenko, Jacob Austin, James Keeling, Jane Labanowski, Jean{-}Baptiste Lespiau, Jeff Stanway, Jenny Brennan, Jeremy Chen, Johan Ferret, Justin Chiu, and et~al.
\newblock Gemma: Open models based on gemini research and technology.
\newblock \emph{CoRR}, abs/2403.08295, 2024.
\newblock \doi{10.48550/ARXIV.2403.08295}.
\newblock URL \url{https://doi.org/10.48550/arXiv.2403.08295}.

\bibitem[Mishra et~al.(2022)Mishra, Finlayson, Lu, Tang, Welleck, Baral, Rajpurohit, Tafjord, Sabharwal, Clark, et~al.]{mishra2022lila}
Swaroop Mishra, Matthew Finlayson, Pan Lu, Leonard Tang, Sean Welleck, Chitta Baral, Tanmay Rajpurohit, Oyvind Tafjord, Ashish Sabharwal, Peter Clark, et~al.
\newblock Lila: A unified benchmark for mathematical reasoning.
\newblock \emph{arXiv preprint arXiv:2210.17517}, 2022.

\bibitem[Mitra et~al.(2024)Mitra, Khanpour, Rosset, and Awadallah]{mitra2024orcamath}
Arindam Mitra, Hamed Khanpour, Corby Rosset, and Ahmed Awadallah.
\newblock Orca-math: Unlocking the potential of slms in grade school math, 2024.

\bibitem[Muennighoff et~al.(2023)Muennighoff, Liu, Zebaze, Zheng, Hui, Zhuo, Singh, Tang, Von~Werra, and Longpre]{muennighoff2023octopack}
Niklas Muennighoff, Qian Liu, Armel Zebaze, Qinkai Zheng, Binyuan Hui, Terry~Yue Zhuo, Swayam Singh, Xiangru Tang, Leandro Von~Werra, and Shayne Longpre.
\newblock Octopack: Instruction tuning code large language models.
\newblock In \emph{NeurIPS 2023 Workshop on Instruction Tuning and Instruction Following}, 2023.

\bibitem[Naseer \& Nandakumar()Naseer and Nandakumar]{naseer2024probing}
Muzammal Naseer and Karthik Nandakumar.
\newblock Probing the efficacy of federated parameter-efficient fine-tuning of vision transformers for medical image.
\newblock In \emph{Medical Image Computing and Computer Assisted Intervention--MICCAI 2024 Workshops: ISIC 2024, iMIMIC 2024, EARTH 2024, DeCaF 2024, Held in Conjunction with MICCAI 2024, Marrakesh, Morocco, October 6--10, 2024, Proceedings}, pp.\  236. Springer Nature.

\bibitem[Naveed et~al.(2023)Naveed, Khan, Qiu, Saqib, Anwar, Usman, Akhtar, Barnes, and Mian]{naveed2023comprehensive}
Humza Naveed, Asad~Ullah Khan, Shi Qiu, Muhammad Saqib, Saeed Anwar, Muhammad Usman, Naveed Akhtar, Nick Barnes, and Ajmal Mian.
\newblock A comprehensive overview of large language models.
\newblock \emph{arXiv preprint arXiv:2307.06435}, 2023.

\bibitem[Nguyen et~al.(2022)Nguyen, Pham, Pathirana, Ding, Seneviratne, Lin, Dobre, and Hwang]{nguyen2022federated}
Dinh~C Nguyen, Quoc-Viet Pham, Pubudu~N Pathirana, Ming Ding, Aruna Seneviratne, Zihuai Lin, Octavia Dobre, and Won-Joo Hwang.
\newblock Federated learning for smart healthcare: A survey.
\newblock \emph{ACM Computing Surveys (Csur)}, 55\penalty0 (3):\penalty0 1--37, 2022.

\bibitem[Nie et~al.(2024)Nie, Yan, Guo, Liu, Wang, He, Zheng, Wang, Li, Sun, et~al.]{nie2024cfinbench}
Ying Nie, Binwei Yan, Tianyu Guo, Hao Liu, Haoyu Wang, Wei He, Binfan Zheng, Weihao Wang, Qiang Li, Weijian Sun, et~al.
\newblock Cfinbench: A comprehensive chinese financial benchmark for large language models.
\newblock \emph{arXiv preprint arXiv:2407.02301}, 2024.

\bibitem[Nijkamp et~al.(2022)Nijkamp, Pang, Hayashi, Tu, Wang, Zhou, Savarese, and Xiong]{nijkamp2022codegen}
Erik Nijkamp, Bo~Pang, Hiroaki Hayashi, Lifu Tu, Huan Wang, Yingbo Zhou, Silvio Savarese, and Caiming Xiong.
\newblock Codegen: An open large language model for code with multi-turn program synthesis.
\newblock \emph{arXiv preprint arXiv:2203.13474}, 2022.

\bibitem[Niklaus et~al.(2023)Niklaus, Matoshi, Rani, Galassi, St{\"u}rmer, and Chalkidis]{niklaus2023lextreme}
Joel Niklaus, Veton Matoshi, Pooja Rani, Andrea Galassi, Matthias St{\"u}rmer, and Ilias Chalkidis.
\newblock Lextreme: A multi-lingual and multi-task benchmark for the legal domain.
\newblock \emph{arXiv preprint arXiv:2301.13126}, 2023.

\bibitem[Ning et~al.(2024)Ning, Tian, Xiao, Fan, Wang, Li, Wang, and Zhou]{ning2024fedgcs}
Zhiyuan Ning, Chunlin Tian, Meng Xiao, Wei Fan, Pengyang Wang, Li~Li, Pengfei Wang, and Yuanchun Zhou.
\newblock Fedgcs: A generative framework for efficient client selection in federated learning via gradient-based optimization.
\newblock \emph{arXiv preprint arXiv:2405.06312}, 2024.

\bibitem[OpenAI(2023)]{GPT4}
OpenAI.
\newblock {GPT-4} technical report.
\newblock \emph{CoRR}, abs/2303.08774, 2023.
\newblock \doi{10.48550/ARXIV.2303.08774}.
\newblock URL \url{https://doi.org/10.48550/arXiv.2303.08774}.

\bibitem[{OpenAI}(2024)]{gpt4o}
{OpenAI}.
\newblock Hello {GPT-4o}, 2024.
\newblock URL \url{https://openai.com/index/hello-gpt-4o/}.

\bibitem[Ouyang et~al.(2023)Ouyang, Xie, Fu, Cheng, Pan, Ling, Xing, Zhou, and Huang]{ouyang2023harmony}
Xiaomin Ouyang, Zhiyuan Xie, Heming Fu, Sitong Cheng, Li~Pan, Neiwen Ling, Guoliang Xing, Jiayu Zhou, and Jianwei Huang.
\newblock Harmony: Heterogeneous multi-modal federated learning through disentangled model training.
\newblock In \emph{Proceedings of the 21st Annual International Conference on Mobile Systems, Applications and Services}, pp.\  530--543, 2023.

\bibitem[Pan et~al.(2024{\natexlab{a}})Pan, Liu, He, Gao, Zhao, Lu, Zhou, Liu, Hu, Wen, et~al.]{pan2024markllm}
Leyi Pan, Aiwei Liu, Zhiwei He, Zitian Gao, Xuandong Zhao, Yijian Lu, Binglin Zhou, Shuliang Liu, Xuming Hu, Lijie Wen, et~al.
\newblock Markllm: An open-source toolkit for llm watermarking.
\newblock \emph{arXiv preprint arXiv:2405.10051}, 2024{\natexlab{a}}.

\bibitem[Pan et~al.(2024{\natexlab{b}})Pan, Liu, Diao, Pi, Zhang, Han, and Zhang]{pan2024lisa}
Rui Pan, Xiang Liu, Shizhe Diao, Renjie Pi, Jipeng Zhang, Chi Han, and Tong Zhang.
\newblock Lisa: layerwise importance sampling for memory-efficient large language model fine-tuning.
\newblock \emph{Advances in Neural Information Processing Systems}, 37:\penalty0 57018--57049, 2024{\natexlab{b}}.

\bibitem[Penedo et~al.(2025)Penedo, Lozhkov, Kydlíček, Allal, Beeching, Lajarín, Gallouédec, Habib, Tunstall, and von Werra]{penedo2025codeforces}
Guilherme Penedo, Anton Lozhkov, Hynek Kydlíček, Loubna~Ben Allal, Edward Beeching, Agustín~Piqueres Lajarín, Quentin Gallouédec, Nathan Habib, Lewis Tunstall, and Leandro von Werra.
\newblock Codeforces cots.
\newblock \url{https://huggingface.co/datasets/open-r1/codeforces-cots}, 2025.

\bibitem[Peng et~al.(2023)Peng, Li, He, Galley, and Gao]{peng2023instruction}
Baolin Peng, Chunyuan Li, Pengcheng He, Michel Galley, and Jianfeng Gao.
\newblock Instruction tuning with gpt-4.
\newblock \emph{arXiv preprint arXiv:2304.03277}, 2023.

\bibitem[Peng et~al.(2024)Peng, Bian, and Xu]{peng2024fedmm}
Yuanzhe Peng, Jieming Bian, and Jie Xu.
\newblock Fedmm: Federated multi-modal learning with modality heterogeneity in computational pathology.
\newblock In \emph{ICASSP 2024-2024 IEEE International Conference on Acoustics, Speech and Signal Processing (ICASSP)}, pp.\  1696--1700. IEEE, 2024.

\bibitem[Pfeiffer et~al.(2020)Pfeiffer, R{\"u}ckl{\'e}, Poth, Kamath, Vuli{\'c}, Ruder, Cho, and Gurevych]{pfeiffer2020adapterhub}
Jonas Pfeiffer, Andreas R{\"u}ckl{\'e}, Clifton Poth, Aishwarya Kamath, Ivan Vuli{\'c}, Sebastian Ruder, Kyunghyun Cho, and Iryna Gurevych.
\newblock Adapterhub: A framework for adapting transformers.
\newblock \emph{arXiv preprint arXiv:2007.07779}, 2020.

\bibitem[Puppala et~al.(2024)Puppala, Hossain, Alam, and Talukder]{puppala2024scan}
Sai Puppala, Ismail Hossain, Md~Jahangir Alam, and Sajedul Talukder.
\newblock Scan: A healthcare personalized chatbot with federated learning based gpt.
\newblock In \emph{2024 IEEE 48th Annual Computers, Software, and Applications Conference (COMPSAC)}, pp.\  1945--1951. IEEE, 2024.

\bibitem[Qi et~al.(2024{\natexlab{a}})Qi, Luan, Huang, Fung, Yang, and Qian]{qi2024fdlora}
Jiaxing Qi, Zhongzhi Luan, Shaohan Huang, Carol Fung, Hailong Yang, and Depei Qian.
\newblock Fdlora: Personalized federated learning of large language model via dual lora tuning.
\newblock \emph{arXiv preprint arXiv:2406.07925}, 2024{\natexlab{a}}.

\bibitem[Qi et~al.(2024{\natexlab{b}})Qi, Xu, Guo, Wang, Zhang, and Xu]{qi2024long}
Zehan Qi, Rongwu Xu, Zhijiang Guo, Cunxiang Wang, Hao Zhang, and Wei Xu.
\newblock Long2rag: Evaluating long-context \& long-form retrieval-augmented generation with key point recall.
\newblock \emph{arXiv preprint arXiv:2410.23000}, 2024{\natexlab{b}}.

\bibitem[Qiao et~al.(2024)Qiao, Tan, Dong, Wu, Sun, Song, GongQue, Lei, Wei, Zhang, et~al.]{qiao2024we}
Runqi Qiao, Qiuna Tan, Guanting Dong, Minhui Wu, Chong Sun, Xiaoshuai Song, Zhuoma GongQue, Shanglin Lei, Zhe Wei, Miaoxuan Zhang, et~al.
\newblock We-math: Does your large multimodal model achieve human-like mathematical reasoning?
\newblock \emph{arXiv preprint arXiv:2407.01284}, 2024.

\bibitem[Qin et~al.(2023{\natexlab{a}})Qin, Liang, Ye, Zhu, Yan, Lu, Lin, Cong, Tang, Qian, et~al.]{qin2023toolllm}
Yujia Qin, Shihao Liang, Yining Ye, Kunlun Zhu, Lan Yan, Yaxi Lu, Yankai Lin, Xin Cong, Xiangru Tang, Bill Qian, et~al.
\newblock Toolllm: Facilitating large language models to master 16000+ real-world apis.
\newblock \emph{arXiv preprint arXiv:2307.16789}, 2023{\natexlab{a}}.

\bibitem[Qin et~al.(2023{\natexlab{b}})Qin, Chen, Qian, Ding, Li, and Deng]{qin2023federated}
Zhen Qin, Daoyuan Chen, Bingchen Qian, Bolin Ding, Yaliang Li, and Shuiguang Deng.
\newblock Federated full-parameter tuning of billion-sized language models with communication cost under 18 kilobytes.
\newblock \emph{arXiv preprint arXiv:2312.06353}, 2023{\natexlab{b}}.

\bibitem[Qin et~al.(2024)Qin, Wu, He, and Deng]{qin2024federated}
Zhen Qin, Zhaomin Wu, Bingsheng He, and Shuiguang Deng.
\newblock Federated data-efficient instruction tuning for large language models.
\newblock \emph{arXiv preprint arXiv:2410.10926}, 2024.

\bibitem[Qiu et~al.(2023)Qiu, Li, Mummadi, Ganesh, Li, Peng, and Lin]{qiu2023text}
Chen Qiu, Xingyu Li, Chaithanya~Kumar Mummadi, Madan~Ravi Ganesh, Zhenzhen Li, Lu~Peng, and Wan-Yi Lin.
\newblock Text-driven prompt generation for vision-language models in federated learning.
\newblock \emph{arXiv preprint arXiv:2310.06123}, 2023.

\bibitem[Quan \& Liu(2024)Quan and Liu]{quan2024econlogicqa}
Yinzhu Quan and Zefang Liu.
\newblock Econlogicqa: A question-answering benchmark for evaluating large language models in economic sequential reasoning.
\newblock \emph{arXiv preprint arXiv:2405.07938}, 2024.

\bibitem[Rasiah et~al.(2023)Rasiah, Stern, Matoshi, St{\"u}rmer, Chalkidis, Ho, and Niklaus]{rasiah2023scale}
Vishvaksenan Rasiah, Ronja Stern, Veton Matoshi, Matthias St{\"u}rmer, Ilias Chalkidis, Daniel~E Ho, and Joel Niklaus.
\newblock Scale: Scaling up the complexity for advanced language model evaluation.
\newblock \emph{arXiv preprint arXiv:2306.09237}, 2023.

\bibitem[Reddy et~al.(2024)Reddy, Koncel-Kedziorski, Lai, Krumdick, Lovering, and Tanner]{reddy2024docfinqa}
Varshini Reddy, Rik Koncel-Kedziorski, Viet~Dac Lai, Michael Krumdick, Charles Lovering, and Chris Tanner.
\newblock Docfinqa: A long-context financial reasoning dataset.
\newblock \emph{arXiv preprint arXiv:2401.06915}, 2024.

\bibitem[Ren et~al.(2024)Ren, Yu, Peng, Tang, Li, Gao, Tan, Zhao, Li, Li, et~al.]{ren2024advances}
Chao Ren, Han Yu, Hongyi Peng, Xiaoli Tang, Anran Li, Yulan Gao, Alysa~Ziying Tan, Bo~Zhao, Xiaoxiao Li, Zengxiang Li, et~al.
\newblock Advances and open challenges in federated learning with foundation models.
\newblock \emph{arXiv preprint arXiv:2404.15381}, 2024.

\bibitem[Sabah et~al.(2024)Sabah, Chen, Yang, Azam, Ahmad, and Sarwar]{sabah2024model}
Fahad Sabah, Yuwen Chen, Zhen Yang, Muhammad Azam, Nadeem Ahmad, and Raheem Sarwar.
\newblock Model optimization techniques in personalized federated learning: A survey.
\newblock \emph{Expert Systems with Applications}, 243:\penalty0 122874, 2024.

\bibitem[Sanh et~al.(2020)Sanh, Debut, Chaumond, and Wolf]{sanh2019distilbert}
Victor Sanh, Lysandre Debut, Julien Chaumond, and Thomas Wolf.
\newblock Distilbert, a distilled version of bert: smaller, faster, cheaper and lighter, 2020.
\newblock URL \url{https://arxiv.org/abs/1910.01108}.

\bibitem[Sarwar(2025)]{sarwar2025fedmentalcare}
SM~Sarwar.
\newblock Fedmentalcare: Towards privacy-preserving fine-tuned llms to analyze mental health status using federated learning framework.
\newblock \emph{arXiv preprint arXiv:2503.05786}, 2025.

\bibitem[Shabani(2024)]{shabani2024harnessing}
Naser Shabani.
\newblock Harnessing federated learning for llm fine-tuning: A distributed approach, 2024.

\bibitem[Shah et~al.(2023)Shah, Vithani, Gullapalli, and Chava]{shah2023finer}
Agam Shah, Ruchit Vithani, Abhinav Gullapalli, and Sudheer Chava.
\newblock Finer: Financial named entity recognition dataset and weak-supervision model.
\newblock \emph{arXiv e-prints}, pp.\  arXiv--2302, 2023.

\bibitem[Shah et~al.(2022)Shah, Chawla, Eidnani, Shah, Du, Chava, Raman, Smiley, Chen, and Yang]{shah2022flue}
Raj~Sanjay Shah, Kunal Chawla, Dheeraj Eidnani, Agam Shah, Wendi Du, Sudheer Chava, Natraj Raman, Charese Smiley, Jiaao Chen, and Diyi Yang.
\newblock When flue meets flang: Benchmarks and large pre-trained language model for financial domain.
\newblock \emph{arXiv preprint arXiv:2211.00083}, 2022.

\bibitem[Shaham et~al.(2023)Shaham, Ivgi, Efrat, Berant, and Levy]{shaham2023zeroscrolls}
Uri Shaham, Maor Ivgi, Avia Efrat, Jonathan Berant, and Omer Levy.
\newblock Zeroscrolls: A zero-shot benchmark for long text understanding.
\newblock \emph{arXiv preprint arXiv:2305.14196}, 2023.

\bibitem[shareAI(2023)]{ShareGPT-Chinese-English-90k}
shareAI.
\newblock Sharegpt-chinese-english-90k bilingual human-machine qa dataset.
\newblock \url{https://huggingface.co/datasets/shareAI/ShareGPT-Chinese-English-90k}, 2023.

\bibitem[Sharma et~al.(2022)Sharma, Nayak, Bose, Meena, Dasgupta, Ganguly, and Goyal]{sharma2022finred}
Soumya Sharma, Tapas Nayak, Arusarka Bose, Ajay~Kumar Meena, Koustuv Dasgupta, Niloy Ganguly, and Pawan Goyal.
\newblock Finred: A dataset for relation extraction in financial domain.
\newblock In \emph{Companion Proceedings of the Web Conference 2022}, pp.\  595--597, 2022.

\bibitem[Shen(2024)]{shen2024rethinking}
Ming Shen.
\newblock Rethinking data selection for supervised fine-tuning.
\newblock \emph{arXiv preprint arXiv:2402.06094}, 2024.

\bibitem[Shin et~al.(2023)Shin, Ahn, Kang, and Kang]{shin2023fedsplitx}
Jiyun Shin, Jinhyun Ahn, Honggu Kang, and Joonhyuk Kang.
\newblock Fedsplitx: Federated split learning for computationally-constrained heterogeneous clients.
\newblock \emph{arXiv preprint arXiv:2310.14579}, 2023.

\bibitem[Singhal et~al.(2023)Singhal, Azizi, Tu, Mahdavi, Wei, Chung, Scales, Tanwani, Cole-Lewis, Pfohl, et~al.]{singhal2023large}
Karan Singhal, Shekoofeh Azizi, Tao Tu, S~Sara Mahdavi, Jason Wei, Hyung~Won Chung, Nathan Scales, Ajay Tanwani, Heather Cole-Lewis, Stephen Pfohl, et~al.
\newblock Large language models encode clinical knowledge.
\newblock \emph{Nature}, 620\penalty0 (7972):\penalty0 172--180, 2023.

\bibitem[Srivastava et~al.(2022)Srivastava, Rastogi, Rao, Shoeb, Abid, Fisch, Brown, Santoro, Gupta, Garriga-Alonso, et~al.]{srivastava2022beyond}
Aarohi Srivastava, Abhinav Rastogi, Abhishek Rao, Abu Awal~Md Shoeb, Abubakar Abid, Adam Fisch, Adam~R Brown, Adam Santoro, Aditya Gupta, Adri{\`a} Garriga-Alonso, et~al.
\newblock Beyond the imitation game: Quantifying and extrapolating the capabilities of language models.
\newblock \emph{arXiv preprint arXiv:2206.04615}, 2022.

\bibitem[Su et~al.(2023)Su, Li, and Xue]{su2023fedra}
Shangchao Su, Bin Li, and Xiangyang Xue.
\newblock Fedra: A random allocation strategy for federated tuning to unleash the power of heterogeneous clients.
\newblock \emph{arXiv preprint arXiv:2311.11227}, 2023.

\bibitem[Su et~al.(2024)Su, Yang, Li, and Xue]{su2024federated2}
Shangchao Su, Mingzhao Yang, Bin Li, and Xiangyang Xue.
\newblock Federated adaptive prompt tuning for multi-domain collaborative learning.
\newblock In \emph{Proceedings of the AAAI Conference on Artificial Intelligence}, volume~38, pp.\  15117--15125, 2024.

\bibitem[Sun et~al.(2022)Sun, Khalid, Mendieta, Yang, Wang, Lee, and Chen]{sun2022conquering}
Guangyu Sun, Umar Khalid, Matias Mendieta, Taojiannan Yang, Pu~Wang, Minwoo Lee, and Chen Chen.
\newblock Conquering the communication constraints to enable large pre-trained models in federated learning.
\newblock \emph{arXiv preprint arXiv:2210.01708}, 2022.

\bibitem[Sun et~al.(2023)Sun, Xu, Yin, Yang, Xu, Chen, and Roth]{sun2023fedbpt}
Jingwei Sun, Ziyue Xu, Hongxu Yin, Dong Yang, Daguang Xu, Yiran Chen, and Holger~R Roth.
\newblock Fedbpt: Efficient federated black-box prompt tuning for large language models.
\newblock \emph{arXiv preprint arXiv:2310.01467}, 2023.

\bibitem[Sun et~al.(2024)Sun, Xie, Ding, Li, and Zhang]{sun2024exploring}
Yuchang Sun, Yuexiang Xie, Bolin Ding, Yaliang Li, and Jun Zhang.
\newblock Exploring selective layer fine-tuning in federated learning.
\newblock \emph{arXiv preprint arXiv:2408.15600}, 2024.

\bibitem[Tam et~al.(2023{\natexlab{a}})Tam, Li, Han, Xu, and Fu]{tam2023federated}
Kahou Tam, Li~Li, Bo~Han, Chengzhong Xu, and Huazhu Fu.
\newblock Federated noisy client learning.
\newblock \emph{IEEE Transactions on Neural Networks and Learning Systems}, 2023{\natexlab{a}}.

\bibitem[Tam et~al.(2023{\natexlab{b}})Tam, Li, Zhao, and Xu]{tam2023fedcoop}
Kahou Tam, Li~Li, Yan Zhao, and Chengzhong Xu.
\newblock Fedcoop: Cooperative federated learning for noisy labels.
\newblock In \emph{ECAI 2023}, pp.\  2298--2306. IOS Press, 2023{\natexlab{b}}.

\bibitem[Tam et~al.(2024{\natexlab{a}})Tam, Tian, Li, Zhao, and Xu]{tam2024fedhybrid}
Kahou Tam, Chunlin Tian, Li~Li, Haikai Zhao, and ChengZhong Xu.
\newblock Fedhybrid: Breaking the memory wall of federated learning via hybrid tensor management.
\newblock In \emph{Proceedings of the 22nd ACM Conference on Embedded Networked Sensor Systems}, pp.\  394--408, 2024{\natexlab{a}}.

\bibitem[Tam et~al.(2024{\natexlab{b}})Tam, Xu, Li, and Fu]{tam2024towards}
Kahou Tam, Kewei Xu, Li~Li, and Huazhu Fu.
\newblock Towards federated domain unlearning: Verification methodologies and challenges.
\newblock \emph{arXiv preprint arXiv:2406.03078}, 2024{\natexlab{b}}.

\bibitem[Tang et~al.(2025)Tang, Shao, Sohn, Chen, Zhang, Xiang, Wu, Zhao, Wu, Shi, et~al.]{tang2025medagentsbench}
Xiangru Tang, Daniel Shao, Jiwoong Sohn, Jiapeng Chen, Jiayi Zhang, Jinyu Xiang, Fang Wu, Yilun Zhao, Chenglin Wu, Wenqi Shi, et~al.
\newblock Medagentsbench: Benchmarking thinking models and agent frameworks for complex medical reasoning.
\newblock \emph{arXiv preprint arXiv:2503.07459}, 2025.

\bibitem[Tang et~al.(2024{\natexlab{a}})Tang, Zhou, Li, Ji, Hou, and Zhang]{tang2024citeeval}
Zecheng Tang, Keyan Zhou, Juntao Li, Baibei Ji, Jianye Hou, and Min Zhang.
\newblock L-citeeval: Do long-context models truly leverage context for responding?
\newblock \emph{arXiv preprint arXiv:2410.02115}, 2024{\natexlab{a}}.

\bibitem[Tang et~al.(2024{\natexlab{b}})Tang, Zhang, Wang, and Wei]{tang2024mathscale}
Zhengyang Tang, Xingxing Zhang, Benyou Wang, and Furu Wei.
\newblock Mathscale: Scaling instruction tuning for mathematical reasoning.
\newblock \emph{arXiv preprint arXiv:2403.02884}, 2024{\natexlab{b}}.

\bibitem[Taori et~al.(2023)Taori, Gulrajani, Zhang, Dubois, Li, Guestrin, Liang, and Hashimoto]{taori2023stanford}
Rohan Taori, Ishaan Gulrajani, Tianyi Zhang, Yann Dubois, Xuechen Li, Carlos Guestrin, Percy Liang, and Tatsunori~B Hashimoto.
\newblock Stanford alpaca: An instruction-following llama model, 2023.

\bibitem[Tavallaie \& Nazemi$^1$()Tavallaie and Nazemi$^1$]{tavallaie2024rbla}
Omid Tavallaie and Niousha Nazemi$^1$.
\newblock Rbla: Rank-based-lora-aggregation for fine-tuning heterogeneous models.
\newblock In \emph{Web Services--ICWS 2024: 31st International Conference, Held as Part of the Services Conference Federation, SCF 2024, Bangkok, Thailand, November 16--19, 2024, Proceedings}, pp.\ ~47. Springer Nature.

\bibitem[Team et~al.(2023)Team, Anil, Borgeaud, Alayrac, Yu, Soricut, Schalkwyk, Dai, Hauth, Millican, et~al.]{team2023gemini}
Gemini Team, Rohan Anil, Sebastian Borgeaud, Jean-Baptiste Alayrac, Jiahui Yu, Radu Soricut, Johan Schalkwyk, Andrew~M Dai, Anja Hauth, Katie Millican, et~al.
\newblock Gemini: a family of highly capable multimodal models.
\newblock \emph{arXiv preprint arXiv:2312.11805}, 2023.

\bibitem[Thapa et~al.(2022)Thapa, Arachchige, Camtepe, and Sun]{thapa2022splitfed}
Chandra Thapa, Pathum Chamikara~Mahawaga Arachchige, Seyit Camtepe, and Lichao Sun.
\newblock Splitfed: When federated learning meets split learning.
\newblock In \emph{Proceedings of the AAAI Conference on Artificial Intelligence}, volume~36, pp.\  8485--8493, 2022.

\bibitem[Thirunavukarasu et~al.(2023)Thirunavukarasu, Ting, Elangovan, Gutierrez, Tan, and Ting]{thirunavukarasu2023large}
Arun~James Thirunavukarasu, Darren Shu~Jeng Ting, Kabilan Elangovan, Laura Gutierrez, Ting~Fang Tan, and Daniel Shu~Wei Ting.
\newblock Large language models in medicine.
\newblock \emph{Nature medicine}, 29\penalty0 (8):\penalty0 1930--1940, 2023.

\bibitem[Tian et~al.(2022{\natexlab{a}})Tian, Li, Shi, Wang, and Xu]{tian2022harmony}
Chunlin Tian, Li~Li, Zhan Shi, Jun Wang, and ChengZhong Xu.
\newblock Harmony: Heterogeneity-aware hierarchical management for federated learning system.
\newblock In \emph{2022 55th IEEE/ACM International Symposium on Microarchitecture (MICRO)}, pp.\  631--645. IEEE, 2022{\natexlab{a}}.

\bibitem[Tian et~al.(2023)Tian, Shi, and Li]{DBLP:conf/iclr/TianSL23}
Chunlin Tian, Zhan Shi, and Li~Li.
\newblock Learn to select: Efficient cross-device federated learning via reinforcement learning.
\newblock In Krystal Maughan, Rosanne Liu, and Thomas~F. Burns (eds.), \emph{The First Tiny Papers Track at {ICLR} 2023, Tiny Papers @ {ICLR} 2023, Kigali, Rwanda, May 5, 2023}. OpenReview.net, 2023.
\newblock URL \url{https://openreview.net/forum?id=wecTsVkrjit}.

\bibitem[Tian et~al.(2024{\natexlab{a}})Tian, Li, Tam, Wu, and Xu]{tian2024breaking}
Chunlin Tian, Li~Li, Kahou Tam, Yebo Wu, and Cheng-Zhong Xu.
\newblock Breaking the memory wall for heterogeneous federated learning via model splitting.
\newblock \emph{IEEE Transactions on Parallel and Distributed Systems}, 2024{\natexlab{a}}.

\bibitem[Tian et~al.(2024{\natexlab{b}})Tian, Shi, Guo, Li, and Xu]{tian2024hydralora}
Chunlin Tian, Zhan Shi, Zhijiang Guo, Li~Li, and Chengzhong Xu.
\newblock Hydralora: An asymmetric lora architecture for efficient fine-tuning.
\newblock \emph{arXiv preprint arXiv:2404.19245}, 2024{\natexlab{b}}.

\bibitem[Tian et~al.(2024{\natexlab{c}})Tian, Shi, Li, Xu, et~al.]{tian2024ranking}
Chunlin Tian, Zhan Shi, Li~Li, Cheng-zhong Xu, et~al.
\newblock Ranking-based client imitation selection for efficient federated learning.
\newblock In \emph{Forty-first International Conference on Machine Learning}, 2024{\natexlab{c}}.

\bibitem[Tian et~al.(2025)Tian, Qin, Tam, Li, Wang, Zhao, Zhang, and Xu]{tian2025clone}
Chunlin Tian, Xinpeng Qin, Kahou Tam, Li~Li, Zijian Wang, Yuanzhe Zhao, Minglei Zhang, and Chengzhong Xu.
\newblock Clone: Customizing llms for efficient latency-aware inference at the edge.
\newblock \emph{arXiv preprint arXiv:2506.02847}, 2025.

\bibitem[Tian et~al.(2022{\natexlab{b}})Tian, Wan, Lyu, Yao, Jin, and Sun]{tian2022fedbert}
Yuanyishu Tian, Yao Wan, Lingjuan Lyu, Dezhong Yao, Hai Jin, and Lichao Sun.
\newblock Fedbert: When federated learning meets pre-training.
\newblock \emph{ACM Transactions on Intelligent Systems and Technology (TIST)}, 13\penalty0 (4):\penalty0 1--26, 2022{\natexlab{b}}.

\bibitem[Tiwari et~al.(2022)Tiwari, Killamsetty, Iyer, and Shenoy]{tiwari2022gcr}
Rishabh Tiwari, Krishnateja Killamsetty, Rishabh Iyer, and Pradeep Shenoy.
\newblock Gcr: Gradient coreset based replay buffer selection for continual learning.
\newblock In \emph{Proceedings of the IEEE/CVF Conference on Computer Vision and Pattern Recognition}, pp.\  99--108, 2022.

\bibitem[Toshniwal et~al.(2024)Toshniwal, Moshkov, Narenthiran, Gitman, Jia, and Gitman]{toshniwal2024openmath}
Shubham Toshniwal, Ivan Moshkov, Sean Narenthiran, Daria Gitman, Fei Jia, and Igor Gitman.
\newblock Openmathinstruct-1: A 1.8 million math instruction tuning dataset.
\newblock \emph{arXiv preprint arXiv: Arxiv-2402.10176}, 2024.

\bibitem[Touvron et~al.(2023{\natexlab{a}})Touvron, Lavril, Izacard, Martinet, Lachaux, Lacroix, Rozi{\`e}re, Goyal, Hambro, Azhar, et~al.]{llama}
Hugo Touvron, Thibaut Lavril, Gautier Izacard, Xavier Martinet, Marie-Anne Lachaux, Timoth{\'e}e Lacroix, Baptiste Rozi{\`e}re, Naman Goyal, Eric Hambro, Faisal Azhar, et~al.
\newblock Llama: Open and efficient foundation language models.
\newblock \emph{arXiv preprint arXiv:2302.13971}, 2023{\natexlab{a}}.

\bibitem[Touvron et~al.(2023{\natexlab{b}})Touvron, Martin, Stone, Albert, Almahairi, Babaei, Bashlykov, Batra, Bhargava, Bhosale, et~al.]{llama2}
Hugo Touvron, Louis Martin, Kevin Stone, Peter Albert, Amjad Almahairi, Yasmine Babaei, Nikolay Bashlykov, Soumya Batra, Prajjwal Bhargava, Shruti Bhosale, et~al.
\newblock Llama 2: Open foundation and fine-tuned chat models.
\newblock \emph{arXiv preprint arXiv:2307.09288}, 2023{\natexlab{b}}.

\bibitem[Trinh et~al.(2024)Trinh, Wu, Le, He, and Luong]{trinh2024solving}
Trieu~H Trinh, Yuhuai Wu, Quoc~V Le, He~He, and Thang Luong.
\newblock Solving olympiad geometry without human demonstrations.
\newblock \emph{Nature}, 625\penalty0 (7995):\penalty0 476--482, 2024.

\bibitem[Voigt \& Von~dem Bussche(2017)Voigt and Von~dem Bussche]{voigt2017eu}
Paul Voigt and Axel Von~dem Bussche.
\newblock The eu general data protection regulation (gdpr).
\newblock \emph{A Practical Guide, 1st Ed., Cham: Springer International Publishing}, 10\penalty0 (3152676):\penalty0 10--5555, 2017.

\bibitem[Wang et~al.(2023{\natexlab{a}})Wang, Cheng, Zhan, Li, Song, and Liu]{wang2023openchat}
Guan Wang, Sijie Cheng, Xianyuan Zhan, Xiangang Li, Sen Song, and Yang Liu.
\newblock Openchat: Advancing open-source language models with mixed-quality data.
\newblock \emph{arXiv preprint arXiv:2309.11235}, 2023{\natexlab{a}}.

\bibitem[Wang et~al.(2022{\natexlab{a}})Wang, Wu, He, Huang, and Church]{wang2022progress}
Haifeng Wang, Hua Wu, Zhongjun He, Liang Huang, and Kenneth~Ward Church.
\newblock Progress in machine translation.
\newblock \emph{Engineering}, 18:\penalty0 143--153, 2022{\natexlab{a}}.

\bibitem[Wang et~al.(2020{\natexlab{a}})Wang, Yurochkin, Sun, Papailiopoulos, and Khazaeni]{wang2020federated}
Hongyi Wang, Mikhail Yurochkin, Yuekai Sun, Dimitris Papailiopoulos, and Yasaman Khazaeni.
\newblock Federated learning with matched averaging.
\newblock \emph{arXiv preprint arXiv:2002.06440}, 2020{\natexlab{a}}.

\bibitem[Wang et~al.(2020{\natexlab{b}})Wang, Liu, Liang, Joshi, and Poor]{wang2020tackling}
Jianyu Wang, Qinghua Liu, Hao Liang, Gauri Joshi, and H~Vincent Poor.
\newblock Tackling the objective inconsistency problem in heterogeneous federated optimization.
\newblock \emph{Advances in neural information processing systems}, 33:\penalty0 7611--7623, 2020{\natexlab{b}}.

\bibitem[Wang et~al.(2023{\natexlab{b}})Wang, Wu, Liu, Wu, Qu, Geng, and Zhang]{wang2023fedins2}
Jie Wang, Yebo Wu, Erwu Liu, Xiaolong Wu, Xinyu Qu, Yuanzhe Geng, and Hanfu Zhang.
\newblock Fedins2: A federated-edge-learning-based inertial navigation system with segment fusion.
\newblock \emph{IEEE Internet of Things Journal}, 11\penalty0 (2):\penalty0 3653--3661, 2023{\natexlab{b}}.

\bibitem[Wang et~al.(2025)Wang, Wu, Tian, Liu, Wu, Lai, and Tian]{wang2025indoor}
Jie Wang, Xiaolong Wu, Jindong Tian, Erwu Liu, Yebo Wu, Rucong Lai, and Yong Tian.
\newblock Indoor localization fusing inertial navigation with monocular depth estimation in federated learning framework with data heterogeneity.
\newblock \emph{IEEE Transactions on Instrumentation and Measurement}, 2025.

\bibitem[Wang et~al.(2024{\natexlab{a}})Wang, Dong, Xu, Dong, Wang, Saha, Lim, Xiong, and Sahoo]{wang2024mathhay}
Lei Wang, Shan Dong, Yuhui Xu, Hanze Dong, Yalu Wang, Amrita Saha, Ee-Peng Lim, Caiming Xiong, and Doyen Sahoo.
\newblock Mathhay: An automated benchmark for long-context mathematical reasoning in llms.
\newblock \emph{arXiv preprint arXiv:2410.04698}, 2024{\natexlab{a}}.

\bibitem[Wang et~al.(2023{\natexlab{c}})Wang, Xu, Xu, Jiang, Chen, Zhang, and Qian]{wang2023bose}
Lun Wang, Yang Xu, Hongli Xu, Zhida Jiang, Min Chen, Wuyang Zhang, and Chen Qian.
\newblock Bose: Block-wise federated learning in heterogeneous edge computing.
\newblock \emph{IEEE/ACM Transactions on Networking}, 2023{\natexlab{c}}.

\bibitem[Wang et~al.(2024{\natexlab{b}})Wang, Chen, Fu, Liao, Zhang, Wu, Yu, Xu, Zhang, Luo, et~al.]{wang2024leave}
Minzheng Wang, Longze Chen, Cheng Fu, Shengyi Liao, Xinghua Zhang, Bingli Wu, Haiyang Yu, Nan Xu, Lei Zhang, Run Luo, et~al.
\newblock Leave no document behind: Benchmarking long-context llms with extended multi-doc qa.
\newblock \emph{arXiv preprint arXiv:2406.17419}, 2024{\natexlab{b}}.

\bibitem[Wang et~al.(2024{\natexlab{c}})Wang, Yu, Zhang, Kim, Rossi, Zhao, Wu, Mitra, Yao, and Henao]{wang2024personalized}
Rui Wang, Tong Yu, Ruiyi Zhang, Sungchul Kim, Ryan Rossi, Handong Zhao, Junda Wu, Subrata Mitra, Lina Yao, and Ricardo Henao.
\newblock Personalized federated learning for text classification with gradient-free prompt tuning.
\newblock In \emph{Findings of the Association for Computational Linguistics: NAACL 2024}, pp.\  4597--4612, 2024{\natexlab{c}}.

\bibitem[Wang et~al.(2019{\natexlab{a}})Wang, Tuor, Salonidis, Leung, Makaya, He, and Chan]{wang2019adaptive}
Shiqiang Wang, Tiffany Tuor, Theodoros Salonidis, Kin~K Leung, Christian Makaya, Ting He, and Kevin Chan.
\newblock Adaptive federated learning in resource constrained edge computing systems.
\newblock \emph{IEEE journal on selected areas in communications}, 37\penalty0 (6):\penalty0 1205--1221, 2019{\natexlab{a}}.

\bibitem[Wang et~al.(2024{\natexlab{d}})Wang, Wang, Xiao, Chen, and Ma]{wang2024fedkim}
Xiaochen Wang, Jiaqi Wang, Houping Xiao, Jinghui Chen, and Fenglong Ma.
\newblock Fedkim: Adaptive federated knowledge injection into medical foundation models.
\newblock \emph{arXiv preprint arXiv:2408.10276}, 2024{\natexlab{d}}.

\bibitem[Wang et~al.(2023{\natexlab{d}})Wang, Hu, Lu, Zhu, Zhang, Subramaniam, Loomba, Zhang, Sun, and Wang]{wang2023scibench}
Xiaoxuan Wang, Ziniu Hu, Pan Lu, Yanqiao Zhu, Jieyu Zhang, Satyen Subramaniam, Arjun~R Loomba, Shichang Zhang, Yizhou Sun, and Wei Wang.
\newblock Scibench: Evaluating college-level scientific problem-solving abilities of large language models.
\newblock \emph{arXiv preprint arXiv:2307.10635}, 2023{\natexlab{d}}.

\bibitem[Wang et~al.(2023{\natexlab{e}})Wang, Chen, Song, Zhang, Chen, Xiao, Jiang, Li, Wan, Wang, et~al.]{wang2023cmb}
Xidong Wang, Guiming~Hardy Chen, Dingjie Song, Zhiyi Zhang, Zhihong Chen, Qingying Xiao, Feng Jiang, Jianquan Li, Xiang Wan, Benyou Wang, et~al.
\newblock Cmb: A comprehensive medical benchmark in chinese.
\newblock \emph{arXiv preprint arXiv:2308.08833}, 2023{\natexlab{e}}.

\bibitem[Wang et~al.(2023{\natexlab{f}})Wang, Yu, Zeng, Yang, Wang, Chen, Jiang, Xie, Wang, Xie, et~al.]{wang2023pandalm}
Yidong Wang, Zhuohao Yu, Zhengran Zeng, Linyi Yang, Cunxiang Wang, Hao Chen, Chaoya Jiang, Rui Xie, Jindong Wang, Xing Xie, et~al.
\newblock Pandalm: An automatic evaluation benchmark for llm instruction tuning optimization.
\newblock \emph{arXiv preprint arXiv:2306.05087}, 2023{\natexlab{f}}.

\bibitem[Wang et~al.(2022{\natexlab{b}})Wang, Kordi, Mishra, Liu, Smith, Khashabi, and Hajishirzi]{wang2022self}
Yizhong Wang, Yeganeh Kordi, Swaroop Mishra, Alisa Liu, Noah~A Smith, Daniel Khashabi, and Hannaneh Hajishirzi.
\newblock Self-instruct: Aligning language models with self-generated instructions.
\newblock \emph{arXiv preprint arXiv:2212.10560}, 2022{\natexlab{b}}.

\bibitem[Wang et~al.(2022{\natexlab{c}})Wang, Zhou, Fried, and Neubig]{wang2022execution}
Zhiruo Wang, Shuyan Zhou, Daniel Fried, and Graham Neubig.
\newblock Execution-based evaluation for open-domain code generation.
\newblock \emph{arXiv preprint arXiv:2212.10481}, 2022{\natexlab{c}}.

\bibitem[Wang et~al.(2024{\natexlab{e}})Wang, Wu, Yu, Zhou, Hu, and Min]{wang2024federated}
Zi~Wang, Fei Wu, Feng Yu, Yurui Zhou, Jia Hu, and Geyong Min.
\newblock Federated continual learning for edge-ai: A comprehensive survey.
\newblock \emph{arXiv preprint arXiv:2411.13740}, 2024{\natexlab{e}}.

\bibitem[Wang et~al.(2019{\natexlab{b}})Wang, Wohlwend, and Lei]{wang2019structured}
Ziheng Wang, Jeremy Wohlwend, and Tao Lei.
\newblock Structured pruning of large language models.
\newblock \emph{arXiv preprint arXiv:1910.04732}, 2019{\natexlab{b}}.

\bibitem[Wang et~al.(2024{\natexlab{f}})Wang, Shen, He, Sun, Wang, Lyu, and Li]{wang2024flora}
Ziyao Wang, Zheyu Shen, Yexiao He, Guoheng Sun, Hongyi Wang, Lingjuan Lyu, and Ang Li.
\newblock Flora: Federated fine-tuning large language models with heterogeneous low-rank adaptations.
\newblock \emph{arXiv preprint arXiv:2409.05976}, 2024{\natexlab{f}}.

\bibitem[Wankhade et~al.(2022)Wankhade, Rao, and Kulkarni]{wankhade2022survey}
Mayur Wankhade, Annavarapu Chandra~Sekhara Rao, and Chaitanya Kulkarni.
\newblock A survey on sentiment analysis methods, applications, and challenges.
\newblock \emph{Artificial Intelligence Review}, 55\penalty0 (7):\penalty0 5731--5780, 2022.

\bibitem[Watson et~al.(2024)Watson, Meyer, Van~Segbroeck, Grossman, Torbey, Mlocek, and Greco]{gretel-synthetic-pii-finance-multilingual-2024}
Alex Watson, Yev Meyer, Maarten Van~Segbroeck, Matthew Grossman, Sami Torbey, Piotr Mlocek, and Johnny Greco.
\newblock {Synthetic-PII-Financial-Documents-North-America}: A synthetic dataset for training language models to label and detect pii in domain specific formats, June 2024.
\newblock URL \url{https://huggingface.co/datasets/gretelai/synthetic_pii_finance_multilingual}.

\bibitem[Wei et~al.(2023)Wei, Wang, Shah, and Chellappa]{wei2023dual}
Guoyizhe Wei, Feng Wang, Anshul Shah, and Rama Chellappa.
\newblock Dual prompt tuning for domain-aware federated learning.
\newblock \emph{arXiv preprint arXiv:2310.03103}, 2023.

\bibitem[Weng et~al.()Weng, Hoang, Nguyen, Thai, Weng, and Hoang]{wengprobabilistic}
Pei-Yau Weng, Minh Hoang, Lam~M Nguyen, My~T Thai, Tsui-Wei Weng, and Trong~Nghia Hoang.
\newblock Probabilistic federated prompt-tuning with non-iid and imbalanced data.
\newblock In \emph{The Thirty-eighth Annual Conference on Neural Information Processing Systems}.

\bibitem[Woisetschl{\"a}ger et~al.(2024)Woisetschl{\"a}ger, Isenko, Wang, Mayer, and Jacobsen]{woisetschlager2024survey}
Herbert Woisetschl{\"a}ger, Alexander Isenko, Shiqiang Wang, Ruben Mayer, and Hans-Arno Jacobsen.
\newblock A survey on efficient federated learning methods for foundation model training.
\newblock \emph{arXiv preprint arXiv:2401.04472}, 2024.

\bibitem[Wu et~al.(2024{\natexlab{a}})Wu, Wang, Yu, Zhang, Chang, and Yu]{wu2024longmemeval}
Di~Wu, Hongwei Wang, Wenhao Yu, Yuwei Zhang, Kai-Wei Chang, and Dong Yu.
\newblock Longmemeval: Benchmarking chat assistants on long-term interactive memory.
\newblock \emph{arXiv preprint arXiv:2410.10813}, 2024{\natexlab{a}}.

\bibitem[Wu et~al.(2024{\natexlab{b}})Wu, Li, Li, Ding, and Gao]{wu2024fedbiot}
Feijie Wu, Zitao Li, Yaliang Li, Bolin Ding, and Jing Gao.
\newblock Fedbiot: Llm local fine-tuning in federated learning without full model.
\newblock In \emph{Proceedings of the 30th ACM SIGKDD Conference on Knowledge Discovery and Data Mining}, pp.\  3345--3355, 2024{\natexlab{b}}.

\bibitem[Wu et~al.(2024{\natexlab{c}})Wu, Liu, Wang, Wang, and Gao]{wu2024client}
Feijie Wu, Xiaoze Liu, Haoyu Wang, Xingchen Wang, and Jing Gao.
\newblock On the client preference of llm fine-tuning in federated learning.
\newblock \emph{arXiv preprint arXiv:2407.03038}, 2024{\natexlab{c}}.

\bibitem[Wu et~al.(2024{\natexlab{d}})Wu, Zhao, Zhang, Wu, Zhu, Zhang, Ouyang, Zhang, Wang, Yang, et~al.]{wu2024medjourney}
Xian Wu, Yutian Zhao, Yunyan Zhang, Jiageng Wu, Zhihong Zhu, Yingying Zhang, Yi~Ouyang, Ziheng Zhang, Huimin Wang, Jie Yang, et~al.
\newblock Medjourney: Benchmark and evaluation of large language models over patient clinical journey.
\newblock \emph{Advances in Neural Information Processing Systems}, 37:\penalty0 87621--87646, 2024{\natexlab{d}}.

\bibitem[Wu et~al.(2024{\natexlab{e}})Wu, Wang, Liu, Shi, Yan, Lu, Zhu, and Zhang]{wu2024lifbench}
Xiaodong Wu, Minhao Wang, Yichen Liu, Xiaoming Shi, He~Yan, Xiangju Lu, Junmin Zhu, and Wei Zhang.
\newblock Lifbench: Evaluating the instruction following performance and stability of large language models in long-context scenarios.
\newblock \emph{arXiv preprint arXiv:2411.07037}, 2024{\natexlab{e}}.

\bibitem[Wu et~al.(2024{\natexlab{f}})Wu, Liu, Niu, Wang, Tang, and Zhu]{wu2024fedlora}
Xinghao Wu, Xuefeng Liu, Jianwei Niu, Haolin Wang, Shaojie Tang, and Guogang Zhu.
\newblock Fedlora: When personalized federated learning meets low-rank adaptation.
\newblock 2024{\natexlab{f}}.

\bibitem[Wu et~al.(2024{\natexlab{g}})Wu, Niu, Liu, Shi, Zhu, and Tang]{wu2024tackling}
Xinghao Wu, Jianwei Niu, Xuefeng Liu, Mingjia Shi, Guogang Zhu, and Shaojie Tang.
\newblock Tackling feature-classifier mismatch in federated learning via prompt-driven feature transformation.
\newblock \emph{arXiv preprint arXiv:2407.16139}, 2024{\natexlab{g}}.

\bibitem[Wu et~al.(2024{\natexlab{h}})Wu, Li, Tian, Chang, Lin, Wang, and Xu]{wu2024heterogeneity}
Yebo Wu, Li~Li, Chunlin Tian, Tao Chang, Chi Lin, Cong Wang, and Cheng-Zhong Xu.
\newblock Heterogeneity-aware memory efficient federated learning via progressive layer freezing.
\newblock In \emph{2024 IEEE/ACM 32nd International Symposium on Quality of Service (IWQoS)}, pp.\  1--10. IEEE, 2024{\natexlab{h}}.

\bibitem[Wu et~al.(2024{\natexlab{i}})Wu, Li, Tian, Chen, and Xu]{wu2024neulite}
Yebo Wu, Li~Li, Chunlin Tian, Dubing Chen, and Chengzhong Xu.
\newblock Neulite: Memory-efficient federated learning via elastic progressive training.
\newblock \emph{arXiv preprint arXiv:2408.10826}, 2024{\natexlab{i}}.

\bibitem[Wu et~al.(2025{\natexlab{a}})Wu, Li, Guo, and Li]{wu2025elastic}
Yebo Wu, Jingguang Li, Zhijiang Guo, and Li~Li.
\newblock Elastic mixture of rank-wise experts for knowledge reuse in federated fine-tuning.
\newblock \emph{arXiv preprint arXiv:2512.00902}, 2025{\natexlab{a}}.

\bibitem[Wu et~al.(2025{\natexlab{b}})Wu, Li, Guo, and Li]{wu2025learning}
Yebo Wu, Jingguang Li, Zhijiang Guo, and Li~Li.
\newblock Learning like humans: Resource-efficient federated fine-tuning through cognitive developmental stages.
\newblock \emph{arXiv preprint arXiv:2508.00041}, 2025{\natexlab{b}}.

\bibitem[Wu et~al.(2025{\natexlab{c}})Wu, Li, Tian, Guo, and Li]{wu2025memory}
Yebo Wu, Jingguang Li, Chunlin Tian, Zhijiang Guo, and Li~Li.
\newblock Memory-efficient federated fine-tuning of large language models via layer pruning.
\newblock \emph{arXiv preprint arXiv:2508.17209}, 2025{\natexlab{c}}.

\bibitem[Wu et~al.(2025{\natexlab{d}})Wu, Li, and Xu]{wu2025breaking}
Yebo Wu, Li~Li, and Cheng-zhong Xu.
\newblock Breaking the memory wall for heterogeneous federated learning via progressive training.
\newblock In \emph{Proceedings of the 31st ACM SIGKDD Conference on Knowledge Discovery and Data Mining V. 1}, pp.\  1623--1632, 2025{\natexlab{d}}.

\bibitem[Xie et~al.(2023)Xie, Han, Zhang, Lai, Peng, Lopez-Lira, and Huang]{xie2023pixiu}
Qianqian Xie, Weiguang Han, Xiao Zhang, Yanzhao Lai, Min Peng, Alejandro Lopez-Lira, and Jimin Huang.
\newblock Pixiu: A large language model, instruction data and evaluation benchmark for finance.
\newblock \emph{arXiv preprint arXiv:2306.05443}, 2023.

\bibitem[Xie et~al.(2024)Xie, Han, Chen, Xiang, Zhang, He, Xiao, Li, Dai, Feng, et~al.]{xie2024finben}
Qianqian Xie, Weiguang Han, Zhengyu Chen, Ruoyu Xiang, Xiao Zhang, Yueru He, Mengxi Xiao, Dong Li, Yongfu Dai, Duanyu Feng, et~al.
\newblock Finben: A holistic financial benchmark for large language models.
\newblock \emph{Advances in Neural Information Processing Systems}, 37:\penalty0 95716--95743, 2024.

\bibitem[Xin et~al.(2024)Xin, Luo, Zhou, Du, Liu, Fan, Li, and Du]{xin2024parameter}
Yi~Xin, Siqi Luo, Haodi Zhou, Junlong Du, Xiaohong Liu, Yue Fan, Qing Li, and Yuntao Du.
\newblock Parameter-efficient fine-tuning for pre-trained vision models: A survey.
\newblock \emph{arXiv preprint arXiv:2402.02242}, 2024.

\bibitem[Xiong et~al.(2023)Xiong, Li, Zheng, Guo, Yin, Xie, Yang, Cao, Wang, Han, et~al.]{xiong2023dq}
Jing Xiong, Zixuan Li, Chuanyang Zheng, Zhijiang Guo, Yichun Yin, Enze Xie, Zhicheng Yang, Qingxing Cao, Haiming Wang, Xiongwei Han, et~al.
\newblock Dq-lore: Dual queries with low rank approximation re-ranking for in-context learning.
\newblock \emph{arXiv preprint arXiv:2310.02954}, 2023.

\bibitem[Xiong et~al.(2024)Xiong, Shen, Ye, Tao, Wan, Lu, Wu, Zheng, Guo, Kong, et~al.]{xiong2024uncomp}
Jing Xiong, Jianghan Shen, Fanghua Ye, Chaofan Tao, Zhongwei Wan, Jianqiao Lu, Xun Wu, Chuanyang Zheng, Zhijiang Guo, Lingpeng Kong, et~al.
\newblock Uncomp: Uncertainty-aware long-context compressor for efficient large language model inference.
\newblock \emph{arXiv preprint arXiv:2410.03090}, 2024.

\bibitem[Xiong et~al.(2025)Xiong, Shen, Zheng, Wan, Zhao, Yang, Ye, Yang, Kong, and Wong]{xiong2025parallelcomp}
Jing Xiong, Jianghan Shen, Chuanyang Zheng, Zhongwei Wan, Chenyang Zhao, Chiwun Yang, Fanghua Ye, Hongxia Yang, Lingpeng Kong, and Ngai Wong.
\newblock Parallelcomp: Parallel long-context compressor for length extrapolation.
\newblock \emph{arXiv preprint arXiv:2502.14317}, 2025.

\bibitem[Xu et~al.(2023{\natexlab{a}})Xu, Sun, Zheng, Geng, Zhao, Feng, Tao, and Jiang]{xu2023wizardlm}
Can Xu, Qingfeng Sun, Kai Zheng, Xiubo Geng, Pu~Zhao, Jiazhan Feng, Chongyang Tao, and Daxin Jiang.
\newblock Wizardlm: Empowering large language models to follow complex instructions.
\newblock \emph{arXiv preprint arXiv:2304.12244}, 2023{\natexlab{a}}.

\bibitem[Xu et~al.(2023{\natexlab{b}})Xu, Guo, Duan, and McAuley]{xu2023baize}
Canwen Xu, Daya Guo, Nan Duan, and Julian McAuley.
\newblock Baize: An open-source chat model with parameter-efficient tuning on self-chat data.
\newblock \emph{arXiv preprint arXiv:2304.01196}, 2023{\natexlab{b}}.

\bibitem[Xu et~al.(2023{\natexlab{c}})Xu, Lin, Han, Zhao, Liu, and Cambria]{xu2023large}
Fangzhi Xu, Qika Lin, Jiawei Han, Tianzhe Zhao, Jun Liu, and Erik Cambria.
\newblock Are large language models really good logical reasoners? a comprehensive evaluation from deductive, inductive and abductive views.
\newblock \emph{arXiv preprint arXiv:2306.09841}, 2023{\natexlab{c}}.

\bibitem[Xu et~al.(2023{\natexlab{d}})Xu, Li, Zhu, Xue, Zhu, Zhao, He, Zhang, Kang, and Lan]{xu2023superclue}
Liang Xu, Anqi Li, Lei Zhu, Hang Xue, Changtai Zhu, Kangkang Zhao, Haonan He, Xuanwei Zhang, Qiyue Kang, and Zhenzhong Lan.
\newblock Superclue: A comprehensive chinese large language model benchmark.
\newblock \emph{arXiv preprint arXiv:2307.15020}, 2023{\natexlab{d}}.

\bibitem[Xu et~al.(2024{\natexlab{a}})Xu, Zhu, Wu, and Xue]{xu2024superclue}
Liang Xu, Lei Zhu, Yaotong Wu, and Hang Xue.
\newblock Superclue-fin: Graded fine-grained analysis of chinese llms on diverse financial tasks and applications.
\newblock \emph{arXiv preprint arXiv:2404.19063}, 2024{\natexlab{a}}.

\bibitem[Xu et~al.(2023{\natexlab{e}})Xu, Wu, Cai, Li, and Wang]{xu2023federated}
Mengwei Xu, Yaozong Wu, Dongqi Cai, Xiang Li, and Shangguang Wang.
\newblock Federated fine-tuning of billion-sized language models across mobile devices.
\newblock \emph{arXiv preprint arXiv:2308.13894}, 2023{\natexlab{e}}.

\bibitem[Xu et~al.(2024{\natexlab{b}})Xu, Cai, Wu, Li, and Wang]{xu2024fwdllm}
Mengwei Xu, Dongqi Cai, Yaozong Wu, Xiang Li, and Shangguang Wang.
\newblock $\{$FwdLLM$\}$: Efficient federated finetuning of large language models with perturbed inferences.
\newblock In \emph{2024 USENIX Annual Technical Conference (USENIX ATC 24)}, pp.\  579--596, 2024{\natexlab{b}}.

\bibitem[Xu et~al.(2024{\natexlab{c}})Xu, Yin, Cai, Yi, Xu, Wang, Wu, Zhao, Yang, Wang, et~al.]{xu2024survey}
Mengwei Xu, Wangsong Yin, Dongqi Cai, Rongjie Yi, Daliang Xu, Qipeng Wang, Bingyang Wu, Yihao Zhao, Chen Yang, Shihe Wang, et~al.
\newblock A survey of resource-efficient llm and multimodal foundation models.
\newblock \emph{arXiv preprint arXiv:2401.08092}, 2024{\natexlab{c}}.

\bibitem[Xu et~al.(2024{\natexlab{d}})Xu, Li, Du, Li, Luo, Liang, Li, Zhang, Han, Yin, et~al.]{xu2024copyrightmeter}
Naen Xu, Changjiang Li, Tianyu Du, Minxi Li, Wenjie Luo, Jiacheng Liang, Yuyuan Li, Xuhong Zhang, Meng Han, Jianwei Yin, et~al.
\newblock Copyrightmeter: Revisiting copyright protection in text-to-image models.
\newblock \emph{arXiv preprint arXiv:2411.13144}, 2024{\natexlab{d}}.

\bibitem[Xu et~al.(2025)Xu, Zhang, Li, Chen, Zhou, Li, Du, and Ji]{xu2025videoeraser}
Naen Xu, Jinghuai Zhang, Changjiang Li, Zhi Chen, Chunyi Zhou, Qingming Li, Tianyu Du, and Shouling Ji.
\newblock Videoeraser: Concept erasure in text-to-video diffusion models.
\newblock In \emph{Proceedings of the 2025 Conference on Empirical Methods in Natural Language Processing}, pp.\  5965--5994, 2025.

\bibitem[Xu et~al.(2026{\natexlab{a}})Xu, An, Shi, Zhang, Zhou, Li, Du, Fu, Wang, and Ji]{xu2026agents}
Naen Xu, Hengyu An, Shuo Shi, Jinghuai Zhang, Chunyi Zhou, Changjiang Li, Tianyu Du, Zhihui Fu, Jun Wang, and Shouling Ji.
\newblock When agents" misremember" collectively: Exploring the mandela effect in llm-based multi-agent systems.
\newblock \emph{arXiv preprint arXiv:2602.00428}, 2026{\natexlab{a}}.

\bibitem[Xu et~al.(2026{\natexlab{b}})Xu, Zhang, He, Zhou, Wang, Fu, Du, Wang, and Ji]{xu2026fraudshield}
Naen Xu, Jinghuai Zhang, Ping He, Chunyi Zhou, Jun Wang, Zhihui Fu, Tianyu Du, Zhaoxiang Wang, and Shouling Ji.
\newblock Fraudshield: Knowledge graph empowered defense for llms against fraud attacks.
\newblock \emph{arXiv preprint arXiv:2601.22485}, 2026{\natexlab{b}}.

\bibitem[Xu et~al.(2024{\natexlab{e}})Xu, Ye, Liu, Liu, Sun, Liu, Guo, Li, Liu, Huang, et~al.]{xu2024detectiveqa}
Zhe Xu, Jiasheng Ye, Xiaoran Liu, Xiangyang Liu, Tianxiang Sun, Zhigeng Liu, Qipeng Guo, Linlin Li, Qun Liu, Xuanjing Huang, et~al.
\newblock Detectiveqa: Evaluating long-context reasoning on detective novels.
\newblock \emph{arXiv preprint arXiv:2409.02465}, 2024{\natexlab{e}}.

\bibitem[Yan et~al.(2024)Yan, Yang, Tang, and Shi]{yan2024federa}
Yuxuan Yan, Qianqian Yang, Shunpu Tang, and Zhiguo Shi.
\newblock Federa: Efficient fine-tuning of language models in federated learning leveraging weight decomposition.
\newblock \emph{arXiv preprint arXiv:2404.18848}, 2024.

\bibitem[Yang et~al.(2025)Yang, Li, Yang, Zhang, Hui, Zheng, Yu, Gao, Huang, Lv, et~al.]{yang2025qwen3}
An~Yang, Anfeng Li, Baosong Yang, Beichen Zhang, Binyuan Hui, Bo~Zheng, Bowen Yu, Chang Gao, Chengen Huang, Chenxu Lv, et~al.
\newblock Qwen3 technical report.
\newblock \emph{arXiv preprint arXiv:2505.09388}, 2025.

\bibitem[Yang et~al.(2023{\natexlab{a}})Yang, Wang, and Wang]{yang2023efficient}
Fu-En Yang, Chien-Yi Wang, and Yu-Chiang~Frank Wang.
\newblock Efficient model personalization in federated learning via client-specific prompt generation.
\newblock In \emph{Proceedings of the IEEE/CVF International Conference on Computer Vision}, pp.\  19159--19168, 2023{\natexlab{a}}.

\bibitem[Yang(2023)]{Firefly}
Jianxin Yang.
\newblock Firefly: Chinese conversational large language models.
\newblock \url{https://github.com/yangjianxin1/Firefly}, 2023.

\bibitem[Yang et~al.(2022)Yang, Zhang, Qin, Li, Wang, Liu, Wang, Xie, and Zhang]{yang2022glue}
Linyi Yang, Shuibai Zhang, Libo Qin, Yafu Li, Yidong Wang, Hanmeng Liu, Jindong Wang, Xing Xie, and Yue Zhang.
\newblock Glue-x: Evaluating natural language understanding models from an out-of-distribution generalization perspective.
\newblock \emph{arXiv preprint arXiv:2211.08073}, 2022.

\bibitem[Yang et~al.(2024{\natexlab{a}})Yang, Zhao, Zhu, Zhou, Xu, Jia, and Zan]{yang2024zhongjing}
Songhua Yang, Hanjie Zhao, Senbin Zhu, Guangyu Zhou, Hongfei Xu, Yuxiang Jia, and Hongying Zan.
\newblock Zhongjing: Enhancing the chinese medical capabilities of large language model through expert feedback and real-world multi-turn dialogue.
\newblock In \emph{Proceedings of the AAAI conference on artificial intelligence}, volume~38, pp.\  19368--19376, 2024{\natexlab{a}}.

\bibitem[Yang et~al.(2024{\natexlab{b}})Yang, Long, Shen, Jiang, and Blumenstein]{yang2024dual}
Yiyuan Yang, Guodong Long, Tao Shen, Jing Jiang, and Michael Blumenstein.
\newblock Dual-personalizing adapter for federated foundation models.
\newblock \emph{arXiv preprint arXiv:2403.19211}, 2024{\natexlab{b}}.

\bibitem[Yang et~al.(2024{\natexlab{c}})Yang, Liu, Gao, Xu, and Wang]{yang2024sa}
Yuning Yang, Xiaohong Liu, Tianrun Gao, Xiaodong Xu, and Guangyu Wang.
\newblock Sa-fedlora: Adaptive parameter allocation for efficient federated learning with lora tuning.
\newblock \emph{arXiv preprint arXiv:2405.09394}, 2024{\natexlab{c}}.

\bibitem[Yang et~al.(2023{\natexlab{b}})Yang, Li, Lin, Wang, Lin, Liu, and Wang]{yang2023dawn}
Zhengyuan Yang, Linjie Li, Kevin Lin, Jianfeng Wang, Chung-Ching Lin, Zicheng Liu, and Lijuan Wang.
\newblock The dawn of lmms: Preliminary explorations with gpt-4v (ision).
\newblock \emph{arXiv preprint arXiv:2309.17421}, 9\penalty0 (1):\penalty0 1, 2023{\natexlab{b}}.

\bibitem[Yang et~al.(2019)Yang, Dai, Yang, Carbonell, Salakhutdinov, and Le]{yang2019xlnet}
Zhilin Yang, Zihang Dai, Yiming Yang, Jaime Carbonell, Russ~R Salakhutdinov, and Quoc~V Le.
\newblock Xlnet: Generalized autoregressive pretraining for language understanding.
\newblock \emph{Advances in neural information processing systems}, 32, 2019.

\bibitem[Yao et~al.(2021)Yao, Pan, O'Neill, Dai, Wan, Jin, and Sun]{yao2021fedhm}
Dezhong Yao, Wanning Pan, Michael~J O'Neill, Yutong Dai, Yao Wan, Hai Jin, and Lichao Sun.
\newblock Fedhm: Efficient federated learning for heterogeneous models via low-rank factorization.
\newblock \emph{arXiv preprint arXiv:2111.14655}, 2021.

\bibitem[Yao et~al.(2024)Yao, Zhang, Wu, Huang, Xia, Yu, Zhang, Kim, Rossi, Li, et~al.]{yao2024federated}
Yuhang Yao, Jianyi Zhang, Junda Wu, Chengkai Huang, Yu~Xia, Tong Yu, Ruiyi Zhang, Sungchul Kim, Ryan Rossi, Ang Li, et~al.
\newblock Federated large language models: Current progress and future directions.
\newblock \emph{arXiv preprint arXiv:2409.15723}, 2024.

\bibitem[Ye et~al.(2023)Ye, Fang, Du, Yuen, and Tao]{ye2023heterogeneous}
Mang Ye, Xiuwen Fang, Bo~Du, Pong~C Yuen, and Dacheng Tao.
\newblock Heterogeneous federated learning: State-of-the-art and research challenges.
\newblock \emph{ACM Computing Surveys}, 56\penalty0 (3):\penalty0 1--44, 2023.

\bibitem[Ye et~al.(2024{\natexlab{a}})Ye, Kou, and Tang]{ye2024praffl}
Rongguang Ye, Wei-Bin Kou, and Ming Tang.
\newblock Praffl: A preference-aware scheme in fair federated learning.
\newblock \emph{arXiv preprint arXiv:2404.08973}, 2024{\natexlab{a}}.

\bibitem[Ye et~al.(2024{\natexlab{b}})Ye, Wang, Chai, Li, Li, Xu, Du, Wang, and Chen]{ye2024openfedllm}
Rui Ye, Wenhao Wang, Jingyi Chai, Dihan Li, Zexi Li, Yinda Xu, Yaxin Du, Yanfeng Wang, and Siheng Chen.
\newblock Openfedllm: Training large language models on decentralized private data via federated learning.
\newblock In \emph{Proceedings of the 30th ACM SIGKDD conference on knowledge discovery and data mining}, pp.\  6137--6147, 2024{\natexlab{b}}.

\bibitem[Yen et~al.(2024)Yen, Gao, Hou, Ding, Fleischer, Izsak, Wasserblat, and Chen]{yen2024helmet}
Howard Yen, Tianyu Gao, Minmin Hou, Ke~Ding, Daniel Fleischer, Peter Izsak, Moshe Wasserblat, and Danqi Chen.
\newblock Helmet: How to evaluate long-context language models effectively and thoroughly.
\newblock \emph{arXiv preprint arXiv:2410.02694}, 2024.

\bibitem[Yi et~al.(2022)Yi, Wu, Zhang, Zhu, Qi, Sun, and Xie]{yi2022robust}
Jingwei Yi, Fangzhao Wu, Huishuai Zhang, Bin Zhu, Tao Qi, Guangzhong Sun, and Xing Xie.
\newblock Robust quantity-aware aggregation for federated learning.
\newblock \emph{arXiv preprint arXiv:2205.10848}, 2022.

\bibitem[Yi et~al.(2023)Yi, Yu, Wang, and Liu]{yi2023fedlora}
Liping Yi, Han Yu, Gang Wang, and Xiaoguang Liu.
\newblock Fedlora: Model-heterogeneous personalized federated learning with lora tuning.
\newblock \emph{arXiv preprint arXiv:2310.13283}, 2023.

\bibitem[Yi et~al.(2025)Yi, Zhang, Chen, Bao, and Xing]{yi2025fedfld}
Xiaoyang Yi, Jian Zhang, Jing Chen, Yuru Bao, and Lingkai Xing.
\newblock Fedfld: Heterogeneous federated learning via forget-less distillation.
\newblock In \emph{ICASSP 2025-2025 IEEE International Conference on Acoustics, Speech and Signal Processing (ICASSP)}, pp.\  1--5. IEEE, 2025.

\bibitem[Yin et~al.(2023)Yin, Huang, and Wan]{yin2023alcuna}
Xunjian Yin, Baizhou Huang, and Xiaojun Wan.
\newblock Alcuna: Large language models meet new knowledge.
\newblock \emph{arXiv preprint arXiv:2310.14820}, 2023.

\bibitem[Yoon et~al.(2021)Yoon, Jeong, Lee, Yang, and Hwang]{yoon2021federated}
Jaehong Yoon, Wonyong Jeong, Giwoong Lee, Eunho Yang, and Sung~Ju Hwang.
\newblock Federated continual learning with weighted inter-client transfer.
\newblock In \emph{International Conference on Machine Learning}, pp.\  12073--12086. PMLR, 2021.

\bibitem[Yu et~al.(2024)Yu, Yang, Gao, Kang, Wang, Zhang, and Li]{yu2024personalized}
Hao Yu, Xin Yang, Xin Gao, Yan Kang, Hao Wang, Junbo Zhang, and Tianrui Li.
\newblock Personalized federated continual learning via multi-granularity prompt.
\newblock In \emph{Proceedings of the 30th ACM SIGKDD Conference on Knowledge Discovery and Data Mining}, pp.\  4023--4034, 2024.

\bibitem[Yu et~al.(2023{\natexlab{a}})Yu, Wang, Tu, Cao, Zhang-Li, Lv, Peng, Yao, Zhang, Li, et~al.]{yu2023kola}
Jifan Yu, Xiaozhi Wang, Shangqing Tu, Shulin Cao, Daniel Zhang-Li, Xin Lv, Hao Peng, Zijun Yao, Xiaohan Zhang, Hanming Li, et~al.
\newblock Kola: Carefully benchmarking world knowledge of large language models.
\newblock \emph{arXiv preprint arXiv:2306.09296}, 2023{\natexlab{a}}.

\bibitem[Yu et~al.(2023{\natexlab{b}})Yu, Jiang, Shi, Yu, Liu, Zhang, Kwok, Li, Weller, and Liu]{yu2023metamath}
Longhui Yu, Weisen Jiang, Han Shi, Jincheng Yu, Zhengying Liu, Yu~Zhang, James~T Kwok, Zhenguo Li, Adrian Weller, and Weiyang Liu.
\newblock Metamath: Bootstrap your own mathematical questions for large language models.
\newblock \emph{arXiv preprint arXiv:2309.12284}, 2023{\natexlab{b}}.

\bibitem[Yu et~al.(2023{\natexlab{c}})Yu, Liu, Wang, Xu, and Liu]{yu2023multimodal}
Qiying Yu, Yang Liu, Yimu Wang, Ke~Xu, and Jingjing Liu.
\newblock Multimodal federated learning via contrastive representation ensemble.
\newblock \emph{arXiv preprint arXiv:2302.08888}, 2023{\natexlab{c}}.

\bibitem[Yu et~al.(2025)Yu, Zhang, Zhu, Yuan, Zuo, Yue, Fan, Liu, Liu, Liu, Lin, Lin, Ma, Sheng, Tong, Zhang, Zhang, Zhang, Zhu, Zhu, Chen, Chen, Wang, Yu, Dai, Song, Wei, Zhou, Liu, Ma, Zhang, Yan, Qiao, Wu, and Wang]{yu2025dapoopensourcellmreinforcement}
Qiying Yu, Zheng Zhang, Ruofei Zhu, Yufeng Yuan, Xiaochen Zuo, Yu~Yue, Tiantian Fan, Gaohong Liu, Lingjun Liu, Xin Liu, Haibin Lin, Zhiqi Lin, Bole Ma, Guangming Sheng, Yuxuan Tong, Chi Zhang, Mofan Zhang, Wang Zhang, Hang Zhu, Jinhua Zhu, Jiaze Chen, Jiangjie Chen, Chengyi Wang, Hongli Yu, Weinan Dai, Yuxuan Song, Xiangpeng Wei, Hao Zhou, Jingjing Liu, Wei-Ying Ma, Ya-Qin Zhang, Lin Yan, Mu~Qiao, Yonghui Wu, and Mingxuan Wang.
\newblock Dapo: An open-source llm reinforcement learning system at scale, 2025.
\newblock URL \url{https://arxiv.org/abs/2503.14476}.

\bibitem[Yu et~al.(2023{\natexlab{d}})Yu, Mu{\~n}oz, and Jannesari]{yu2023bridging}
Sixing Yu, J~Pablo Mu{\~n}oz, and Ali Jannesari.
\newblock Bridging the gap between foundation models and heterogeneous federated learning.
\newblock \emph{arXiv preprint arXiv:2310.00247}, 2023{\natexlab{d}}.

\bibitem[Yu et~al.(2022)Yu, Zhu, Li, Hu, Wang, Ji, and Jiang]{yu2022survey}
Wenhao Yu, Chenguang Zhu, Zaitang Li, Zhiting Hu, Qingyun Wang, Heng Ji, and Meng Jiang.
\newblock A survey of knowledge-enhanced text generation.
\newblock \emph{ACM Computing Surveys}, 54\penalty0 (11s):\penalty0 1--38, 2022.

\bibitem[Yuan et~al.(2024{\natexlab{a}})Yuan, Wang, Sun, Philip, and Brinton]{yuan2024decentralized}
Liangqi Yuan, Ziran Wang, Lichao Sun, S~Yu Philip, and Christopher~G Brinton.
\newblock Decentralized federated learning: A survey and perspective.
\newblock \emph{IEEE Internet of Things Journal}, 2024{\natexlab{a}}.

\bibitem[Yuan et~al.(2023)Yuan, Chen, Cui, Gao, Zou, Cheng, Ji, Liu, and Sun]{yuan2023revisiting}
Lifan Yuan, Yangyi Chen, Ganqu Cui, Hongcheng Gao, Fangyuan Zou, Xingyi Cheng, Heng Ji, Zhiyuan Liu, and Maosong Sun.
\newblock Revisiting out-of-distribution robustness in nlp: Benchmarks, analysis, and llms evaluations.
\newblock \emph{Advances in Neural Information Processing Systems}, 36:\penalty0 58478--58507, 2023.

\bibitem[Yuan et~al.(2024{\natexlab{b}})Yuan, Ning, Zhou, Yang, Li, Zhuang, Tan, Yao, Lin, Li, et~al.]{yuan2024lv}
Tao Yuan, Xuefei Ning, Dong Zhou, Zhijie Yang, Shiyao Li, Minghui Zhuang, Zheyue Tan, Zhuyu Yao, Dahua Lin, Boxun Li, et~al.
\newblock Lv-eval: A balanced long-context benchmark with 5 length levels up to 256k.
\newblock \emph{arXiv preprint arXiv:2402.05136}, 2024{\natexlab{b}}.

\bibitem[Yuan et~al.(2024{\natexlab{c}})Yuan, Yang, Ye, Chen, Hung, and Yin]{yuan2024fellas}
Wei Yuan, Chaoqun Yang, Guanhua Ye, Tong Chen, Nguyen Quoc~Viet Hung, and Hongzhi Yin.
\newblock Fellas: Enhancing federated sequential recommendation with llm as external services.
\newblock \emph{ACM Transactions on Information Systems}, 2024{\natexlab{c}}.

\bibitem[Yue et~al.(2023{\natexlab{a}})Yue, Chen, Wang, Li, Shen, Liu, Zhou, Xiao, Yun, Huang, et~al.]{yue2023disc}
Shengbin Yue, Wei Chen, Siyuan Wang, Bingxuan Li, Chenchen Shen, Shujun Liu, Yuxuan Zhou, Yao Xiao, Song Yun, Xuanjing Huang, et~al.
\newblock Disc-lawllm: Fine-tuning large language models for intelligent legal services.
\newblock \emph{arXiv preprint arXiv:2309.11325}, 2023{\natexlab{a}}.

\bibitem[Yue et~al.(2023{\natexlab{b}})Yue, Qu, Zhang, Fu, Huang, Sun, Su, and Chen]{yue2023mammoth}
Xiang Yue, Xingwei Qu, Ge~Zhang, Yao Fu, Wenhao Huang, Huan Sun, Yu~Su, and Wenhu Chen.
\newblock Mammoth: Building math generalist models through hybrid instruction tuning.
\newblock \emph{arXiv preprint arXiv:2309.05653}, 2023{\natexlab{b}}.

\bibitem[Zaken et~al.(2021)Zaken, Ravfogel, and Goldberg]{zaken2021bitfit}
Elad~Ben Zaken, Shauli Ravfogel, and Yoav Goldberg.
\newblock Bitfit: Simple parameter-efficient fine-tuning for transformer-based masked language-models.
\newblock \emph{arXiv preprint arXiv:2106.10199}, 2021.

\bibitem[Zellers et~al.(2019)Zellers, Holtzman, Bisk, Farhadi, and Choi]{zellers2019hellaswag}
Rowan Zellers, Ari Holtzman, Yonatan Bisk, Ali Farhadi, and Yejin Choi.
\newblock Hellaswag: Can a machine really finish your sentence?
\newblock \emph{arXiv preprint arXiv:1905.07830}, 2019.

\bibitem[Zeng et~al.(2020)Zeng, Yang, Ju, Yang, Wang, Zhang, Zhou, Zeng, Dong, Zhang, et~al.]{zeng2020meddialog}
Guangtao Zeng, Wenmian Yang, Zeqian Ju, Yue Yang, Sicheng Wang, Ruisi Zhang, Meng Zhou, Jiaqi Zeng, Xiangyu Dong, Ruoyu Zhang, et~al.
\newblock Meddialog: Large-scale medical dialogue datasets.
\newblock In \emph{Proceedings of the 2020 conference on empirical methods in natural language processing (EMNLP)}, pp.\  9241--9250, 2020.

\bibitem[Zeng(2023)]{zeng2023measuring}
Hui Zeng.
\newblock Measuring massive multitask chinese understanding.
\newblock \emph{arXiv preprint arXiv:2304.12986}, 2023.

\bibitem[Zeng et~al.(2024{\natexlab{a}})Zeng, Yue, Jiang, and Wang]{zeng2024federated}
Huimin Zeng, Zhenrui Yue, Qian Jiang, and Dong Wang.
\newblock Federated recommendation via hybrid retrieval augmented generation.
\newblock \emph{arXiv preprint arXiv:2403.04256}, 2024{\natexlab{a}}.

\bibitem[Zeng et~al.(2024{\natexlab{b}})Zeng, Chen, Deng, Chen, Mao, and Li]{zeng2024fine}
Juntao Zeng, Bo~Chen, Yuandan Deng, Weiqin Chen, Yanlin Mao, and Jiawei Li.
\newblock Fine-tuning of financial large language model and application at edge device.
\newblock In \emph{Proceedings of the 3rd International Conference on Computer, Artificial Intelligence and Control Engineering}, pp.\  42--47, 2024{\natexlab{b}}.

\bibitem[Zhan et~al.(2024)Zhan, Wu, Tian, Zhao, and Li]{zhan2024heterogeneity}
Shichen Zhan, Yebo Wu, Chunlin Tian, Yan Zhao, and Li~Li.
\newblock Heterogeneity-aware coordination for federated learning via stitching pre-trained blocks.
\newblock In \emph{2024 IEEE/ACM 32nd International Symposium on Quality of Service (IWQoS)}, pp.\  1--10. IEEE, 2024.

\bibitem[Zhang et~al.(2023{\natexlab{a}})Zhang, Yang, and Liu]{zhang2023instructfingpt}
Boyu Zhang, Hongyang Yang, and Xiao-Yang Liu.
\newblock Instruct-fingpt: Financial sentiment analysis by instruction tuning of general-purpose large language models.
\newblock \emph{FinLLM Symposium at IJCAI 2023}, 2023{\natexlab{a}}.

\bibitem[Zhang et~al.(2024{\natexlab{a}})Zhang, Yu, Dong, Li, Su, Chu, and Yu]{zhang2024mm}
Duzhen Zhang, Yahan Yu, Jiahua Dong, Chenxing Li, Dan Su, Chenhui Chu, and Dong Yu.
\newblock Mm-llms: Recent advances in multimodal large language models.
\newblock \emph{arXiv preprint arXiv:2401.13601}, 2024{\natexlab{a}}.

\bibitem[Zhang et~al.(2024{\natexlab{b}})Zhang, Vahidian, Kuo, Li, Zhang, Yu, Wang, and Chen]{zhang2024towards}
Jianyi Zhang, Saeed Vahidian, Martin Kuo, Chunyuan Li, Ruiyi Zhang, Tong Yu, Guoyin Wang, and Yiran Chen.
\newblock Towards building the federatedgpt: Federated instruction tuning.
\newblock In \emph{ICASSP 2024-2024 IEEE International Conference on Acoustics, Speech and Signal Processing (ICASSP)}, pp.\  6915--6919. IEEE, 2024{\natexlab{b}}.

\bibitem[Zhang et~al.(2024{\natexlab{c}})Zhang, Yu, Dai, and Xu]{zhang2024citalaw}
Kepu Zhang, Weijie Yu, Sunhao Dai, and Jun Xu.
\newblock Citalaw: Enhancing llm with citations in legal domain.
\newblock \emph{arXiv preprint arXiv:2412.14556}, 2024{\natexlab{c}}.

\bibitem[Zhang et~al.(2023{\natexlab{b}})Zhang, Li, Liu, Liu, Chen, Luo, Yang, et~al.]{zhang2023marathon}
Lei Zhang, Yunshui Li, Ziqiang Liu, Junhao Liu, Longze Chen, Run Luo, Min Yang, et~al.
\newblock Marathon: A race through the realm of long context with large language models.
\newblock \emph{arXiv preprint arXiv:2312.09542}, 2023{\natexlab{b}}.

\bibitem[Zhang et~al.(2023{\natexlab{c}})Zhang, Cai, Liu, Yang, Dai, Liao, Qin, Li, Liu, Liu, et~al.]{zhang2023fineval}
Liwen Zhang, Weige Cai, Zhaowei Liu, Zhi Yang, Wei Dai, Yujie Liao, Qianru Qin, Yifei Li, Xingyu Liu, Zhiqiang Liu, et~al.
\newblock Fineval: A chinese financial domain knowledge evaluation benchmark for large language models.
\newblock \emph{arXiv preprint arXiv:2308.09975}, 2023{\natexlab{c}}.

\bibitem[Zhang et~al.(2021)Zhang, Chen, Bi, Liang, Li, Shang, Yin, Tan, Xu, Huang, et~al.]{zhang2021cblue}
Ningyu Zhang, Mosha Chen, Zhen Bi, Xiaozhuan Liang, Lei Li, Xin Shang, Kangping Yin, Chuanqi Tan, Jian Xu, Fei Huang, et~al.
\newblock Cblue: A chinese biomedical language understanding evaluation benchmark.
\newblock \emph{arXiv preprint arXiv:2106.08087}, 2021.

\bibitem[Zhang et~al.(2024{\natexlab{d}})Zhang, Zeng, Wang, and Lu]{zhang2024tinyllama}
Peiyuan Zhang, Guangtao Zeng, Tianduo Wang, and Wei Lu.
\newblock Tinyllama: An open-source small language model.
\newblock \emph{arXiv preprint arXiv:2401.02385}, 2024{\natexlab{d}}.

\bibitem[Zhang et~al.(2024{\natexlab{e}})Zhang, Zhou, Hu, Feng, Weng, and Chen]{zhang2024personalized}
Pengyu Zhang, Yingbo Zhou, Ming Hu, Junxian Feng, Jiawen Weng, and Mingsong Chen.
\newblock Personalized federated instruction tuning via neural architecture search.
\newblock \emph{arXiv preprint arXiv:2402.16919}, 2024{\natexlab{e}}.

\bibitem[Zhang et~al.(2024{\natexlab{f}})Zhang, Jiang, Zhang, Lin, Guo, Qiu, Zhou, Lu, Chang, Qiao, et~al.]{zhang2024mathverse}
Renrui Zhang, Dongzhi Jiang, Yichi Zhang, Haokun Lin, Ziyu Guo, Pengshuo Qiu, Aojun Zhou, Pan Lu, Kai-Wei Chang, Yu~Qiao, et~al.
\newblock Mathverse: Does your multi-modal llm truly see the diagrams in visual math problems?
\newblock In \emph{European Conference on Computer Vision}, pp.\  169--186. Springer, 2024{\natexlab{f}}.

\bibitem[Zhang et~al.(2023{\natexlab{d}})Zhang, Fang, Zhang, Ma, Zhou, Huang, Bu, Gui, Chen, Chen, et~al.]{zhang2023bayling}
Shaolei Zhang, Qingkai Fang, Zhuocheng Zhang, Zhengrui Ma, Yan Zhou, Langlin Huang, Mengyu Bu, Shangtong Gui, Yunji Chen, Xilin Chen, et~al.
\newblock Bayling: Bridging cross-lingual alignment and instruction following through interactive translation for large language models.
\newblock \emph{arXiv preprint arXiv:2306.10968}, 2023{\natexlab{d}}.

\bibitem[Zhang et~al.(2023{\natexlab{e}})Zhang, Dong, Li, Zhang, Sun, Wang, Li, Hu, Zhang, Wu, et~al.]{zhang2023instruction}
Shengyu Zhang, Linfeng Dong, Xiaoya Li, Sen Zhang, Xiaofei Sun, Shuhe Wang, Jiwei Li, Runyi Hu, Tianwei Zhang, Fei Wu, et~al.
\newblock Instruction tuning for large language models: A survey.
\newblock \emph{arXiv preprint arXiv:2308.10792}, 2023{\natexlab{e}}.

\bibitem[Zhang et~al.(2023{\natexlab{f}})Zhang, Aljunied, Gao, Chia, and Bing]{zhang2023m3exam}
Wenxuan Zhang, Mahani Aljunied, Chang Gao, Yew~Ken Chia, and Lidong Bing.
\newblock M3exam: A multilingual, multimodal, multilevel benchmark for examining large language models.
\newblock \emph{Advances in Neural Information Processing Systems}, 36:\penalty0 5484--5505, 2023{\natexlab{f}}.

\bibitem[Zhang et~al.(2024{\natexlab{g}})Zhang, Xiang, Yuan, Feng, Han, Lopez-Lira, Liu, Qiu, Ananiadou, Peng, et~al.]{zhang2024dolares}
Xiao Zhang, Ruoyu Xiang, Chenhan Yuan, Duanyu Feng, Weiguang Han, Alejandro Lopez-Lira, Xiao-Yang Liu, Meikang Qiu, Sophia Ananiadou, Min Peng, et~al.
\newblock D{\'o}lares or dollars? unraveling the bilingual prowess of financial llms between spanish and english.
\newblock In \emph{Proceedings of the 30th ACM SIGKDD Conference on Knowledge Discovery and Data Mining}, pp.\  6236--6246, 2024{\natexlab{g}}.

\bibitem[Zhang et~al.(2023{\natexlab{g}})Zhang, Yu, Yu, Lv, Liu, Huang, Xu, and Li]{zhang2023wider}
Xinghua Zhang, Bowen Yu, Haiyang Yu, Yangyu Lv, Tingwen Liu, Fei Huang, Hongbo Xu, and Yongbin Li.
\newblock Wider and deeper llm networks are fairer llm evaluators.
\newblock \emph{arXiv preprint arXiv:2308.01862}, 2023{\natexlab{g}}.

\bibitem[Zhang et~al.(2023{\natexlab{h}})Zhang, Tian, Yang, Chen, Li, and Petzold]{zhang2023alpacare}
Xinlu Zhang, Chenxin Tian, Xianjun Yang, Lichang Chen, Zekun Li, and Linda~Ruth Petzold.
\newblock Alpacare: Instruction-tuned large language models for medical application.
\newblock \emph{arXiv preprint arXiv:2310.14558}, 2023{\natexlab{h}}.

\bibitem[Zhang et~al.(2024{\natexlab{h}})Zhang, Chen, Hu, Xu, Chen, Hao, Han, Thai, Wang, Liu, et~al.]{zhang2024inftybench}
Xinrong Zhang, Yingfa Chen, Shengding Hu, Zihang Xu, Junhao Chen, Moo~Khai Hao, Xu~Han, Zhen~Leng Thai, Shuo Wang, Zhiyuan Liu, et~al.
\newblock {\i}nftybench: Extending long context evaluation beyond 100k tokens.
\newblock In \emph{ACL (1)}, 2024{\natexlab{h}}.

\bibitem[Zhang et~al.(2024{\natexlab{i}})Zhang, Qin, Wu, and Deng]{zhang2024personalized_new}
Yicheng Zhang, Zhen Qin, Zhaomin Wu, and Shuiguang Deng.
\newblock Personalized federated fine-tuning for llms via data-driven heterogeneous model architectures.
\newblock \emph{arXiv preprint arXiv:2411.19128}, 2024{\natexlab{i}}.

\bibitem[Zhang et~al.(2024{\natexlab{j}})Zhang, Zeng, Luo, Fu, Chen, Xu, and King]{zhang2024survey}
Yifei Zhang, Dun Zeng, Jinglong Luo, Xinyu Fu, Guanzhong Chen, Zenglin Xu, and Irwin King.
\newblock A survey of trustworthy federated learning: Issues, solutions, and challenges.
\newblock \emph{ACM Transactions on Intelligent Systems and Technology}, 15\penalty0 (6):\penalty0 1--47, 2024{\natexlab{j}}.

\bibitem[Zhang et~al.(2024{\natexlab{k}})Zhang, Cao, Ye, Ma, Liao, and Chua]{zhang2024analyzing}
Zhihan Zhang, Yixin Cao, Chenchen Ye, Yunshan Ma, Lizi Liao, and Tat-Seng Chua.
\newblock Analyzing temporal complex events with large language models? a benchmark towards temporal, long context understanding.
\newblock \emph{arXiv preprint arXiv:2406.02472}, 2024{\natexlab{k}}.

\bibitem[Zhang et~al.(2024{\natexlab{l}})Zhang, Xu, Liu, and Hu]{zhang2024fed}
Zikai Zhang, Jiahao Xu, Ping Liu, and Rui Hu.
\newblock Fed-pilot: Optimizing lora assignment for efficient federated foundation model fine-tuning.
\newblock \emph{arXiv preprint arXiv:2410.10200}, 2024{\natexlab{l}}.

\bibitem[Zhao et~al.(2023{\natexlab{a}})Zhao, Du, Li, Li, and Liu]{zhao2023fedprompt}
Haodong Zhao, Wei Du, Fangqi Li, Peixuan Li, and Gongshen Liu.
\newblock Fedprompt: Communication-efficient and privacy-preserving prompt tuning in federated learning.
\newblock In \emph{ICASSP 2023-2023 IEEE International Conference on Acoustics, Speech and Signal Processing (ICASSP)}, pp.\  1--5. IEEE, 2023{\natexlab{a}}.

\bibitem[Zhao et~al.(2024{\natexlab{a}})Zhao, Wang, Xu, Ren, Ng, and Chua]{zhao2024llm}
Jujia Zhao, Wenjie Wang, Chen Xu, Zhaochun Ren, See-Kiong Ng, and Tat-Seng Chua.
\newblock Llm-based federated recommendation.
\newblock \emph{arXiv preprint arXiv:2402.09959}, 2024{\natexlab{a}}.

\bibitem[Zhao et~al.(2024{\natexlab{b}})Zhao, Chen, Lee, Qiu, Gao, Fan, and Lane]{zhao2024breaking}
Wanru Zhao, Yihong Chen, Royson Lee, Xinchi Qiu, Yan Gao, Hongxiang Fan, and Nicholas~Donald Lane.
\newblock Breaking physical and linguistic borders: Multilingual federated prompt tuning for low-resource languages.
\newblock In \emph{The Twelfth International Conference on Learning Representations}, 2024{\natexlab{b}}.

\bibitem[Zhao et~al.(2023{\natexlab{b}})Zhao, Zhou, Li, Tang, Wang, Hou, Min, Zhang, Zhang, Dong, et~al.]{zhao2023survey}
Wayne~Xin Zhao, Kun Zhou, Junyi Li, Tianyi Tang, Xiaolei Wang, Yupeng Hou, Yingqian Min, Beichen Zhang, Junjie Zhang, Zican Dong, et~al.
\newblock A survey of large language models.
\newblock \emph{arXiv preprint arXiv:2303.18223}, 2023{\natexlab{b}}.

\bibitem[Zheng et~al.(2021)Zheng, Han, and Polu]{zheng2021minif2f}
Kunhao Zheng, Jesse~Michael Han, and Stanislas Polu.
\newblock Minif2f: a cross-system benchmark for formal olympiad-level mathematics.
\newblock \emph{arXiv preprint arXiv:2109.00110}, 2021.

\bibitem[Zheng et~al.(2023{\natexlab{a}})Zheng, Chiang, Sheng, Zhuang, Wu, Zhuang, Lin, Li, Li, Xing, Zhang, Gonzalez, and Stoica]{zheng2023judging}
Lianmin Zheng, Wei-Lin Chiang, Ying Sheng, Siyuan Zhuang, Zhanghao Wu, Yonghao Zhuang, Zi~Lin, Zhuohan Li, Dacheng Li, Eric.~P Xing, Hao Zhang, Joseph~E. Gonzalez, and Ion Stoica.
\newblock Judging llm-as-a-judge with mt-bench and chatbot arena, 2023{\natexlab{a}}.

\bibitem[Zheng et~al.(2019)Zheng, Cao, Xu, and Bian]{zheng2019doc2edag}
Shun Zheng, Wei Cao, Wei Xu, and Jiang Bian.
\newblock Doc2edag: An end-to-end document-level framework for chinese financial event extraction.
\newblock \emph{arXiv preprint arXiv:1904.07535}, 2019.

\bibitem[Zheng et~al.(2023{\natexlab{b}})Zheng, Ning, Wang, Zhang, Zheng, Ye, and Chen]{zheng2023survey}
Zibin Zheng, Kaiwen Ning, Yanlin Wang, Jingwen Zhang, Dewu Zheng, Mingxi Ye, and Jiachi Chen.
\newblock A survey of large language models for code: Evolution, benchmarking, and future trends.
\newblock \emph{arXiv preprint arXiv:2311.10372}, 2023{\natexlab{b}}.

\bibitem[Zhong et~al.(2023)Zhong, Cui, Guo, Liang, Lu, Wang, Saied, Chen, and Duan]{zhong2023agieval}
Wanjun Zhong, Ruixiang Cui, Yiduo Guo, Yaobo Liang, Shuai Lu, Yanlin Wang, Amin Saied, Weizhu Chen, and Nan Duan.
\newblock Agieval: A human-centric benchmark for evaluating foundation models.
\newblock \emph{arXiv preprint arXiv:2304.06364}, 2023.

\bibitem[Zhou et~al.(2023)Zhou, Lu, Mishra, Brahma, Basu, Luan, Zhou, and Hou]{zhou2023instruction}
Jeffrey Zhou, Tianjian Lu, Swaroop Mishra, Siddhartha Brahma, Sujoy Basu, Yi~Luan, Denny Zhou, and Le~Hou.
\newblock Instruction-following evaluation for large language models.
\newblock \emph{arXiv preprint arXiv:2311.07911}, 2023.

\bibitem[Zhu et~al.(2021)Zhu, Lei, Wang, Zheng, Poria, and Chua]{zhu2021retrieving}
Fengbin Zhu, Wenqiang Lei, Chao Wang, Jianming Zheng, Soujanya Poria, and Tat-Seng Chua.
\newblock Retrieving and reading: A comprehensive survey on open-domain question answering.
\newblock \emph{arXiv preprint arXiv:2101.00774}, 2021.

\bibitem[Zhu et~al.(2023{\natexlab{a}})Zhu, Wang, Zhou, Wang, Chen, Wang, Yang, Ye, Zhang, Zhenqiang~Gong, et~al.]{zhu2023promptbench}
Kaijie Zhu, Jindong Wang, Jiaheng Zhou, Zichen Wang, Hao Chen, Yidong Wang, Linyi Yang, Wei Ye, Yue Zhang, Neil Zhenqiang~Gong, et~al.
\newblock Promptbench: Towards evaluating the robustness of large language models on adversarial prompts.
\newblock \emph{arXiv e-prints}, pp.\  arXiv--2306, 2023{\natexlab{a}}.

\bibitem[Zhu et~al.(2023{\natexlab{b}})Zhu, Wang, and Wang]{zhu2023chatmed}
Wei Zhu, Xiaoling Wang, and Longyue Wang.
\newblock Chatmed: A chinese medical large language model.
\newblock \emph{Retrieved September}, 18:\penalty0 2023, 2023{\natexlab{b}}.

\bibitem[Zhu \& Wang(2023)Zhu and Wang]{zhu2023shennong}
WY~Wei Zhu and Xiaoling Wang.
\newblock Shennong-tcm: A traditional chinese medicine large language model.
\newblock \emph{GitHub}, 2023.

\bibitem[Zhuang et~al.(2023)Zhuang, Chen, and Lyu]{zhuang2023foundation}
Weiming Zhuang, Chen Chen, and Lingjuan Lyu.
\newblock When foundation model meets federated learning: Motivations, challenges, and future directions.
\newblock \emph{arXiv preprint arXiv:2306.15546}, 2023.

\bibitem[Zhuo et~al.(2024)Zhuo, Vu, Chim, Hu, Yu, Widyasari, Yusuf, Zhan, He, Paul, et~al.]{zhuo2024bigcodebench}
Terry~Yue Zhuo, Minh~Chien Vu, Jenny Chim, Han Hu, Wenhao Yu, Ratnadira Widyasari, Imam Nur~Bani Yusuf, Haolan Zhan, Junda He, Indraneil Paul, et~al.
\newblock Bigcodebench: Benchmarking code generation with diverse function calls and complex instructions.
\newblock \emph{arXiv preprint arXiv:2406.15877}, 2024.

\bibitem[Zou et~al.(2024)Zou, Khalifa, and Wang]{zou2024retrieval}
Kaijian Zou, Muhammad Khalifa, and Lu~Wang.
\newblock Retrieval or global context understanding? on many-shot in-context learning for long-context evaluation.
\newblock \emph{arXiv preprint arXiv:2411.07130}, 2024.

\bibitem[Zuo et~al.(2025)Zuo, Qu, Li, Chen, Zhu, Hua, Zhang, Ding, and Zhou]{zuo2025medxpertqa}
Yuxin Zuo, Shang Qu, Yifei Li, Zhangren Chen, Xuekai Zhu, Ermo Hua, Kaiyan Zhang, Ning Ding, and Bowen Zhou.
\newblock Medxpertqa: Benchmarking expert-level medical reasoning and understanding.
\newblock \emph{arXiv preprint arXiv:2501.18362}, 2025.

\end{thebibliography}
\bibliographystyle{tmlr}


\end{document}